\documentclass[10pt,journal,compsoc]{IEEEtran}

\usepackage{amsmath,amssymb,amsthm}
\usepackage[noend]{algcompatible}
\usepackage{algorithm}
\usepackage{graphics}
\ifCLASSOPTIONcompsoc
  \usepackage[nocompress]{cite}
\else
  \usepackage{cite}
\fi
\usepackage{color}
\usepackage{ragged2e}
\usepackage{booktabs}
\usepackage{stfloats}
\usepackage{graphicx}
\usepackage[normalem]{ulem}
\usepackage{overpic}
\usepackage{multirow}
\usepackage{makecell}
\usepackage{colortbl}
\usepackage[breaklinks=true,colorlinks]{hyperref}
\usepackage{pifont} %
\usepackage{tikz}
\usepackage{silence}
\usepackage{cuted}
\usepackage[section]{placeins}
\usepackage{enumitem}
\usepackage{makecell}
\usepackage{threeparttable} 
\usepackage{needspace}
\usetikzlibrary{decorations,arrows}
\usetikzlibrary{decorations.markings}
\graphicspath{{./Imgs/},{./Imgs/authors/}}
\DeclareGraphicsExtensions{.pdf,.jpg,.png}

\newcommand{\figref}[1]{Fig.~\ref{#1}}
\newcommand{\tabref}[1]{Table~\ref{#1}}
\newcommand{\equref}[1]{Equ.~\ref{#1}}

\ifdefined \GramaCheck
  \newcommand{\CheckRmv}[1]{}
  \renewcommand{\equref}[1]{Equation 1}
  \renewcommand{\figref}[1]{Figure 1}
  \renewcommand{\tabref}[1]{Table 1}
\else
  \newcommand{\CheckRmv}[1]{#1}
\fi

\definecolor{mygray}{gray}{.92}

\theoremstyle{remark}

\begin{document}

\title{Structured Analytic Coherent Point Drift for Non-Rigid Point Set Registration}

\author{Wei~Feng and~Haiyong~Zheng,~\IEEEmembership{Senior~Member,~IEEE}% <-this % stops a space
\IEEEcompsocitemizethanks{%
\IEEEcompsocthanksitem The authors are with the College of Electronic Engineering, Ocean University of China, Qingdao 266404, China.\protect\\
E-mails: weifeng@stu.ouc.edu.cn, zhenghaiyong@ouc.edu.cn.\protect\\
Corresponding author: Haiyong~Zheng.
}
%E-mail: see http://www.michaelshell.org/contact.html
}

\IEEEtitleabstractindextext{%
\begin{abstract} \justifying
Coherent Point Drift (CPD) is a representative probabilistic framework for unsupervised non-rigid point set registration, using Gaussian-mixture posterior correspondences to improve the capture range over hard-correspondence methods. However, its standard non-rigid M-step relies on a point-indexed Gaussian-kernel system whose size grows with the number of moving points,
making the deformation update computationally heavy for large point sets and difficult to control in complexity during registration.
To address these limitations and provide a structured analytic perspective on CPD, we propose \textbf{Analytic-CPD}, a new unsupervised non-rigid point set registration framework. Analytic-CPD retains the probabilistic posterior layer of CPD, but lifts the M-step from point-indexed kernel displacement estimation
to structured analytic mapping estimation. This formulation couples the Gaussian-mixture posterior mechanism of CPD with the smooth-map approximation capacity of Structured Analytic Mappings (SAM). It brings three consequences: the deformation coefficient dimension is governed by the ambient dimension and analytic order rather than by the number of moving points; the deformation
estimation is organized over an interpretable hierarchy of analytic function spaces; and the analytic order can be increased progressively as posterior correspondences become more reliable.
We implement this idea using an increasing-degree continuation strategy with decreasing stage lengths, in which low-order analytic maps first stabilize the posterior correspondence structure and higher-order modes later refine nonlinear residual deformation.Experiments on controlled model-matched, smooth model-mismatch, and registered human-shape data demonstrate the effectiveness and favorable accuracy--efficiency performance of Analytic-CPD. These results support structured analytic mappings as a compact, interpretable, and degree-controllable alternative to point-indexed kernel deformation models in CPD-type non-rigid registration. Code is available at \url{https://github.com/monge-ampere/Analytic-CPD}.
\end{abstract}

}

\maketitle

\IEEEdisplaynontitleabstractindextext

% \fi
\IEEEpeerreviewmaketitle

\IEEEraisesectionheading{\section{Introduction}}\label{sec:intro}

\IEEEPARstart{N}{on-rigid} point set registration is governed by a
correspondence--mapping duality. Correspondences determine how source points should be transported, while the deformation model determines which correspondences are geometrically plausible. Different registration regimes resolve this duality in different ways. When the relation between two point sets is close to a deterministic geometric transport, hard-correspondence
methods such as ICP~\cite{ICP1992} can be highly efficient. However, under large, ambiguous, noisy, or strongly non-rigid deformation, deterministic nearest-neighbor assignments may become unreliable. In such cases, soft-correspondence methods are more appropriate because they represent correspondence uncertainty through assignment weights or posterior probabilities.

Coherent Point Drift (CPD)~\cite{CPD2006,myronenko2010point} is one of the most influential probabilistic frameworks for soft-correspondence non-rigid registration. It models the moving points as Gaussian mixture centroids and estimates posterior correspondences through an EM-style procedure, which often
provides a larger capture range than hard-correspondence methods. However, in standard non-rigid CPD, the deformation is represented by a Gaussian-kernel displacement field whose coefficients are indexed by the moving points. Although this kernel formulation enforces coherent motion, its M-step involves
a dense system whose size grows with the number of moving points, making the deformation update computationally heavy and memory demanding for large point clouds. Moreover, the deformation complexity is controlled indirectly through kernel and regularization parameters, which makes it difficult to explicitly
coordinate the mapping capacity with the evolving reliability of posterior correspondences during registration.

This motivates us to seek a compact and explicitly controllable deformation estimator while preserving the probabilistic posterior correspondence layer of CPD. Structured Analytic Mappings (SAM)~\cite{feng2026structured} provide such a candidate. SAM represents smooth deformations by truncated multivariate Taylor expansions of vector-valued functions, organized into a nested hierarchy of finite-dimensional analytic function spaces. The coefficient dimension is controlled by the ambient dimension and analytic order, rather than by the number of points. This organization gives the deformation model an interpretable function-space structure and allows the mapping complexity to be regulated by analytic degree.

This paper integrates structured analytic mappings into the probabilistic registration framework of CPD. We propose \emph{Analytic-CPD}, a new unsupervised non-rigid point set registration framework that retains the CPD posterior correspondence layer while lifting the non-rigid M-step from
point-indexed kernel displacement estimation to structured analytic mapping estimation. At each iteration, the E-step computes CPD-style posterior correspondences from the current moving configuration. The posterior-weighted deformation update is then formulated as a weighted analytic fitting problem,
where the deformation is represented by a finite-dimensional truncated Taylor mapping of a vector-valued function. Equivalently, the posterior probabilities can be condensed into barycentric soft targets, which provide a clear algebraic
connection between probabilistic correspondence estimation and structured analytic deformation fitting.

This reformulation changes the structure of non-rigid CPD. Standard CPD estimates a point-indexed Gaussian-kernel displacement field, whereas Analytic-CPD estimates an explicit analytic mapping whose coefficient dimension depends on the analytic order and the ambient dimension, rather than on the
number of moving points. The moving set is updated compositionally by a sequence of analytic maps, so the accumulated deformation can acquire high effective expressive complexity even when each individual M-step uses a moderate Taylor
order. To improve stability under large deformation, we further introduce an increasing-degree continuation strategy with decreasing stage lengths. Low-order mappings are used in early iterations to stabilize the posterior correspondence structure, and higher-order modes are activated later for nonlinear refinement.

The main contributions of this paper are summarized as follows:
\begin{itemize}
    \item We propose \textbf{Analytic-CPD}, a new unsupervised non-rigid point-set registration framework that couples the Gaussian-mixture posterior mechanism of CPD with the structured analytic deformation model of SAM.

    \item We lift the non-rigid CPD M-step from point-indexed kernel displacement estimation to weighted structured analytic mapping estimation. Under the squared loss, the posterior-weighted objective can be equivalently expressed through barycentric soft targets.

    \item We introduce a compositional analytic update with an increasing-degree continuation strategy. By organizing deformation estimation over a hierarchy of analytic function spaces, the method coordinates deformation expressiveness with posterior reliability: low-order maps stabilize early
    correspondences, while higher-order modes refine nonlinear residual deformation.

    \item We conduct systematic experiments on controlled analytic deformations, smooth non-analytic model-mismatch tests, multi-model multi-seed statistics, and registered human-motion cases. The results demonstrate strong accuracy--efficiency performance against representative classical,
    probabilistic, analytic, and recent non-rigid registration baselines.
\end{itemize}

\section{Related Work}
\label{sec:related-work}

The proposed Analytic-CPD is related to three lines of work: ICP-type geometric registration, soft-correspondence probabilistic registration, and structured analytic deformation models. We briefly review these directions and clarify the position of our method.

\subsection{ICP-Type Geometric Registration}

Rigid point-set registration is a classical problem in computer vision, robotics, and geometric data analysis. The Iterative Closest Point (ICP) algorithm~\cite{ICP1992} and related iterative point matching methods~\cite{Zhang1994IterativePointMatching} remain representative geometric registration approaches. They alternate between hard correspondence assignment and transformation fitting, typically using nearest-neighbor or closest-point
matches followed by rigid transformation estimation.

Many variants have been proposed to improve robustness and convergence, including robust ICP~\cite{RICP}, trimmed ICP~\cite{trimmedICP}, generalized ICP~\cite{segal2009generalized}, point-to-plane
linearization~\cite{low2004linear}, globally optimal
registration~\cite{yang2015go}, and recent efficient
variants~\cite{zhang2021fast,lv2023kss}. Affine and non-rigid extensions further relax the rigid model by introducing global linear transformations, local affine models, embedded deformation graphs, or motion-regularized deformation
fields~\cite{affine-icp,ho2007new,jenkinson2001global,du2008affine,NICP2007,
Sumner2007EmbeddedDeformation,li2008global}.

ICP-type methods are efficient when reliable hard correspondences can be established. However, under large non-rigid deformation, early nearest-neighbor assignments may be inaccurate and can lead to poor local minima. This limitation motivates soft-correspondence and probabilistic formulations.

\subsection{Soft-Correspondence and Probabilistic Non-Rigid Registration}

Soft-correspondence methods replace deterministic point matches with assignment weights or posterior probabilities. TPS-RPM~\cite{Chui2003TPSRPM} combines deterministic annealing with thin-plate spline deformation~\cite{bookstein1989principal}. Gaussian mixture model based methods formulate point-set registration as probabilistic estimation~\cite{chui2000feature,jian2005robust,jian2010robust}. Among them, Coherent Point Drift (CPD)~\cite{CPD2006,myronenko2010point} is a representative
framework: the moving points are modeled as Gaussian mixture centroids, the fixed points are treated as observations, and EM alternates between posterior correspondence estimation and deformation-field update.

CPD-style registration has been extended in several directions, including ECMPR~\cite{Horaud2011ECMPR}, JRMPC~\cite{evangelidis2017joint}, generalized
CPD~\cite{Ravikumar2017GCPD}, anisotropic or normal-aware
variants~\cite{Min2021GeneralizedCPD}, Bayesian CPD~\cite{BCPD},
geodesic-based BCPD~\cite{hirose2022geodesic}, and accelerated
BCPD++~\cite{hirose2020acceleration}. Recent clustering-based methods, such as ClusterReg~\cite{zhao2024clustereg}, further reinterpret non-rigid registration through soft membership and clustering, while still relying on kernel-regularized
displacement fields.

Other probabilistic or structure-preserving methods incorporate filtering, global-local structure preservation, or topology-aware regularization to improve robustness and efficiency~\cite{gao2019filterreg,ma2015non}.

Despite these developments, most CPD-type non-rigid methods still represent deformation through point-indexed or kernel-indexed displacement fields. Their M-step is therefore closely tied to the moving point set, kernel matrix, or control centers. Analytic-CPD follows a different route: it retains the CPD posterior correspondence layer, but replaces the point-indexed Gaussian-kernel M-step with a finite-dimensional structured analytic mapping estimator.

\subsection{Structured Analytic Deformation Models}

Kernel, spline, and Gaussian basis models are widely used in non-rigid registration because they provide flexible smooth deformation fields. However, their parameterization is often tied to input points, control centers, or dense kernel matrices, which may lead to high-dimensional optimization, large memory
cost, and limited interpretability.

Structured Analytic Mappings (SAM)~\cite{feng2026structured} provide an alternative representation based on truncated multivariate Taylor expansions of vector-valued functions. The deformation model is organized by analytic degree rather than by point or kernel index, and its parameter dimension is controlled
by the ambient dimension and truncation order. Structured Taylor forms also provide a principled approximation framework for smooth registration mappings on compact domains.

Analytic-ICP~\cite{feng2026structured} embeds SAM into a hard-correspondence registration framework and unifies rigid, affine, and nonlinear deformation within a closed-form analytic hierarchy. However, as an ICP-type method, its capture range can be limited when nearest-neighbor correspondences are unreliable
under large deformation. This paper extends SAM to probabilistic
soft-correspondence registration. Analytic-CPD keeps the CPD posterior layer, condenses posterior probabilities into weighted soft targets, and estimates the M-step deformation by weighted structured analytic fitting. In this sense, Analytic-CPD is a structured analytic reformulation of the CPD non-rigid M-step:
it preserves the probabilistic correspondence mechanism of CPD while replacing the point-indexed kernel displacement field with a compact, degree-controllable analytic mapping space.

\section{Structured Analytic Coherent Point Drift}
\label{sec:method}

We build upon the structured analytic mapping framework of~\cite{feng2026structured}, where point-set registration is modeled by finite-dimensional analytic mappings derived from truncated multivariate Taylor expansions of vector-valued functions. In this work, we embed this mapping family into the coherent point drift framework and propose \textbf{Analytic-CPD}. The CPD posterior correspondence layer is retained, while the Gaussian-kernel displacement-field M-step of non-rigid CPD is replaced by a weighted structured analytic mapping fit.

At each iteration, the current moving points are regarded as Gaussian mixture centroids and the fixed points as observations. The E-step computes CPD-style posterior correspondence probabilities. These probabilities are then condensed, for each moving point, into a responsibility-weighted barycenter of the fixed point set. By the weighted least-squares barycentric identity, the original pairwise soft-correspondence objective is equivalent, up to a mapping-independent constant, to a weighted fitting problem between the current moving points and their soft targets.

The M-step therefore becomes a structured analytic mapping estimation problem.  Instead of estimating a point-indexed displacement field whose Gaussian kernel centers are all moving points, Analytic-CPD fits a truncated multivariate Taylor representation of a vector-valued mapping. In standard non-rigid CPD, adding more moving points also enlarges the deformation kernel system; in Analytic-CPD, the number of deformation parameters is determined by the spatial dimension and the analytic truncation order. The fitted analytic mapping is applied to the current moving configuration, yielding a compositional deformation process.

\subsection{Problem Formulation}
\label{subsec:problem-formulation}

Let
\[
X=\{x_n\}_{n=1}^{N}\subset\mathbb{R}^{d},
\qquad
Y=\{y_m\}_{m=1}^{M}\subset\mathbb{R}^{d},
\]
denote the fixed and moving point sets, respectively. The formulation is dimension-independent, and the experiments in this paper focus on the cases \(d=2\) and \(d=3\). The goal of non-rigid point-set registration is to deform the moving point set so that it is aligned with the fixed point set.

Following the CPD formulation, the current moving points are regarded as Gaussian mixture centroids, while the fixed points are treated as observed data. In contrast to non-rigid CPD, which estimates a point-indexed Gaussian-kernel displacement field, we estimate an iteration-wise structured analytic update map that transports the current moving configuration toward the posterior-induced soft targets.

Let
\[
Y^{(t)}=\{y_m^{(t)}\}_{m=1}^{M},
\qquad
Y^{(0)}=Y,
\]
be the moving point set at iteration \(t\). At this iteration, we seek an analytic update map
\[
\mathcal{A}_t(\cdot;\Theta_t,c):\mathbb{R}^{d}\rightarrow\mathbb{R}^{d},
\]
where \(\Theta_t\) denotes the analytic coefficients and \(c\) is the expansion center. The moving point set is updated by
\begin{equation}
Y^{(t+1)}
=
\mathcal{A}_t\bigl(Y^{(t)};\Theta_t,c\bigr),
\qquad
y_m^{(t+1)}
=
\mathcal{A}_t\bigl(y_m^{(t)};\Theta_t,c\bigr).
\label{eq:analytic-compositional-update}
\end{equation}
Hence, the overall registration process can be viewed as a compositional sequence of analytic updates,
\[
Y^{(0)}
\;\xrightarrow{\mathcal{A}_0}\;
Y^{(1)}
\;\xrightarrow{\mathcal{A}_1}\;
\cdots
\;\xrightarrow{\mathcal{A}_{t}}\;
Y^{(t+1)}
\;\cdots.
\]

The correspondences between \(X\) and \(Y^{(t)}\) are unknown. We therefore retain the probabilistic posterior correspondence layer of CPD. At each iteration, posterior probabilities are first estimated from the current moving configuration. These probabilities are then condensed into weighted soft target
points, and the M-step estimates \(\mathcal{A}_t\) by fitting the structured analytic mapping model described in Section~\ref{subsec:structured-analytic-mapping-model}.
The proposed method therefore alternates between CPD-style posterior correspondence estimation and structured analytic mapping estimation.

\subsection{CPD Posterior Layer}
\label{subsec:cpd-posterior-layer}

We retain the probabilistic correspondence layer of coherent point
drift~\cite{myronenko2010point}. At iteration \(t\), the current moving point set
\[
Y^{(t)}=\{y_m^{(t)}\}_{m=1}^{M}
\]
is regarded as the set of Gaussian mixture centroids, while the fixed point set
\[
X=\{x_n\}_{n=1}^{N}
\]
is treated as the set of observed data points. Each centroid \(y_m^{(t)}\) defines an isotropic Gaussian component with covariance \(\sigma_t^2 I\). As in CPD, a uniform outlier component is included to model noise and outliers.

The mixture likelihood of a fixed point \(x_n\) at iteration \(t\) is written as
\begin{equation}
p^{(t)}(x_n)
=
\frac{w}{N}
+
\frac{1-w}{M}
\sum_{m=1}^{M}
\frac{1}{(2\pi\sigma_t^2)^{d/2}}
\exp\left(
-\frac{\|x_n-y_m^{(t)}\|^2}{2\sigma_t^2}
\right),
\label{eq:mixture-density}
\end{equation}
where \(w\in[0,1)\) is the outlier weight.

Equivalently, after removing the common constants in the posterior
normalization, define the Gaussian affinity
\[
K_{mn}^{(t)}
=
\exp\left(
-\frac{\|x_n-y_m^{(t)}\|^2}{2\sigma_t^2}
\right).
\]
The posterior probability that \(x_n\) is generated by the \(m\)-th Gaussian centroid is then
\begin{equation}
P_{mn}^{(t)}
=
P(m\mid x_n)
=
\frac{
K_{mn}^{(t)}
}{
\sum_{k=1}^{M}K_{kn}^{(t)}+c_t
},
\label{eq:posterior}
\end{equation}
where
\begin{equation}
c_t
=
(2\pi\sigma_t^2)^{d/2}
\frac{w}{1-w}
\frac{M}{N}.
\label{eq:outlier-constant}
\end{equation}
The posterior matrix is denoted by
\[
P^{(t)}=\left(P_{mn}^{(t)}\right)\in\mathbb{R}^{M\times N}.
\]

When \(w=0\), each column of \(P^{(t)}\) sums to one. When \(w>0\), part of the probability mass is assigned to the uniform outlier component, and therefore
\[
\sum_{m=1}^{M}P_{mn}^{(t)}\le 1.
\]
The effective posterior mass assigned to the Gaussian centroids is
\[
N_P^{(t)}
=
\sum_{m=1}^{M}\sum_{n=1}^{N}P_{mn}^{(t)}.
\]

For later use, we define the row and column posterior sums as
\begin{equation}
\rho^{(t)}=P^{(t)}\mathbf 1_N\in\mathbb{R}^{M},
\qquad
\eta^{(t)}=(P^{(t)})^\top\mathbf 1_M\in\mathbb{R}^{N}.
\label{eq:posterior-row-column-sums}
\end{equation}
Thus,
\[
\rho_m^{(t)}
=
\sum_{n=1}^{N}P_{mn}^{(t)}
\]
measures the total posterior mass received by the current moving centroid \(y_m^{(t)}\), whereas
\[
\eta_n^{(t)}
=
\sum_{m=1}^{M}P_{mn}^{(t)}
\]
measures how much of the fixed point \(x_n\) is explained by the Gaussian components rather than by the outlier distribution. These quantities satisfy
\[
N_P^{(t)}
=
\mathbf 1_M^\top \rho^{(t)}
=
\mathbf 1_N^\top \eta^{(t)}.
\]

This posterior layer provides a soft correspondence structure between the fixed point set and the current moving point set. The proposed method differs from the original non-rigid CPD not in this correspondence estimation step, but in how the subsequent M-step uses the posterior information. Instead of estimating
a Gaussian-kernel displacement field, we convert the posterior probabilities into weighted soft target points and fit a structured analytic mapping.

\subsection{Barycentric Condensation of Soft Correspondences}
\label{subsec:barycentric-condensation}

The posterior matrix \(P^{(t)}\) encodes soft correspondences between all pairs of current moving points and fixed points. Up to the positive factor \(1/(2\sigma_t^2)\) and terms independent of \(\Theta\), the data term of the CPD M-step can be written as
\begin{equation}
\mathcal{Q}_{\mathrm{pair}}(\Theta)
=
\sum_{m=1}^{M}\sum_{n=1}^{N}
P_{mn}^{(t)}
\left\|
x_n-\mathcal{A}_t(y_m^{(t)};\Theta,c)
\right\|^2 .
\label{eq:pairwise-soft-objective}
\end{equation}
Although this expression is conceptually simple, it is not the most convenient form for structured analytic mapping estimation. Under the squared Euclidean loss, the soft correspondence information can be condensed exactly into a set of weighted soft target points. This condensation changes the form of the
objective but preserves its minimizer with respect to the analytic mapping parameters.

Recall that the posterior mass associated with the current moving point
\(y_m^{(t)}\) is
\[
\rho_m^{(t)}
=
\sum_{n=1}^{N}P_{mn}^{(t)} .
\]
For rows with positive posterior mass, we define the corresponding soft target point as the posterior barycenter of the fixed point set:
\begin{equation}
z_m^{(t)}
=
\frac{1}{\rho_m^{(t)}}
\sum_{n=1}^{N}
P_{mn}^{(t)}x_n .
\label{eq:soft-target}
\end{equation}
Thus, \(z_m^{(t)}\) is the responsibility-weighted average of all fixed points explained by the current moving centroid \(y_m^{(t)}\).

Identifying \(X\) with the coordinate matrix whose rows are \(x_n^\top\),
\[
X=
\begin{bmatrix}
x_1^\top\\
\vdots\\
x_N^\top
\end{bmatrix}
\in\mathbb{R}^{N\times d},
\]
the soft target matrix can be written as
\begin{equation}
Z^{(t)}
=
\operatorname{Diag}\!\left(\rho^{(t)}\right)^{-1}
P^{(t)}X ,
\label{eq:soft-target-matrix}
\end{equation}
where
\[
Z^{(t)}
=
\begin{bmatrix}
(z_1^{(t)})^\top\\
\vdots\\
(z_M^{(t)})^\top
\end{bmatrix}
\in\mathbb{R}^{M\times d}.
\]
Equation~\eqref{eq:soft-target-matrix} is understood only on rows with positive posterior mass. In the implementation, rows with zero or negligible posterior mass are ignored for numerical stability.

The key algebraic step is the weighted barycentric decomposition. For any \(u_m\in\mathbb R^d\), one has
\begin{equation}
\sum_{n=1}^{N}
P_{mn}^{(t)}
\|x_n-u_m\|^2
=
\rho_m^{(t)}
\|z_m^{(t)}-u_m\|^2
+
\sum_{n=1}^{N}
P_{mn}^{(t)}
\|x_n-z_m^{(t)}\|^2 .
\label{eq:barycentric-decomposition}
\end{equation}
This identity follows from the defining property of the weighted barycenter,
\[
\sum_{n=1}^{N}P_{mn}^{(t)}(x_n-z_m^{(t)})=0 .
\]
Taking
\[
u_m=\mathcal{A}_t(y_m^{(t)};\Theta,c),
\]
the second term on the right-hand side of
\eqref{eq:barycentric-decomposition} is independent of \(\Theta\). Therefore, minimizing the pairwise soft objective in
\eqref{eq:pairwise-soft-objective} with respect to the analytic mapping parameters is equivalent to minimizing
\begin{equation}
\Theta_t^\star
=
\arg\min_{\Theta}
\sum_{m\in\mathcal I_t}
\rho_m^{(t)}
\left\|
z_m^{(t)}
-
\mathcal{A}_t(y_m^{(t)};\Theta,c)
\right\|^2 ,
\label{eq:weighted-soft-target-objective}
\end{equation}
where \(\mathcal I_t=\{m:\rho_m^{(t)}>0\}\) denotes the set of active moving centroids. In numerical implementation, this set is formed using a small positive threshold on \(\rho_m^{(t)}\).

Equation~\eqref{eq:weighted-soft-target-objective} converts the CPD posterior information into a weighted point-set fitting problem. The current moving points \(y_m^{(t)}\) serve as the source points, the barycenters \(z_m^{(t)}\) serve as the soft target points, and the posterior masses \(\rho_m^{(t)}\) serve as fitting weights. This condensation is exact for the squared Euclidean loss; it is not an approximation of the original pairwise objective. Its role is to expose the M-step as a weighted structured analytic mapping estimation problem.

Thus, the departure from non-rigid CPD occurs after the posterior computation: the same posterior statistics are used to form weighted soft targets rather than to estimate a Gaussian-kernel displacement field.

\subsection{Structured Analytic Mapping Model}
\label{subsec:structured-analytic-mapping-model}

We use the structured analytic mapping framework introduced in~\cite{feng2026structured} as the deformation model in the M-step. In~\cite{feng2026structured}, vector-valued registration mappings were represented by finite-order multivariate Taylor expansions organized through generalized derivative matrices and generalized monomial vectors. Here we use an equivalent compact multi-index and matrix-vector form. The purpose of this subsection is to specify how this finite-dimensional
analytic mapping family is instantiated as the M-step deformation model in Analytic-CPD.

At iteration \(t\), the mapping to be estimated is
\[
\mathcal{A}_t:\mathbb{R}^{d}\rightarrow\mathbb{R}^{d},
\qquad
y\mapsto \mathcal{A}_t(y;\Theta_t,c),
\]
where \(c\in\mathbb{R}^{d}\) is the expansion center and \(\Theta_t\) denotes the Taylor coefficients. In the present implementation, \(c\) is fixed and typically chosen as the origin after data normalization. The coefficients \(\Theta_t\) are updated at every iteration.

Let \(q_t\) denote the truncation order at iteration \(t\), i.e., the maximum total degree of the Taylor terms included in the analytic update. The coefficient set is
\[
\Theta_t
=
\left\{
a_{\alpha}^{(t)}\in\mathbb{R}^{d}
:\ |\alpha|\le q_t
\right\},
\]
where \(\alpha=(\alpha_1,\ldots,\alpha_d)\) is a multi-index. The structured analytic mapping is represented as
\begin{equation}
\mathcal{A}_t(y;\Theta_t,c)
=
\sum_{r=0}^{q_t}
\sum_{|\alpha|=r}
\frac{(y-c)^{\alpha}}{\alpha!}\,
a_{\alpha}^{(t)},
\qquad
a_{\alpha}^{(t)}\in\mathbb{R}^{d}.
\label{eq:sam-general-expansion}
\end{equation}
Here
\[
\begin{aligned}
|\alpha| &= \alpha_1+\cdots+\alpha_d,
\qquad
\alpha! = \alpha_1!\cdots\alpha_d!,\\
(y-c)^\alpha
&=
\prod_{\ell=1}^{d}(y_\ell-c_\ell)^{\alpha_\ell}.
\end{aligned}
\]
In a classical Taylor expansion, \(a_\alpha^{(t)}\) corresponds to a derivative vector at the expansion center. In registration, the underlying deformation is unknown; these derivative-like coefficient vectors are therefore treated as parameters to be estimated from point-set data.

The number of monomials of exact total degree \(r\) in \(d\) variables is
\begin{equation}
C_{d,r}
=
\binom{r+d-1}{d-1}.
\label{eq:monomial-number}
\end{equation}
Thus, the number of Taylor basis terms of total degree no larger than \(q_t\) is
\begin{equation}
S_{d,q_t}
=
\sum_{r=0}^{q_t} C_{d,r}
=
\binom{q_t+d}{d}.
\label{eq:total-monomial-number}
\end{equation}
Since each basis term has a \(d\)-dimensional vector coefficient, the number of scalar degrees of freedom is
\begin{equation}
K_{d,q_t}
=
dS_{d,q_t}
=
d\binom{q_t+d}{d}.
\label{eq:parameter-number}
\end{equation}
This parameter count depends only on the spatial dimension \(d\) and the truncation order \(q_t\), not on the number of moving points. This point-independent parameterization is a fundamental distinction from non-rigid CPD, whose displacement field is tied to a point-indexed Gaussian kernel matrix.

To expose the linear structure in the unknown coefficients, we collect all Taylor basis terms into a feature vector
\[
\phi_{q_t}(y;c)\in\mathbb{R}^{S_{d,q_t}},
\]
and arrange the corresponding vector coefficients into a matrix
\[
A^{(t)}\in\mathbb{R}^{d\times S_{d,q_t}}.
\]
The mapping can then be written compactly as
\begin{equation}
\mathcal{A}_t(y;\Theta_t,c)
=
A^{(t)}\phi_{q_t}(y;c).
\label{eq:sam-compact-form}
\end{equation}
The mapping is nonlinear in the input coordinate \(y\), but linear in the unknown coefficient matrix \(A^{(t)}\). This linear-in-parameters structure is the key reason why the Analytic-CPD M-step reduces to a weighted least-squares problem.

A simple two-dimensional second-order example illustrates the basis. Let
\[
\delta_1=y_1-c_1,
\qquad
\delta_2=y_2-c_2.
\]
Then
\begin{equation}
\phi_2(y;c)
=
\left[
1,\;
\delta_1,\;
\delta_2,\;
\frac{1}{2}\delta_1^2,\;
\delta_1\delta_2,\;
\frac{1}{2}\delta_2^2
\right]^\top .
\label{eq:sam-2d-second-order-basis}
\end{equation}
In higher dimensions and orders, the basis is constructed analogously by enumerating all multi-indices \(\alpha\) with \(|\alpha|\le q_t\). The ordering of monomials is arbitrary, provided that it is used consistently in basis evaluation, matrix assembly, and mapping evaluation.

The first-order case recovers affine mappings. When \(q_t=1\),
\begin{equation}
\mathcal{A}_t(y;\Theta_t,c)
=
a_{\mathbf{0}}^{(t)}
+
\sum_{\ell=1}^{d}
a_{e_\ell}^{(t)}(y_\ell-c_\ell),
\label{eq:sam-first-order}
\end{equation}
where \(e_\ell\) denotes the \(\ell\)-th canonical multi-index. Equivalently,
\begin{equation}
\mathcal{A}_t(y)=b_t+B_t y,
\label{eq:sam-affine-form}
\end{equation}
with
\[
B_t=
\begin{bmatrix}
a_{e_1}^{(t)} & a_{e_2}^{(t)} & \cdots & a_{e_d}^{(t)}
\end{bmatrix},
\qquad
b_t=a_{\mathbf 0}^{(t)}-B_t c.
\]
Hence general affine mappings are contained as first-order instances of the structured analytic mapping family. Rigid transformations can be obtained by imposing the usual orthogonality constraints on the linear part. Higher-order terms provide nonlinear analytic deformation modes.

With this representation, the weighted soft-target objective derived in Section~\ref{subsec:barycentric-condensation} becomes a linear weighted least-squares problem in the coefficient matrix \(A^{(t)}\), as detailed next.

\subsection{Weighted Analytic Fitting for the CPD M-step}
\label{subsec:weighted-analytic-adjustment}

We now describe how the structured analytic mapping coefficients are estimated from the weighted soft targets produced by the CPD posterior layer. By the barycentric condensation in Section~\ref{subsec:barycentric-condensation}, the pairwise soft-correspondence objective is reduced, up to a constant independent
of the mapping, to the weighted fitting problem in \eqref{eq:weighted-soft-target-objective}. Therefore, the Analytic-CPD M-step can be viewed as a weighted structured analytic fitting problem between the current moving points and their posterior barycentric targets.

For general notation, let
\[
\widehat{Y}=\{\widehat y_m\}_{m=1}^{M_s},
\qquad
\widehat{Z}=\{\widehat z_m\}_{m=1}^{M_s},
\qquad
\omega_m\ge 0 .
\]
The weighted structured analytic fitting problem is
\begin{equation}
\Theta^\star
=
\arg\min_{\Theta}
\sum_{m=1}^{M_s}
\omega_m
\left\|
\widehat z_m-\mathcal A(\widehat y_m;\Theta,c)
\right\|^2 .
\label{eq:weighted-sam-fitting}
\end{equation}
Since \(\mathcal A\) is linear in its coefficients, this is a weighted linear least-squares problem.

Using the compact representation
\[
\mathcal A(y;\Theta,c)=A\phi_q(y;c),
\]
define the Taylor design matrix
\[
\Phi
=
\begin{bmatrix}
\phi_q(\widehat y_1;c)^\top\\
\vdots\\
\phi_q(\widehat y_{M_s};c)^\top
\end{bmatrix}
\in\mathbb R^{M_s\times S_{d,q}},
\]
and let \(\widehat Z\in\mathbb R^{M_s\times d}\) denote the matrix whose \(m\)-th row is \(\widehat z_m^\top\). With
\[
\Omega=\operatorname{Diag}(\omega_1,\ldots,\omega_{M_s}),
\]
the fitting problem becomes
\begin{equation}
A^\star
=
\arg\min_A
\left\|
\Omega^{1/2}
\left(
\widehat Z-\Phi A^\top
\right)
\right\|_F^2 .
\label{eq:weighted-sam-matrix-form}
\end{equation}
The corresponding normal equation is
\begin{equation}
\left(\Phi^\top\Omega\Phi\right)(A^\star)^\top
=
\Phi^\top\Omega\widehat Z .
\label{eq:weighted-sam-normal-equation}
\end{equation}
Thus, all output coordinates share the same weighted Taylor design matrix, and the size of the analytic fitting system is governed by \(S_{d,q}\), rather than by the number of moving points. Further algebraic details, including the vectorized adjustment form and the correction formulation, are provided in Appendix~C.

In the Analytic-CPD iteration, \eqref{eq:weighted-sam-fitting} is applied to the active index set
\[
\mathcal I_t
=
\{\,m:\rho_m^{(t)}>\varepsilon_\rho\,\},
\]
with source points \(y_m^{(t)}\), soft targets \(z_m^{(t)}\), and weights \(\rho_m^{(t)}\). The M-step therefore solves
\begin{equation}
\Theta_t^\star
=
\arg\min_{\Theta_t}
\sum_{m\in\mathcal I_t}
\rho_m^{(t)}
\left\|
z_m^{(t)}
-
\mathcal A_t(y_m^{(t)};\Theta_t,c)
\right\|^2 .
\label{eq:analytic-cpd-mstep-objective}
\end{equation}
The fitted analytic map is then applied to all current moving points:
\begin{equation}
y_m^{(t+1)}
=
\mathcal A_t(y_m^{(t)};\Theta_t^\star,c),
\qquad
m=1,\ldots,M .
\label{eq:analytic-cpd-mapping-update}
\end{equation}

In practice, the weighted least-squares problem can be solved by stable least-squares solvers such as QR or SVD. The normal equation form is used only to expose the algebraic structure of the weighted analytic M-step.

\subsection{Compositional Update and Variance Estimation}
\label{subsec:compositional-update-variance}

After solving the Analytic-CPD M-step in
\eqref{eq:analytic-cpd-mstep-objective}, the estimated coefficient set \(\Theta_t^\star\) defines an analytic map
\[
\mathcal{A}_t(\cdot;\Theta_t^\star,c):\mathbb{R}^d\rightarrow\mathbb{R}^d .
\]
The current moving point set is updated by applying this map to every point:
\begin{equation}
Y^{(t+1)}
=
\mathcal{A}_t(Y^{(t)};\Theta_t^\star,c),
\qquad
y_m^{(t+1)}
=
\mathcal{A}_t(y_m^{(t)};\Theta_t^\star,c).
\label{eq:compositional-point-update}
\end{equation}
Thus, each M-step estimates an analytic update on the current moving configuration, rather than a single point-indexed kernel displacement field. The accumulated deformation from the original moving configuration to the current registered configuration is represented implicitly by the composition
\[
\mathcal{A}_t\circ\mathcal{A}_{t-1}\circ\cdots\circ\mathcal{A}_0 .
\]

This compositional form is important for the expressive power of the method. According to the order-amplification property of composed structured analytic mappings~\cite{feng2026structured}, if the successive analytic updates have orders \(q_0,q_1,\ldots,q_t\), then the accumulated mapping remains analytic,
and its effective polynomial order can grow multiplicatively, with the upper bound
\[
q_{\mathrm{eff}}
\le
\prod_{s=0}^{t} q_s .
\]
Therefore, Analytic-CPD does not rely on fitting the entire deformation by a single high-order Taylor map in one step. A sequence of low-order or staged-order analytic updates can generate highly expressive smooth deformations while keeping each individual M-step relatively stable.

Given the posterior matrix \(P^{(t)}\) computed in the E-step and the updated moving point set \(Y^{(t+1)}\), the variance parameter is updated in the same maximum-likelihood manner as in CPD. For fixed \(P^{(t)}\), the relevant part of the expected negative log-likelihood is
\begin{equation}
Q(\sigma^2)
=
\frac{1}{2\sigma^2}
\sum_{m=1}^{M}\sum_{n=1}^{N}
P_{mn}^{(t)}
\left\|
x_n-y_m^{(t+1)}
\right\|^2
+
\frac{N_P^{(t)}d}{2}\log\sigma^2 ,
\label{eq:sigma-objective}
\end{equation}
where
\[
N_P^{(t)}
=
\sum_{m=1}^{M}\sum_{n=1}^{N}P_{mn}^{(t)}
\]
is the effective posterior mass assigned to the Gaussian components. Taking the derivative of \eqref{eq:sigma-objective} with respect to \(\sigma^2\) and setting it to zero gives
\begin{equation}
\sigma_{t+1}^{2}
=
\frac{1}{N_P^{(t)}d}
\sum_{m=1}^{M}\sum_{n=1}^{N}
P_{mn}^{(t)}
\left\|
x_n-y_m^{(t+1)}
\right\|^2 .
\label{eq:sigma-update-sum}
\end{equation}

For efficient implementation, \eqref{eq:sigma-update-sum} can be written using the posterior statistics already computed in the E-step. Using the posterior row and column sums defined in \eqref{eq:posterior-row-column-sums}, and identifying \(X\) and \(Y^{(t+1)}\) with their coordinate matrices, define
\[
S_X^{(t)}
=
P^{(t)}X
\in\mathbb{R}^{M\times d}.
\]
Then
\begin{equation}
\begin{aligned}
\sigma_{t+1}^{2}
=
\frac{1}{N_P^{(t)}d}
\Big[
&
\operatorname{tr}
\left(
X^\top
\operatorname{Diag}(\eta^{(t)})
X
\right)
-
2\operatorname{tr}
\left(
(S_X^{(t)})^\top
Y^{(t+1)}
\right)
\\
&
+
\operatorname{tr}
\left(
(Y^{(t+1)})^\top
\operatorname{Diag}(\rho^{(t)})
Y^{(t+1)}
\right)
\Big].
\end{aligned}
\label{eq:sigma-update-matrix}
\end{equation}
This expression avoids explicitly summing all pairwise residuals after the M-step.

The quantity
\begin{equation}
E_{\mathrm{soft}}^{(t+1)}
=
\sqrt{d\,\sigma_{t+1}^{2}}
\label{eq:soft-rmse}
\end{equation}
can be interpreted as a CPD-style posterior-weighted RMS residual. It is useful as an internal convergence indicator because it is consistent with the probabilistic objective optimized by the algorithm. When ground-truth pointwise
correspondences are available, an external pointwise error can also be computed from the final registered point set for comparison across different registration methods.

The resulting iteration is an EM-style procedure. The E-step computes the posterior matrix \(P^{(t)}\) from the current moving configuration \(Y^{(t)}\). The M-step first converts \(P^{(t)}\) into weighted soft targets, estimates the structured analytic mapping coefficients by solving \eqref{eq:analytic-cpd-mstep-objective}, updates the moving point set by
\eqref{eq:compositional-point-update}, and finally updates the variance by \eqref{eq:sigma-update-sum} or \eqref{eq:sigma-update-matrix}. Since the deformation update is estimated within a prescribed finite-dimensional analytic mapping family and applied compositionally to the current moving configuration, the procedure can be regarded as a generalized CPD iteration with a structured analytic M-step.

\subsection{Degree-Continuation Strategy}
\label{subsec:degree-continuation}

The truncation order \(q_t\) controls the expressive power of the structured analytic mapping used at iteration \(t\). A larger order provides richer nonlinear deformation modes, but it also increases the number of coefficients and may amplify unreliable posterior correspondences in the early stage of registration. We therefore use a degree-continuation strategy: low-order mappings are kept longer to stabilize the global alignment and posterior
estimation, while higher-order mappings are activated later for nonlinear refinement.

Let \(q_{\max}=D\) be the maximum analytic order. Instead of using a uniform ramp, we adopt an increasing-degree schedule with decreasing stage lengths:
\[
\text{degree:}\quad 1,\ 2,\ \ldots,\ D,
\qquad
\text{stage length:}\quad D,\ D-1,\ \ldots,\ 1 .
\]
Thus, lower-order analytic spaces receive longer optimization intervals, and higher-order spaces are used mainly in the later refinement stage. For a general maximum iteration number \(T_{\max}\), we distribute the total budget according to these triangular weights. Let \(L_q\) denote the number of
iterations assigned to degree \(q\), with
\[
L_1\ge L_2\ge \cdots \ge L_D,
\qquad
\sum_{q=1}^{D}L_q=T_{\max}.
\]
The raw analytic order at iteration \(t\) is then defined by
\begin{equation}
q_t^{\mathrm{raw}}
=
\min
\left\{
q:\;
t < \sum_{r=1}^{q} L_r
\right\},
\qquad
t=0,1,\ldots,T_{\max}-1 .
\label{eq:degree-decreasing-stage-schedule}
\end{equation}
For example, when \(D=10\) and \(T_{\max}=55\), the stage lengths are exactly
\[
10,\ 9,\ 8,\ 7,\ 6,\ 5,\ 4,\ 3,\ 2,\ 1 .
\]
This schedule reflects the fact that high-order Taylor terms are more sensitive to correspondence uncertainty and numerical conditioning, and should therefore be introduced only after the low-order geometry has become sufficiently stable.

In implementation, the selected order is also constrained by a feasibility condition. Let \(M_s\) denote the number of soft target pairs retained in the weighted fitting system after removing rows with zero or numerically negligible posterior mass. Since all output coordinates share the same Taylor design
matrix, a necessary condition for avoiding an underdetermined fitting problem is
\begin{equation}
M_s
\ge
S_{d,q_t}
=
\binom{q_t+d}{d}.
\label{eq:degree-feasibility}
\end{equation}
Equivalently, this condition can be written as
\[
dM_s\ge K_{d,q_t}.
\]
If this condition is violated, the order is reduced from
\(q_t^{\mathrm{raw}}\) until the system is not underdetermined. This feasibility check does not guarantee good conditioning, but it prevents rank deficiency caused solely by insufficient observations.

The strategy can be viewed as a numerical continuation over the nested Taylor
mapping spaces
\[
\mathcal{V}_{1}
\subset
\mathcal{V}_{2}
\subset
\cdots
\subset
\mathcal{V}_{q_{\max}},
\]
where \(\mathcal{V}_{q}\) denotes the finite-dimensional space of structured analytic mappings truncated at order \(q\). The admissible M-step deformation space is therefore expanded along the natural degree filtration of the Taylor model, rather than activating all high-order degrees of freedom from the beginning.

Together with the compositional update discussed in
Section~\ref{subsec:compositional-update-variance}, this schedule exploits the order-amplification property of SAM: high effective deformation complexity can be obtained through a sequence of stable low- or moderate-order analytic updates, rather than by directly activating a single ill-conditioned high-order model at the beginning.

\subsection{Algorithm Summary}
\label{subsec:algorithm-summary}

The proposed method follows the EM-style structure of CPD. The E-step computes the posterior correspondence matrix from the current moving configuration. The M-step condenses the posterior probabilities into weighted soft targets, estimates a structured analytic mapping, updates the moving point set compositionally, and re-estimates the variance. We refer to the resulting algorithm as \textbf{Analytic-CPD}. The complete procedure is summarized in Fig.~\ref{fig:analytic-cpd-algorithm}.

\begin{figure}[t]
\centering
\setlength{\fboxsep}{6pt}
\fbox{
\begin{minipage}{0.96\linewidth}
\footnotesize
\textbf{Analytic-CPD algorithm}

\vspace{1mm}

\textbf{Input:}
Fixed point set \(X=\{x_n\}_{n=1}^{N}\), moving point set
\(Y=\{y_m\}_{m=1}^{M}\), outlier weight \(w\), maximum analytic order
\(q_{\max}\), maximum iteration number \(T_{\max}\), and tolerance
\(\varepsilon\).

\vspace{1mm}

\textbf{Initialization:}
Set \(Y^{(0)}=Y\), initialize \(\sigma_0^2\) as in CPD, and set
\[
t=0,\qquad
Y_{\mathrm{best}}=Y^{(0)},\qquad
s_{\mathrm{best}}=\sqrt{d\sigma_0^2}.
\]

\vspace{1mm}

\textbf{Repeat until convergence or \(t=T_{\max}\):}

\begin{enumerate}
\setlength{\itemsep}{4pt}
\setlength{\parskip}{0pt}

\item \textbf{E-step: posterior estimation.}

Compute the CPD posterior matrix
\[
P^{(t)}=(p_{mn}^{(t)})\in\mathbb R^{M\times N}
\]
from \(X\), \(Y^{(t)}\), \(\sigma_t^2\), and \(w\). Compute posterior masses
and barycentric soft targets:
\[
\rho_m^{(t)}
=
\sum_{n=1}^{N}p_{mn}^{(t)},
\qquad
z_m^{(t)}
=
\frac{1}{\rho_m^{(t)}}
\sum_{n=1}^{N}p_{mn}^{(t)}x_n .
\]
Let
\[
\mathcal I_t=\{m:\rho_m^{(t)}>\varepsilon_\rho\}.
\]

\item \textbf{M-step: analytic fitting and moving-set update.}

Choose \(q_t\) by the degree-continuation schedule and reduce it, if necessary,
until
\[
|\mathcal I_t|
\ge
S_{d,q_t}
=
\binom{q_t+d}{d}.
\]
Estimate the structured analytic map:
\[
\Theta_t^\star
=
\arg\min_{\Theta_t}
\sum_{m\in\mathcal I_t}
\rho_m^{(t)}
\left\|
z_m^{(t)}
-
\mathcal A_t(y_m^{(t)};\Theta_t,c)
\right\|^2 .
\]
Update the moving point set and variance:
\[
\begin{aligned}
Y^{(t+1)}
&=
\mathcal A_t\!\left(Y^{(t)};\Theta_t^\star,c\right),\\
\sigma_{t+1}^2
&\leftarrow
\operatorname{VarUpdate}
\!\left(P^{(t)},X,Y^{(t+1)}\right).
\end{aligned}
\]

\item \textbf{Best-state and stopping rule.}

Set
\[
s_{t+1}=\sqrt{d\sigma_{t+1}^2}.
\]
If \(s_{t+1}<s_{\mathrm{best}}\), update
\[
Y_{\mathrm{best}}=Y^{(t+1)},
\qquad
s_{\mathrm{best}}=s_{t+1}.
\]
Stop if the residual, variance, or moving-set update has converged, or if the
internal residual clearly rebounds from \(s_{\mathrm{best}}\). Otherwise set
\(t\leftarrow t+1\).

\end{enumerate}

\vspace{1mm}

\textbf{Output:}
The best registered moving point set \(Y_{\mathrm{best}}\).

\end{minipage}
}
\caption{Summary of the proposed \textbf{Analytic-CPD} algorithm.}
\label{fig:analytic-cpd-algorithm}
\end{figure}

In the direct implementation, Analytic-CPD retains the \(\mathcal{O}(MN)\) posterior computation of CPD, but replaces the \(M\times M\) kernel-displacement M-step with a compact Taylor least-squares system whose size is governed by \(S_{d,q}=\binom{q+d}{d}\). A detailed complexity analysis is provided in
Appendix~A. The external RMSE rebound guard is used only for ordered synthetic benchmarks and is not used to estimate either the posterior matrix or the analytic mapping parameters.

\subsection{Implementation Details}
\label{subsec:implementation-details}

Before registration, all point sets are centered and scaled to a common normalized coordinate system, so that the coordinate magnitudes are of order one. This normalization is essential for the structured analytic mapping model, because the conditioning and relative contribution of Taylor basis terms depend directly on the coordinate scale. The Analytic-CPD iteration is performed in this normalized coordinate system, with the expansion center fixed at
\[
c=\mathbf 0 .
\]
Only the Taylor coefficient set \(\Theta_t\) is estimated in the M-step.

For numerical stability, rows with zero or negligible posterior mass are omitted from the weighted analytic fitting system. The selected analytic order is also checked against the feasibility condition in \eqref{eq:degree-feasibility} and is reduced when necessary to avoid an underdetermined least-squares problem.
The weighted analytic fitting can be solved using stable least-squares solvers such as QR or SVD; normal-equation forms are used only to expose the algebraic structure and may be used as efficient options in well-conditioned cases.

Unless otherwise specified, the reported implementation uses direct posterior computation and exact weighted analytic fitting. Fast Gaussian summation, low-rank approximation, and other acceleration schemes are not used in the main controlled comparison, so that the reported results isolate the model-level
difference between the CPD Gaussian-kernel M-step and the proposed structured analytic M-step. Further implementation details are provided in Appendix~B.

\section{Experiments}
\label{sec:experiments}

\subsection{Experimental Setup and Evaluation Metrics}
\label{subsec:experimental-setup}

We evaluate the proposed \textbf{Analytic-CPD} method on two-dimensional and three-dimensional point-set registration tasks. The experiments are designed to examine registration accuracy, computational efficiency, and robustness under increasing deformation complexity. Before registration, all point sets are
centered and scaled to a common unit-size normalized coordinate system. Unless otherwise specified, all reported errors are computed in this normalized coordinate system after applying the estimated mapping to the moving point set.

The experimental design follows a progressive validation strategy. The two-dimensional analytic deformation experiments serve as controlled model-matched tests with known pointwise correspondences. The three-dimensional smooth non-analytic deformation experiments provide the main model-mismatch
evaluation, because the ground-truth deformation is smooth but is not a single finite-order analytic map. The registered FAUST human-shape experiments provide additional non-synthetic validation on articulated non-rigid shape variations.
This organization separates model-matched verification, smooth model-mismatch evaluation, and registered real-shape validation.

We compare Analytic-CPD with TPS-RPM, CPD, Analytic-ICP, BCPD, and
ClusterReg~\cite{zhao2024clustereg}. TPS-RPM is included as a classical spline-based non-rigid registration method. CPD is the most directly related probabilistic coherent-motion baseline. Analytic-ICP uses the same structured analytic mapping model as Analytic-CPD but relies on ICP-style hard nearest-neighbor correspondences; it is therefore used to isolate the effect of
replacing hard correspondences with CPD-style posterior correspondences. BCPD and ClusterReg are evaluated using their official implementations.

For experiments with known pointwise correspondences, registration accuracy is evaluated by the same external pointwise RMSE,
\[
E_{\mathrm{rmse}}
=
\left(
\frac{1}{M}
\sum_{m=1}^{M}
\|x_m-y_m^{\mathrm{reg}}\|^2
\right)^{1/2}.
\]
Method-specific internal objective values, including the CPD-style residual \(E_{\mathrm{soft}}=\sqrt{d\sigma^2}\), are not used for cross-method accuracy comparison. When an external method uses its own normalization routine, its output is converted back to the common normalized coordinate system before
evaluation.

Unless otherwise specified, all runtime experiments are conducted on the same computer equipped with an AMD Ryzen 7 5800H CPU at 3.20\,GHz and 32\,GB RAM. Our in-house implementations, including CPD, TPS-RPM, Analytic-ICP, and Analytic-CPD, are 64-bit single-threaded C++ programs without fast Gauss transform, low-rank kernel approximation, multi-threaded BLAS, or GPU acceleration. Thus, their runtimes form a controlled in-house implementation
comparison. BCPD and ClusterReg are reported as official-implementation reference values, because their released packages may use different numerical libraries, memory layouts, parallelization strategies, and stopping rules.

For synthetic deformation experiments, we register the original point set to its deformed version,
\[
Y^{(0)}=X_0,
\qquad
X=X_{\mathrm{def}} .
\]
The reverse direction is not used because the inverse of a large nonlinear deformation may be ill-conditioned or poorly represented by a finite-order analytic mapping. The synthetic experiments include model-matched analytic deformations in 2D and smooth non-analytic deformations in 3D; the latter are used to evaluate the approximation capability of Analytic-CPD beyond exact
finite-order analytic ground-truth mappings.

The main parameters are fixed across experiments unless otherwise specified.
For Analytic-CPD, we use
\[
T_{\max}=55,\qquad q_{\max}=10,\qquad w=0.1 .
\]
The analytic order follows the increasing-degree continuation strategy with decreasing stage lengths described in Section~\ref{subsec:degree-continuation}.
The expansion center is fixed at the origin after normalization, and \(\sigma_0^2\) is initialized as the average pairwise squared distance, as in CPD. CPD, TPS-RPM, Analytic-ICP, BCPD, and ClusterReg are run with fixed parameter settings across the corresponding experiments. Detailed baseline settings, thread-control settings, and implementation notes are provided in Appendix~D.

\subsection{Two-Dimensional Analytic Deformation}
\label{subsec:2d-analytic-deformation}

We first evaluate Analytic-CPD on two-dimensional point sets generated by analytic deformation mappings. The purpose of this experiment is twofold. First, it verifies whether the proposed method can recover smooth nonlinear deformations generated from known analytic mappings. Second, it compares Analytic-CPD with TPS-RPM, CPD, BCPD, and Analytic-ICP under both small-deformation and large-deformation settings.

The two-dimensional experiments are organized into three parts. We first present representative large-deformation examples to visually demonstrate the registration capability of Analytic-CPD. We then report quantitative comparisons on small deformations, where hard nearest-neighbor correspondences are relatively reliable and Analytic-ICP is expected to perform strongly. Finally, we evaluate large-deformation cases, which are the main target regime of Analytic-CPD. In this regime, hard correspondences may become unreliable in the early iterations, while the CPD posterior layer provides smoother soft correspondence information for structured analytic mapping estimation.

\subsubsection{Large-deformation examples.}

We first show representative two-dimensional large-deformation examples to illustrate the qualitative behavior of Analytic-CPD. The examples are generated by applying large structured analytic deformations to the original point sets, following the analytic mapping model in~\cite{feng2026structured}. Red points denote the fixed point sets, and green points denote the moving or registered
moving point sets.

\begin{figure*}[t]
\centering

% ---------- legend ----------
\begingroup
\setlength{\tabcolsep}{1em}
\begin{tabular}{cc}
\textcolor{red}{\large$\bullet$}\ \ Red = fixed
&
\textcolor{green!70!black}{\large$\bullet$}\ \ Green = moving / moved
\end{tabular}
\endgroup

\vspace{0.8em}

% ---------- main table ----------
\begingroup
\setlength{\tabcolsep}{0.35em}
\renewcommand{\arraystretch}{1.15}

\begin{tabular}{
    >{\centering\arraybackslash}m{0.07\textwidth}
    >{\centering\arraybackslash}m{0.17\textwidth}
    >{\centering\arraybackslash}m{0.17\textwidth}
    >{\centering\arraybackslash}m{0.17\textwidth}
    >{\centering\arraybackslash}m{0.17\textwidth}
    >{\centering\arraybackslash}m{0.17\textwidth}
}
&
{\bfseries Initial}
&
{\bfseries Iter 5}
&
{\bfseries Iter 10}
&
{\bfseries Iter 15}
&
{\bfseries Final}
\\[0.5em]

% ---------- Example 1 ----------
{\bfseries (1)}
&
\includegraphics[width=\linewidth]{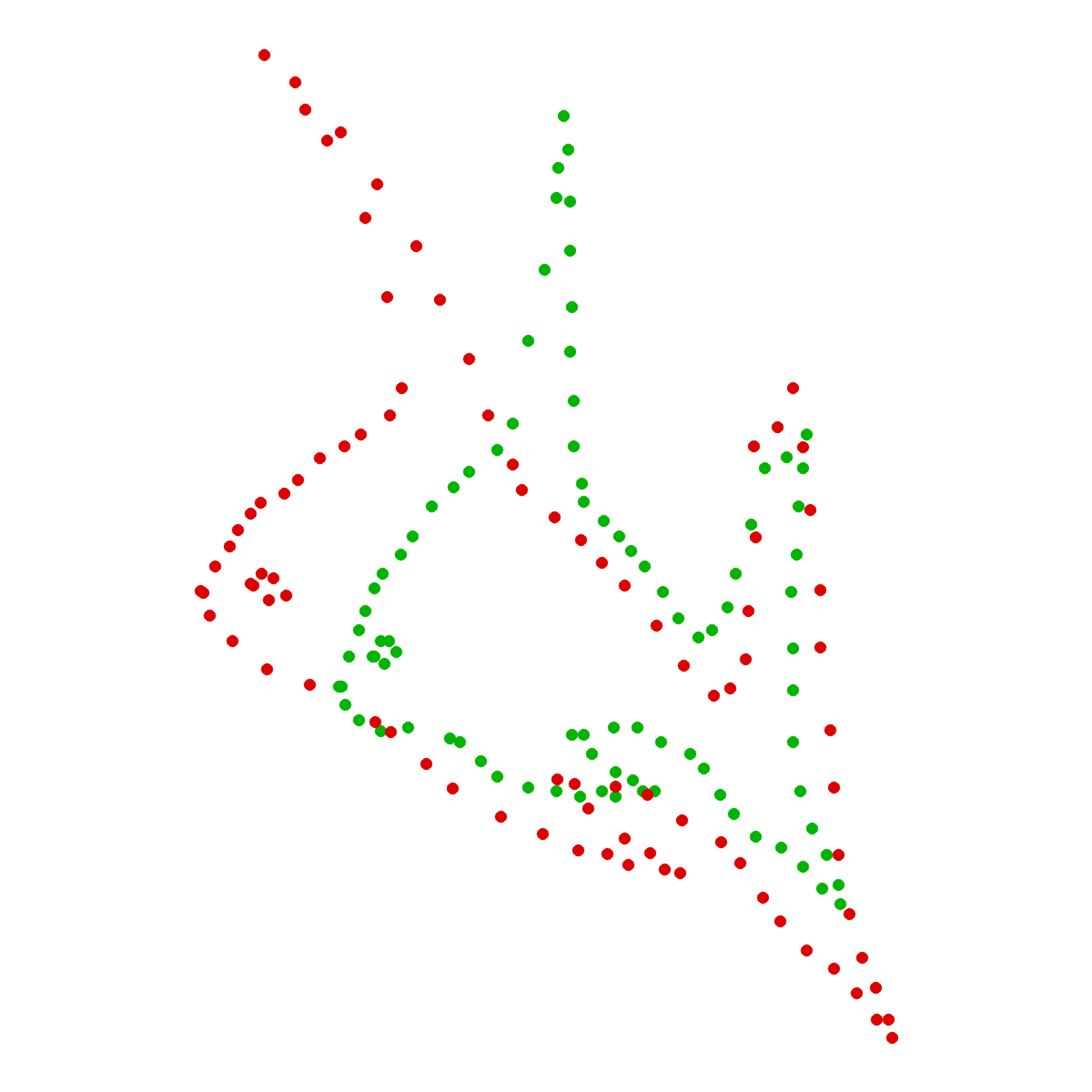}
&
\includegraphics[width=\linewidth]{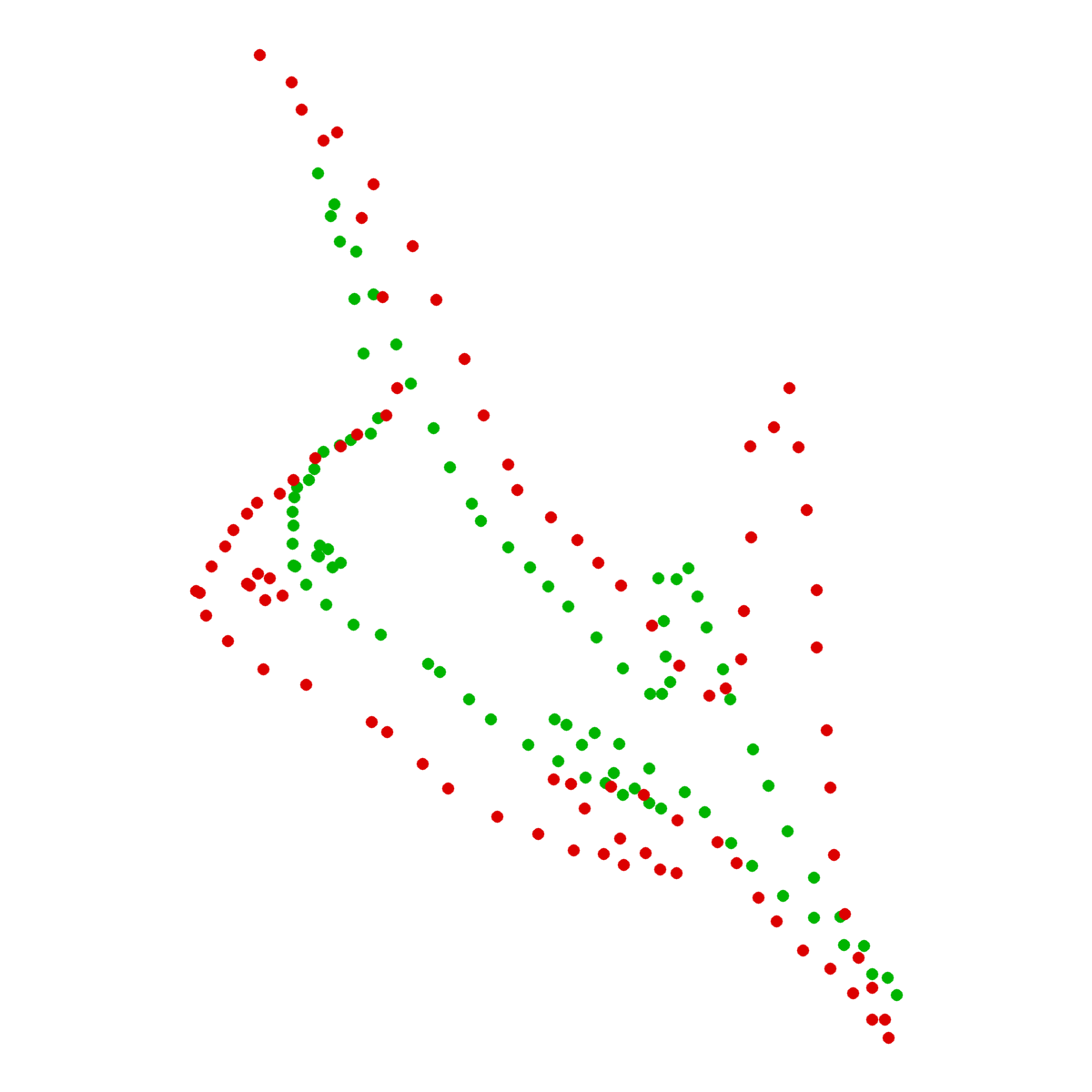}
&
\includegraphics[width=\linewidth]{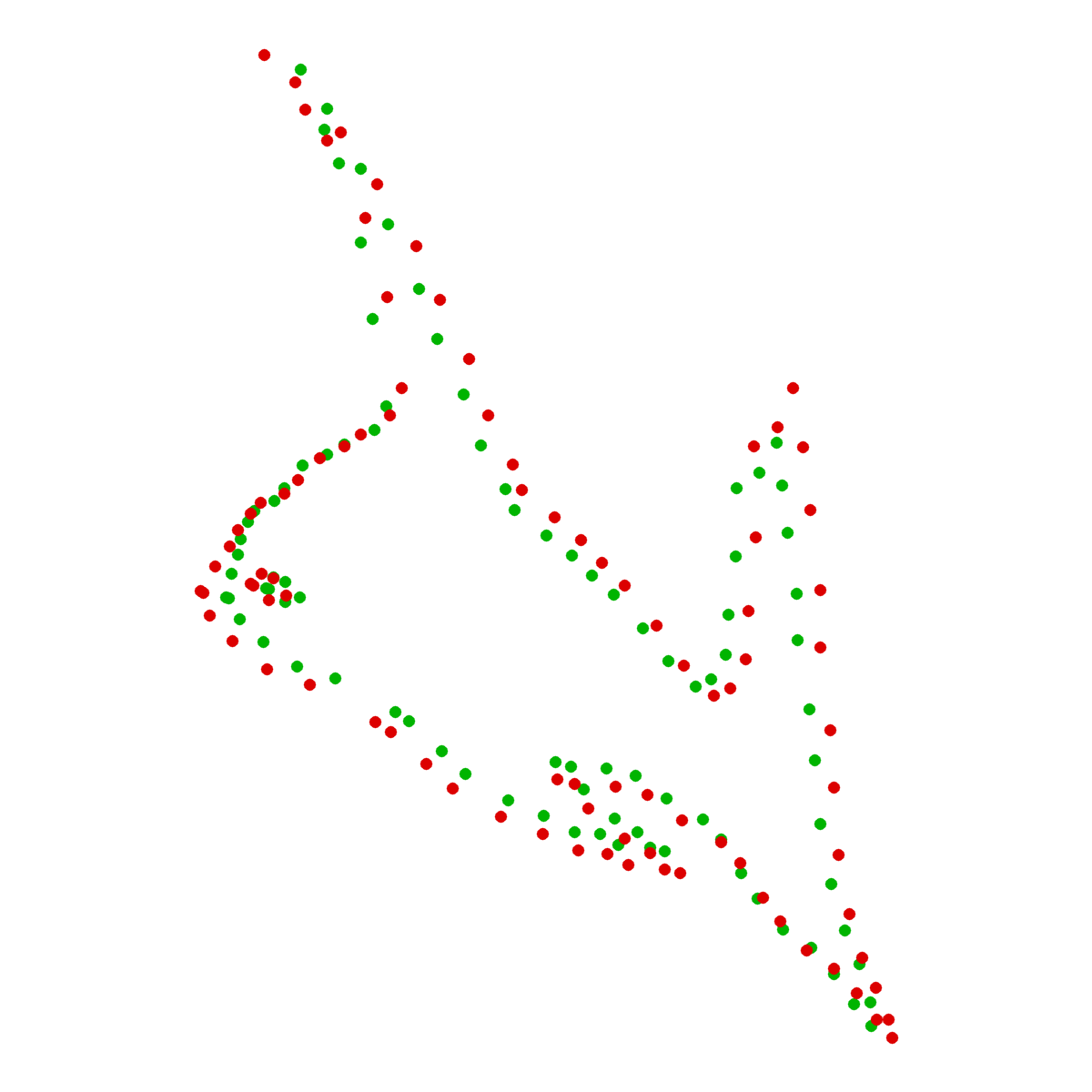}
&
\includegraphics[width=\linewidth]{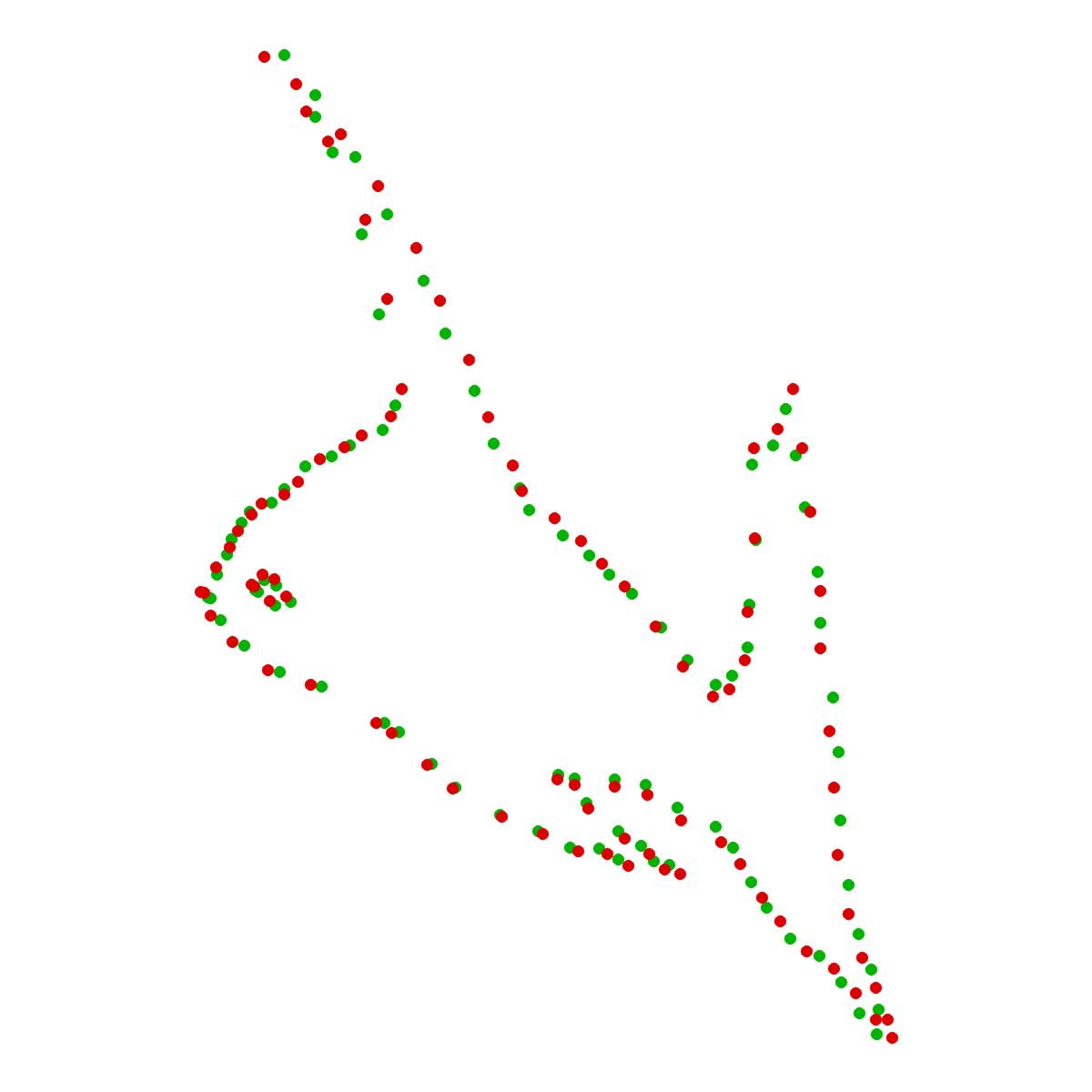}
&
\includegraphics[width=\linewidth]{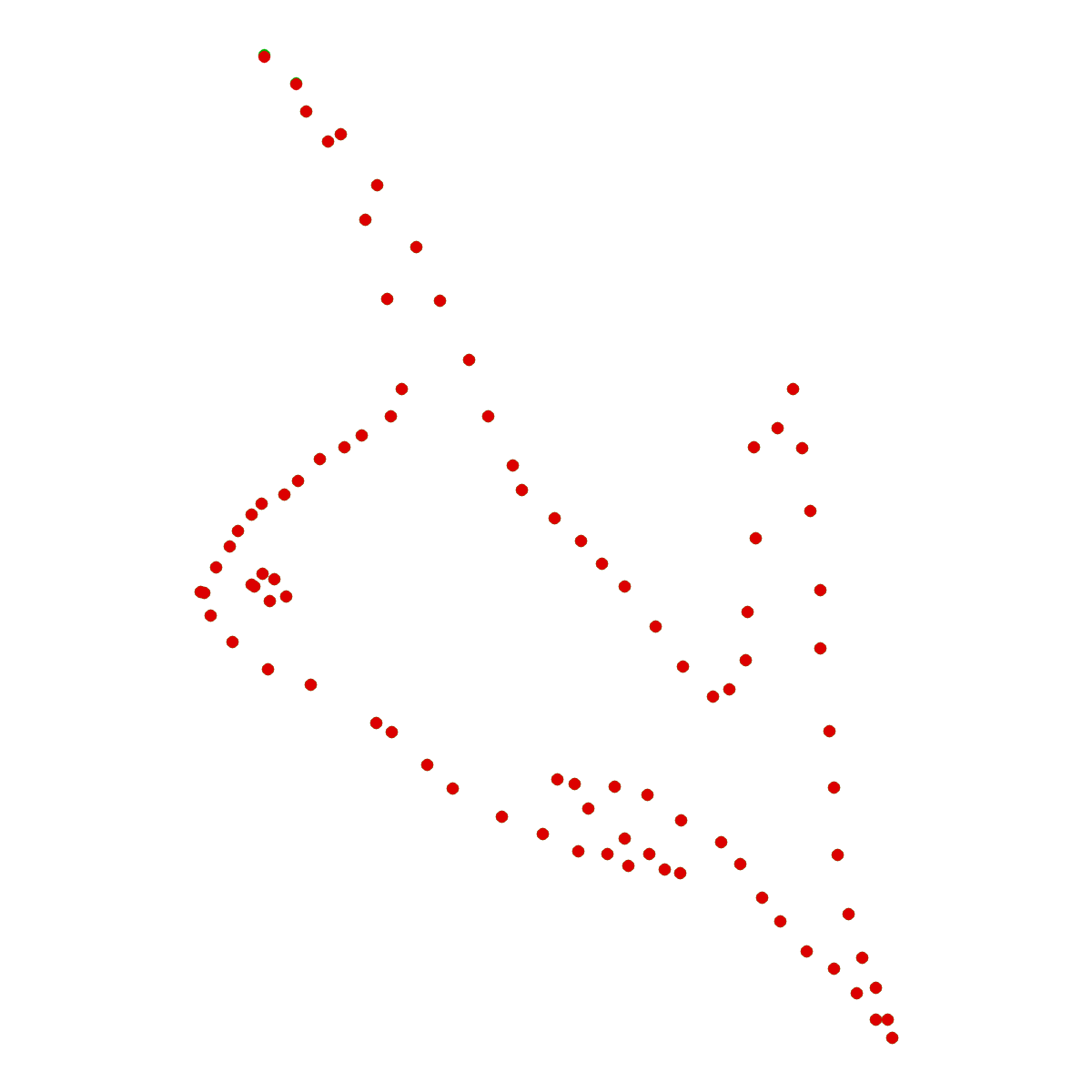}
\\[0.8em]

% ---------- Example 2 ----------
{\bfseries (2)}
&
\includegraphics[width=\linewidth]{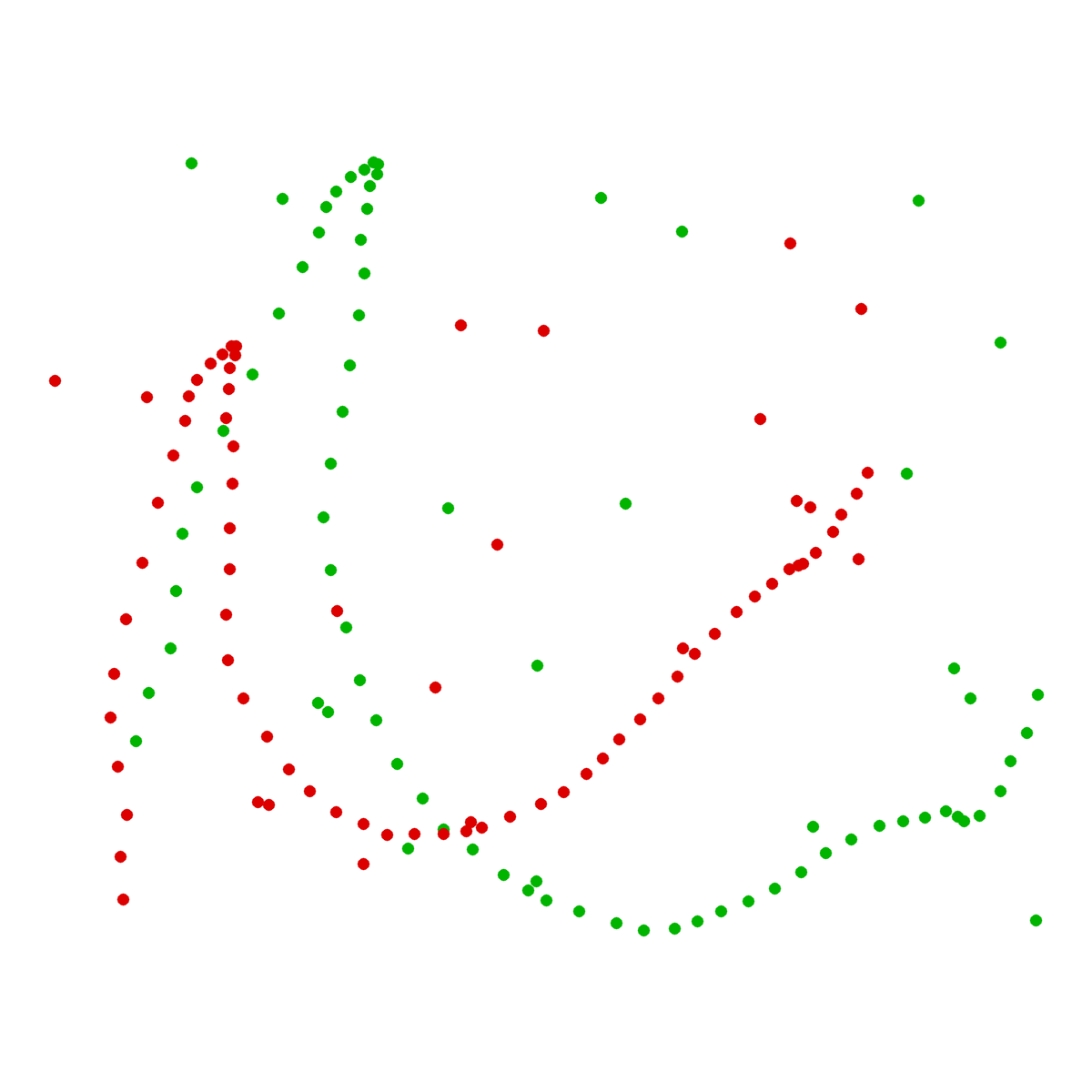}
&
\includegraphics[width=\linewidth]{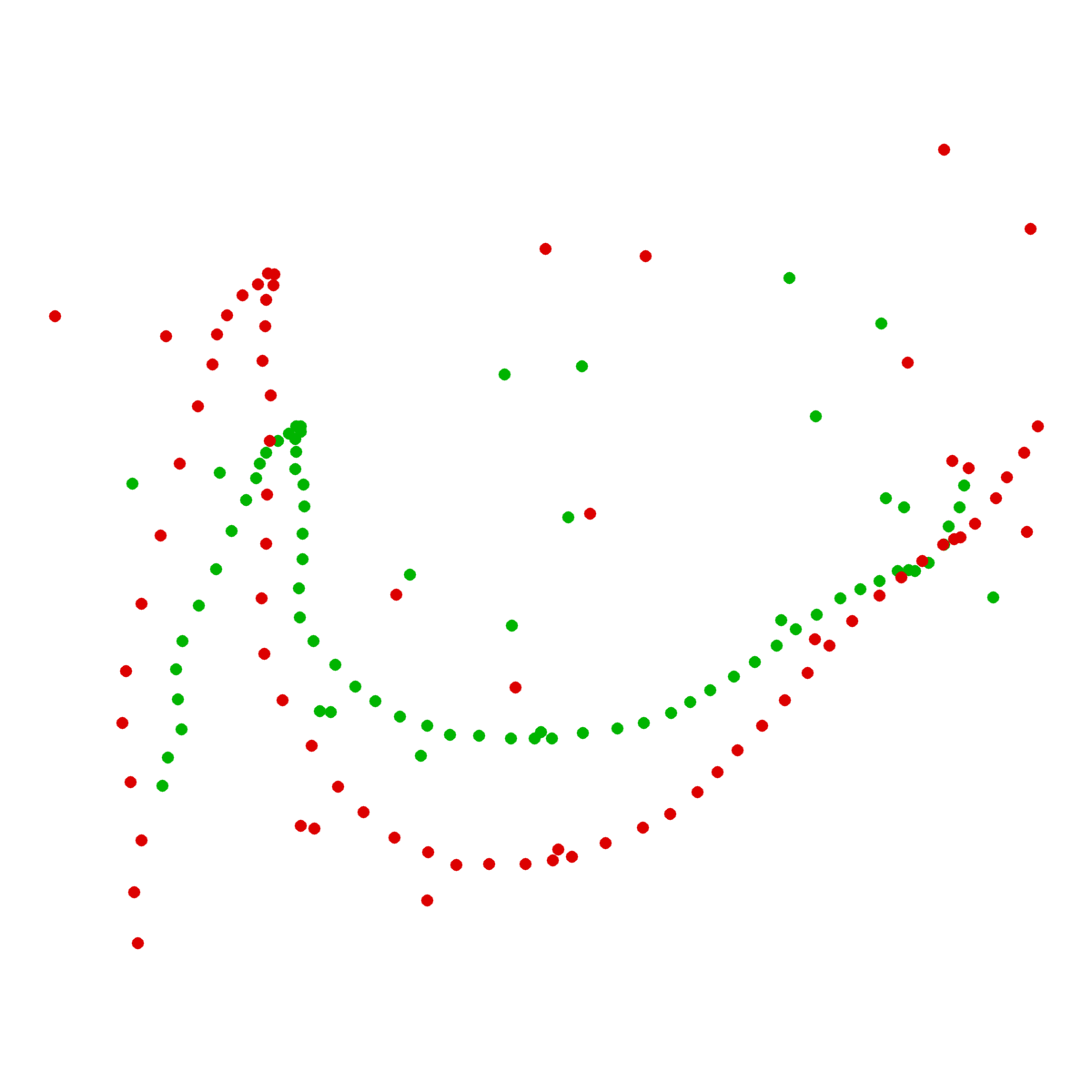}
&
\includegraphics[width=\linewidth]{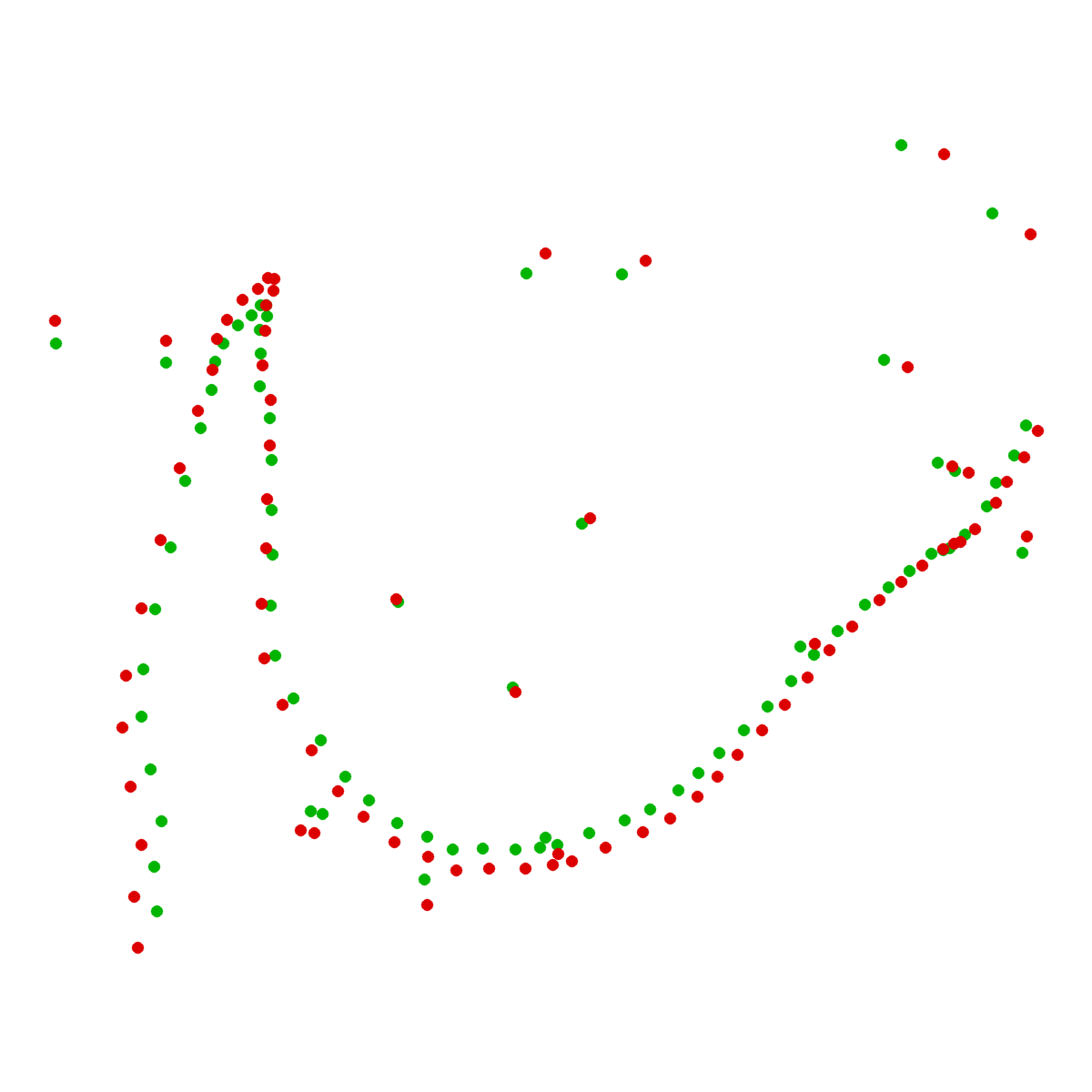}
&
\includegraphics[width=\linewidth]{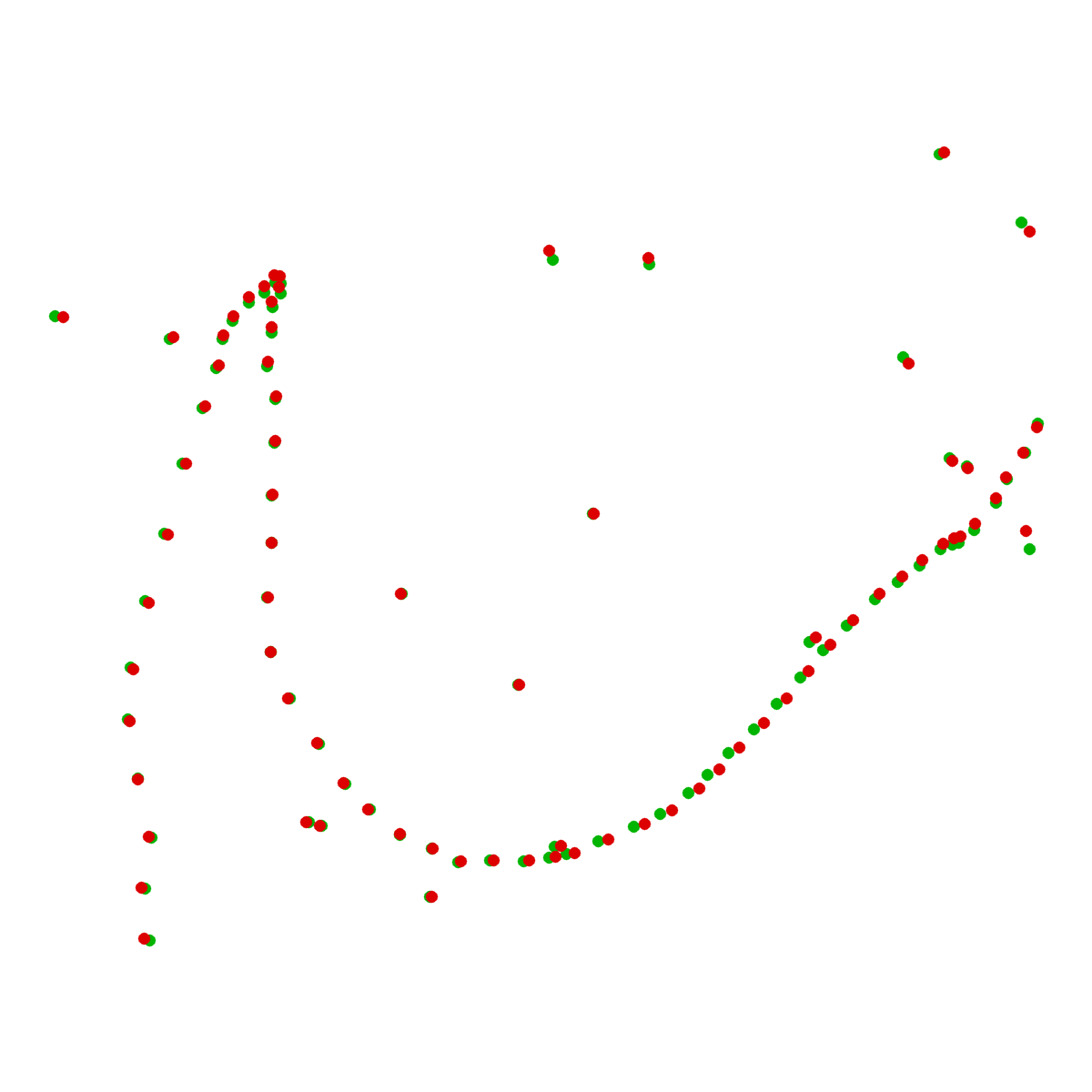}
&
\includegraphics[width=\linewidth]{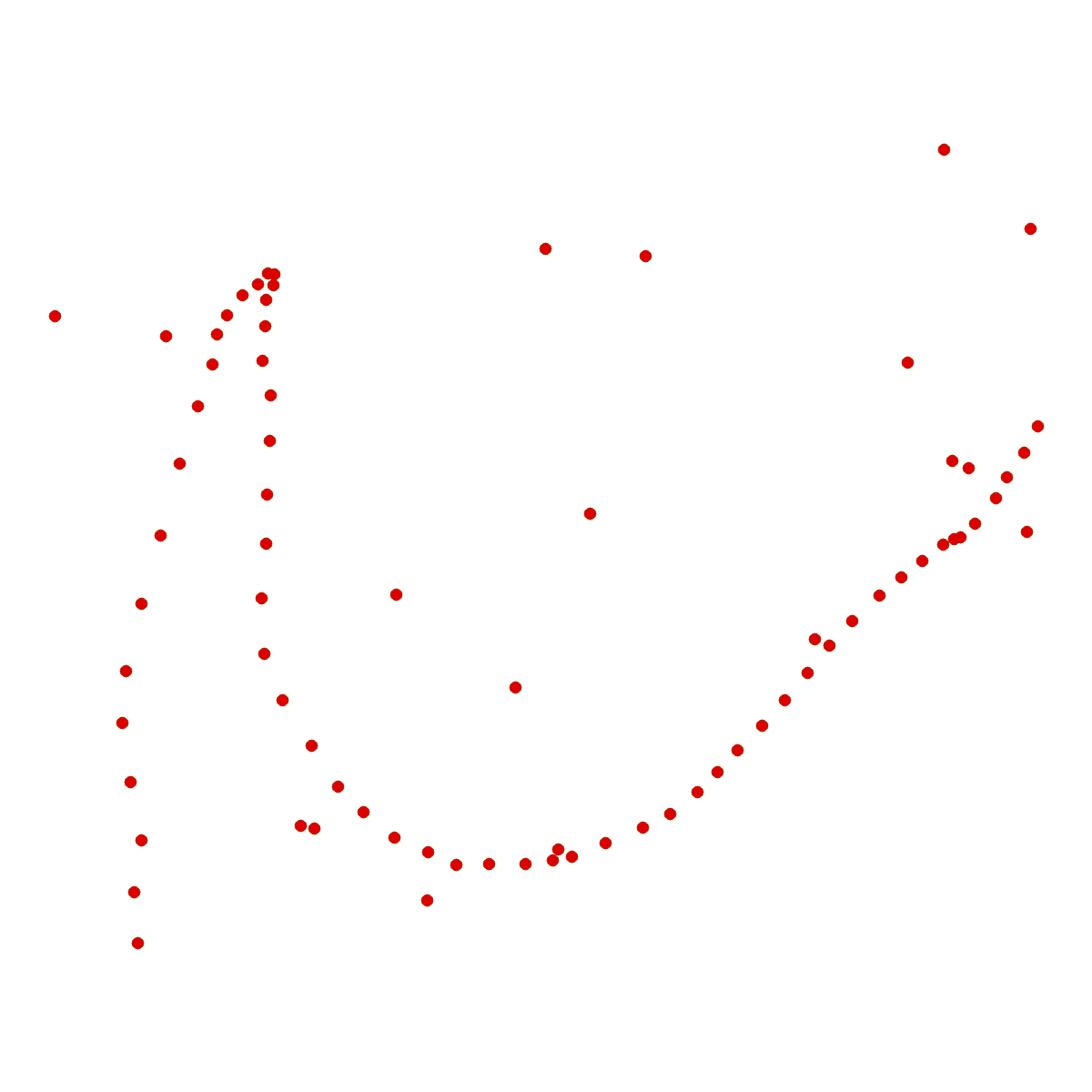}
\\[0.8em]

% ---------- Example 3 ----------
{\bfseries (3)}
&
\includegraphics[width=\linewidth]{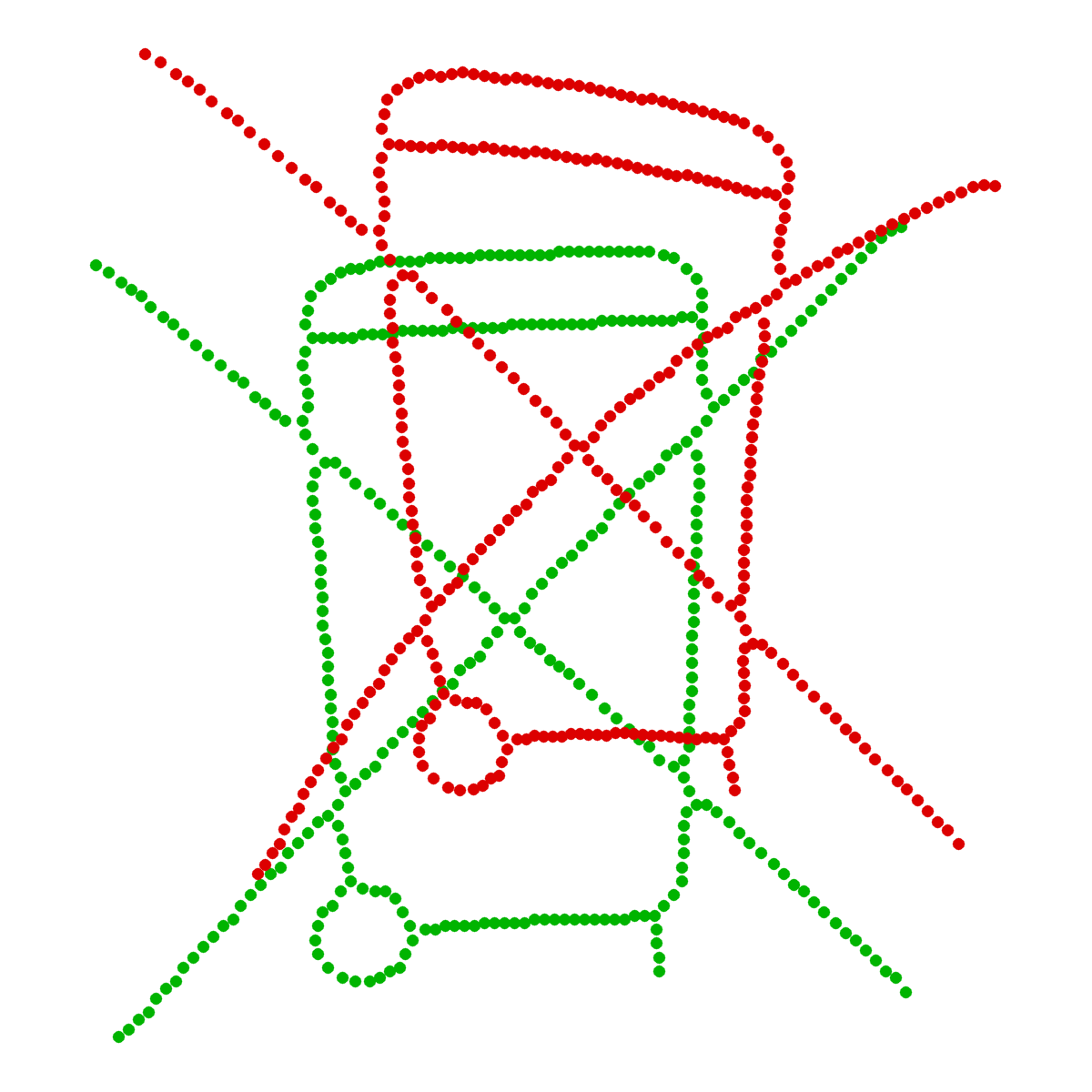}
&
\includegraphics[width=\linewidth]{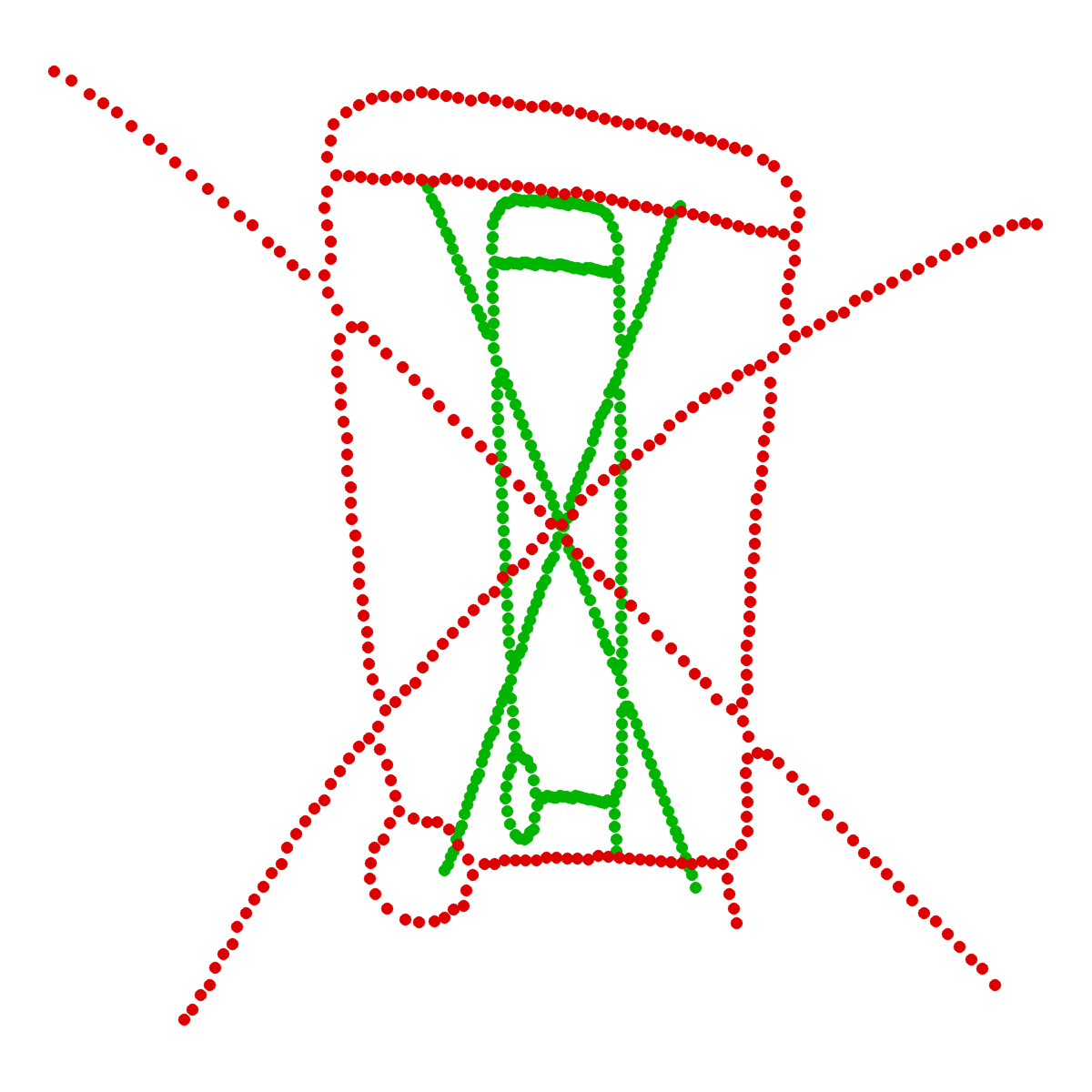}
&
\includegraphics[width=\linewidth]{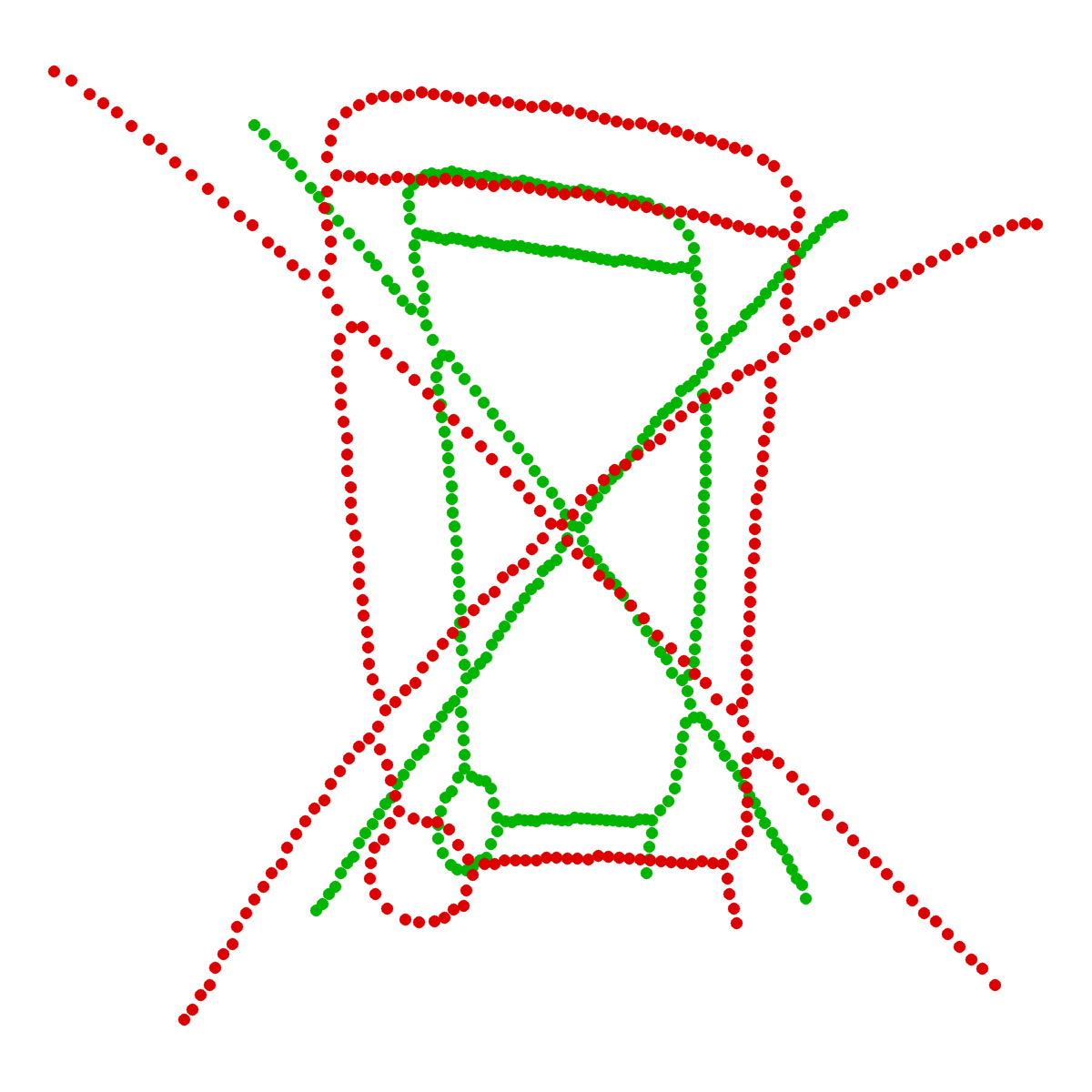}
&
\includegraphics[width=\linewidth]{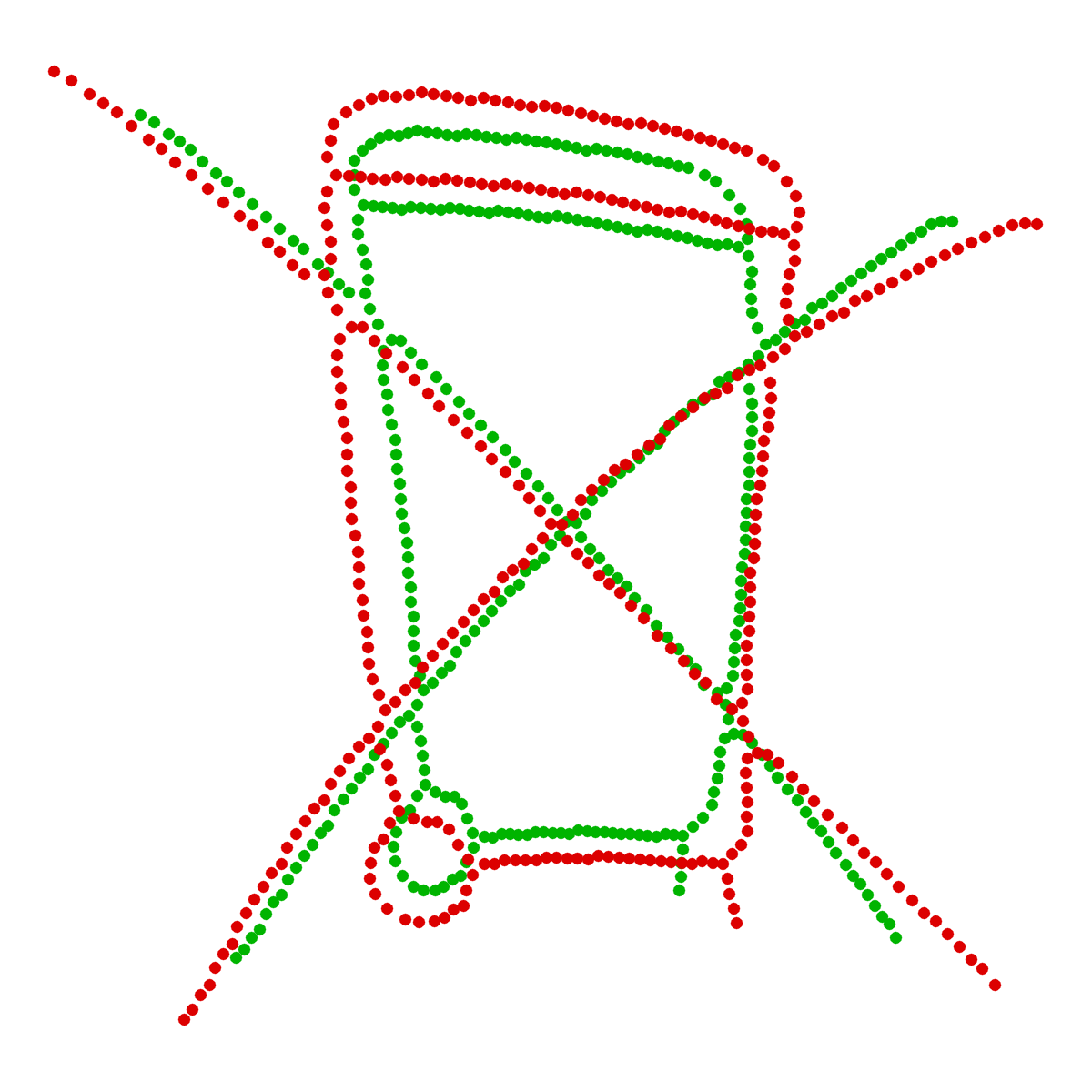}
&
\includegraphics[width=\linewidth]{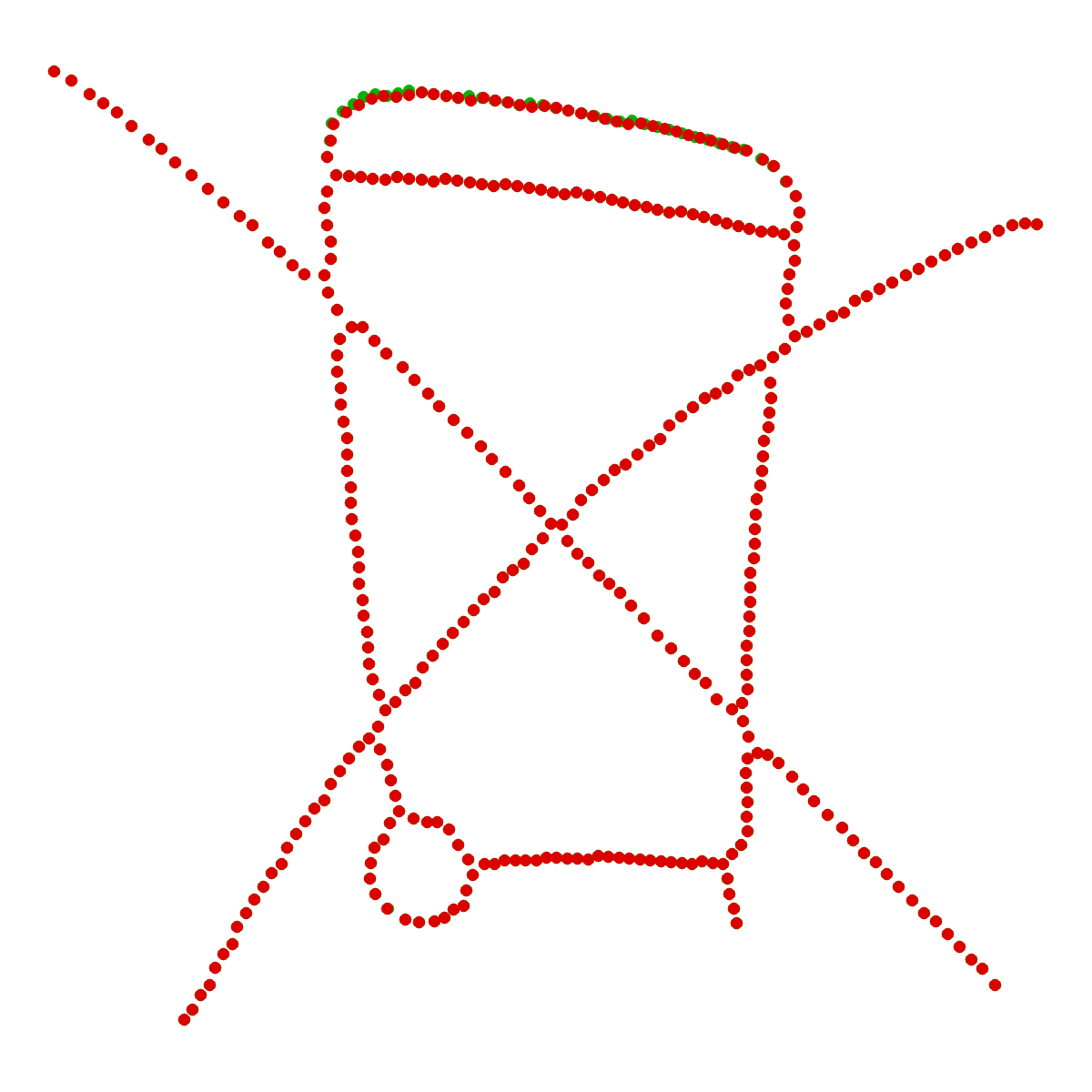}
\end{tabular}
\endgroup

\vspace{0.4em}

\caption{
Two-dimensional large-deformation registration examples using Analytic-CPD.Red points denote the fixed point set, and green points denote the current moving point set during registration. For each example, the process is shown from the initial configuration (Begin), through intermediate iterations (Iter 5, Iter 10, and Iter 15), to the final registered result (End). The
intermediate snapshots are included to visualize the early-stage geometric evolution, while the final stopping iteration is determined by the convergence criterion.
}
\label{fig:2d-large-deformation-examples}

\end{figure*}

Figure~\ref{fig:2d-large-deformation-examples} shows the registration process from the initial configuration to the final result, with intermediate snapshots at iterations 5, 10, and 15. These snapshots are included only to visualize the early-stage geometric evolution, not as fixed stopping iterations. As shown in
the figure, Analytic-CPD progressively aligns the moving point sets with the fixed point sets under clear non-affine deformation. The examples qualitatively show that the CPD posterior layer provides stable soft-correspondence guidance,
while the structured analytic mapping supplies a compact nonlinear deformation model.

\subsubsection{Small-deformation comparison.}

We next evaluate two representative small-deformation examples. In this regime, the initial discrepancy between the moving and fixed point sets is mild, and nearest-neighbor correspondences are usually reliable after only a few iterations. This experiment therefore serves mainly as a consistency check for Analytic-CPD and as a comparison with Analytic-ICP in a regime where
hard-correspondence analytic fitting is expected to be highly effective.

\begin{figure}[t]
\centering
\renewcommand{\arraystretch}{1.15}
\setlength{\tabcolsep}{3pt}

\begin{tabular}{
    >{\centering\arraybackslash}m{0.48\linewidth}
    >{\centering\arraybackslash}m{0.48\linewidth}
}
\toprule
\textbf{\small Initial}
&
\textbf{\small Error--time}
\\
\midrule

\includegraphics[width=\linewidth]{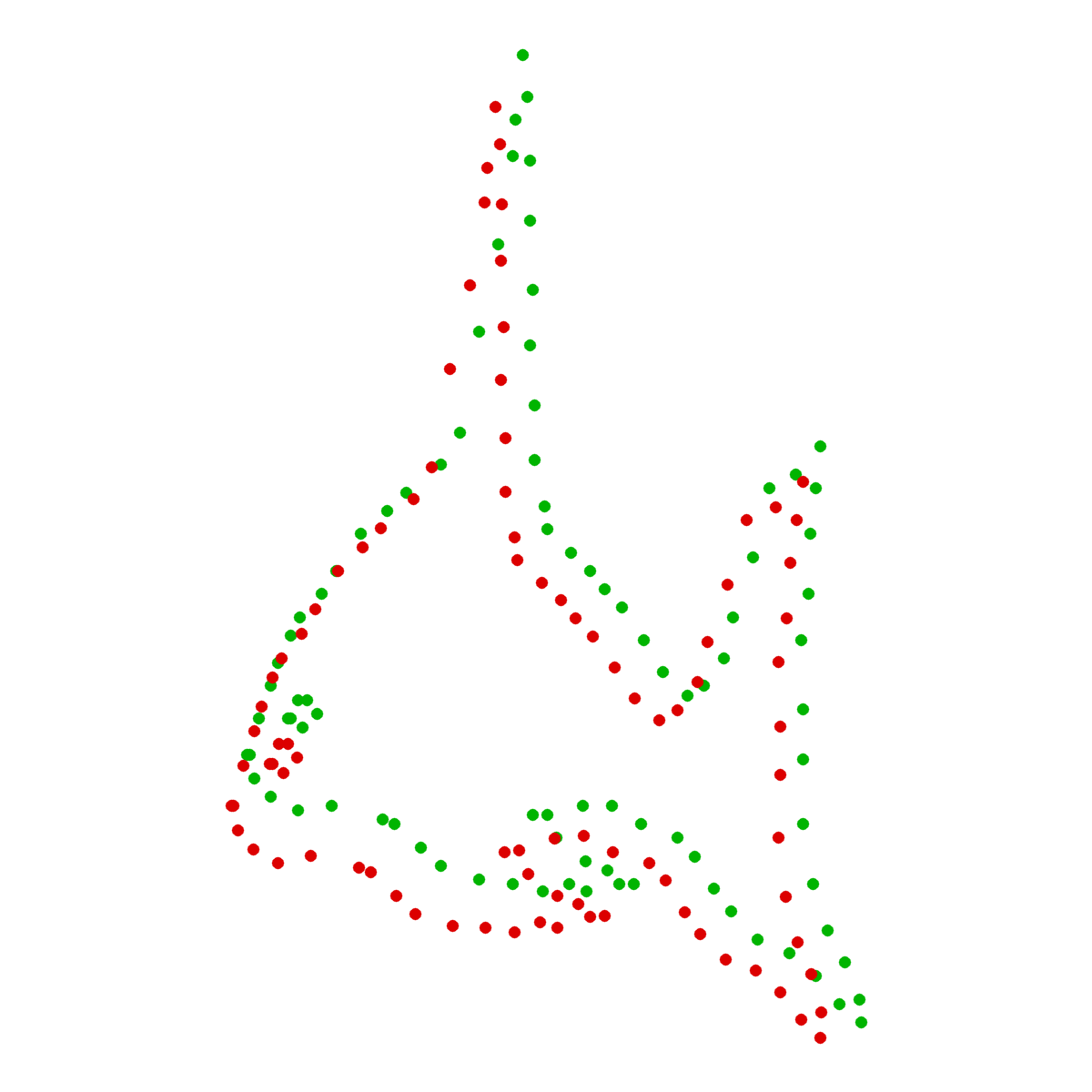}
&
\includegraphics[width=\linewidth]{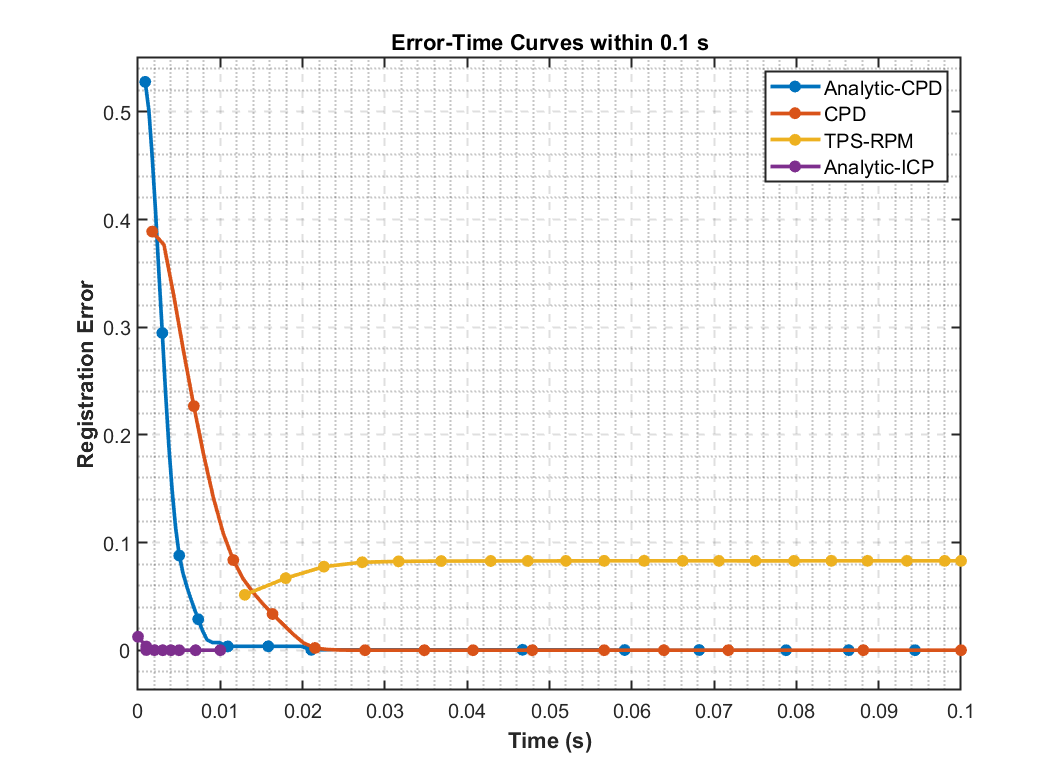}
\\[-0.2em]
\multicolumn{1}{c}{\small (a) Example 1}
&
\multicolumn{1}{c}{\small (b) Example 1}
\\[0.6em]

\includegraphics[width=\linewidth]{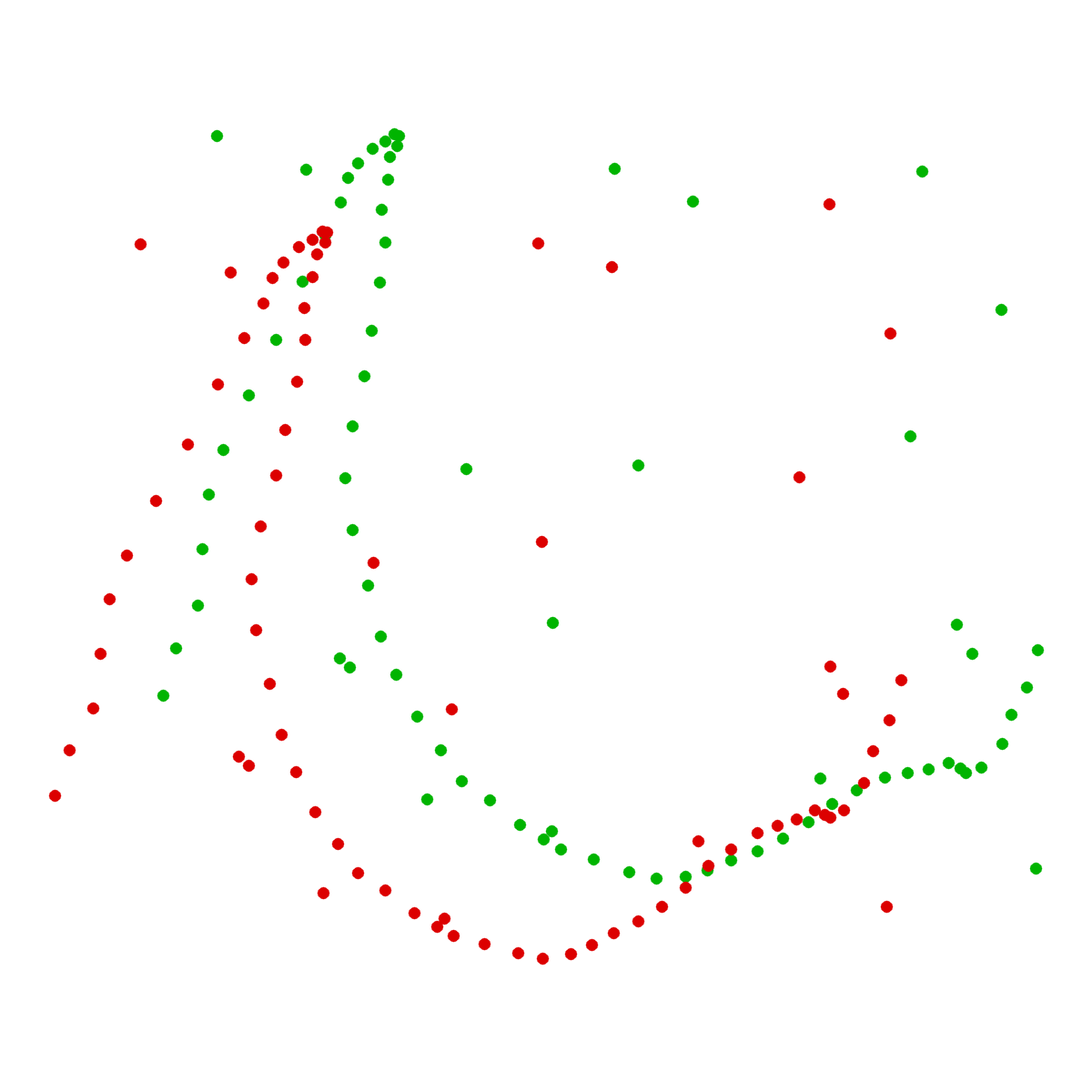}
&
\includegraphics[width=\linewidth]{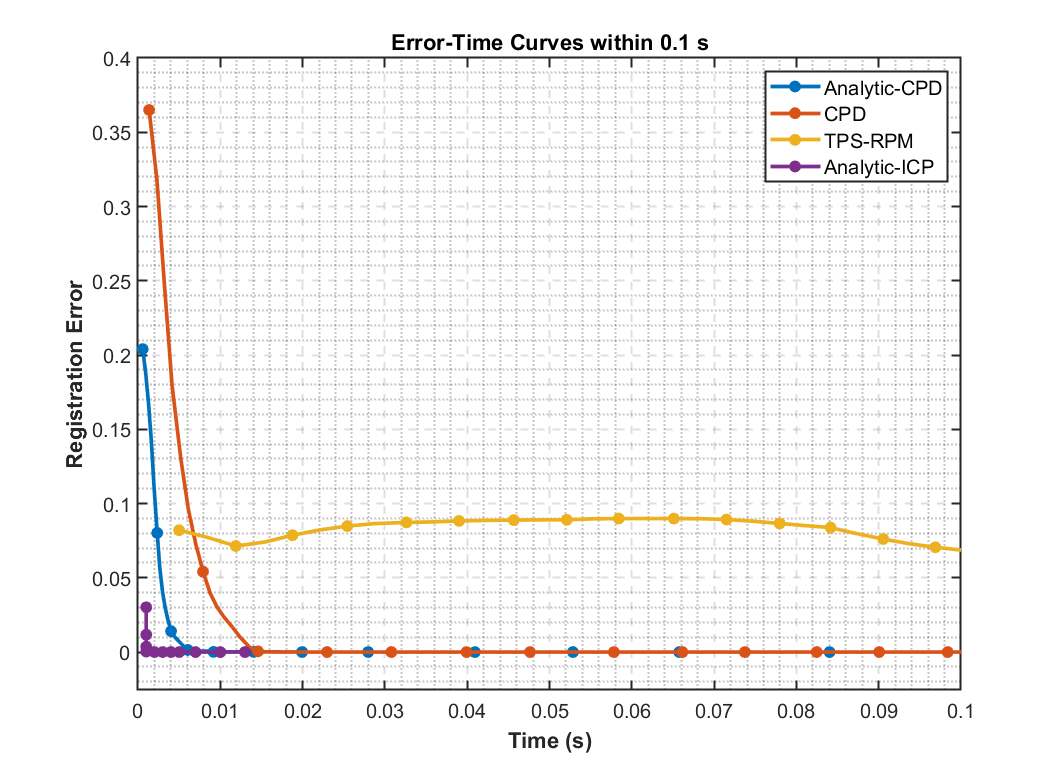}
\\[-0.2em]
\multicolumn{1}{c}{\small (c) Example 2}
&
\multicolumn{1}{c}{\small (d) Example 2}
\\

\bottomrule
\end{tabular}

\caption{
Initial configurations and error--time curves for two representative two-dimensional small-deformation examples. The left column shows the initial relative positions of the moving and fixed point sets, and the right column shows the error--time curves truncated at \(0.1\) seconds to highlight early-stage convergence. BCPD is not shown in the curves because it is evaluated from the final exported result of its official command-line implementation.
}
\label{fig:initial-and-error-time-0p1s}
\end{figure}

Figure~\ref{fig:initial-and-error-time-0p1s} shows the initial configurations and the corresponding error--time curves. The curves are truncated at \(0.1\) seconds to highlight the early-stage convergence behavior. BCPD is not included in the curves because it is evaluated through its official command-line implementation, which does not provide an external RMSE value at
each iteration. Its final exported result is nevertheless evaluated using the same external pointwise RMSE as the other methods. Final errors and total running times are reported in Table~\ref{tab:2d-small-deformation-comparison}.

\begin{table}[t]
\centering
\caption{
Two-dimensional small-deformation comparison.
Errors are measured in the normalized coordinate system.
BCPD is evaluated using its final exported registration result.
}
\label{tab:2d-small-deformation-comparison}

\begin{tabular*}{\linewidth}{@{\extracolsep{\fill}}lccc@{}}
\toprule
Example & Method & Final error & Time (s) \\
\midrule
1 & Analytic-CPD & \(2.33\times 10^{-7}\) & \(0.1847\) \\
1 & Analytic-ICP  & \(8.96\times 10^{-8}\) & \(0.0100\) \\
1 & CPD           & \(3.76\times 10^{-7}\) & \(0.2875\) \\
1 & BCPD          & \(2.10\times 10^{-7}\) & \(0.4400\) \\
1 & TPS-RPM       & \(1.50\times 10^{-3}\) & \(1.5070\) \\
\midrule
2 & Analytic-CPD & \(1.35\times 10^{-7}\) & \(0.1948\) \\
2 & Analytic-ICP  & \(3.31\times 10^{-8}\) & \(0.0130\) \\
2 & CPD           & \(6.40\times 10^{-7}\) & \(0.1721\) \\
2 & BCPD          & \(1.60\times 10^{-7}\) & \(0.3700\) \\
2 & TPS-RPM       & \(5.70\times 10^{-4}\) & \(1.0418\) \\
\bottomrule
\end{tabular*}
\end{table}

The results show that Analytic-CPD, Analytic-ICP, CPD, and BCPD all achieve very small residuals on these small-deformation cases, whereas TPS-RPM produces larger errors and higher running times. Analytic-ICP gives the lowest errors and the fastest runtimes in both examples, which is expected because reliable
nearest-neighbor correspondences are already available in this regime. Analytic-CPD also converges to small residuals: it is comparable to BCPD in Example~1 and slightly more accurate than both CPD and BCPD in Example~2, while remaining faster than BCPD and TPS-RPM.

These results indicate that Analytic-CPD remains competitive even in the small-deformation regime, although this is not the setting where its posterior-guided formulation is expected to have the strongest advantage. The main benefit of Analytic-CPD is examined in the following large-deformation experiments, where hard correspondences can be unreliable in the early iterations and the CPD posterior layer provides a more robust soft-correspondence
structure.

\subsubsection{Large-deformation comparison.}

We next compare the methods on two representative two-dimensional
large-deformation examples. Unlike the small-deformation cases, these examples have much larger initial discrepancies, making early hard correspondences less reliable. This is the regime where the CPD posterior layer is expected to be most useful for guiding the structured analytic mapping.

\begin{figure}[t]
\centering
\renewcommand{\arraystretch}{1.10}
\setlength{\tabcolsep}{0pt}

\begin{tabular}{
@{}c@{\hspace{0.02\columnwidth}}c@{}
}
\toprule
\textbf{\small Initial}
&
\textbf{\small Error--time}
\\
\midrule

% ---------------- Group 1 ----------------
\begin{minipage}[t]{0.47\columnwidth}
\centering
\includegraphics[width=\linewidth]{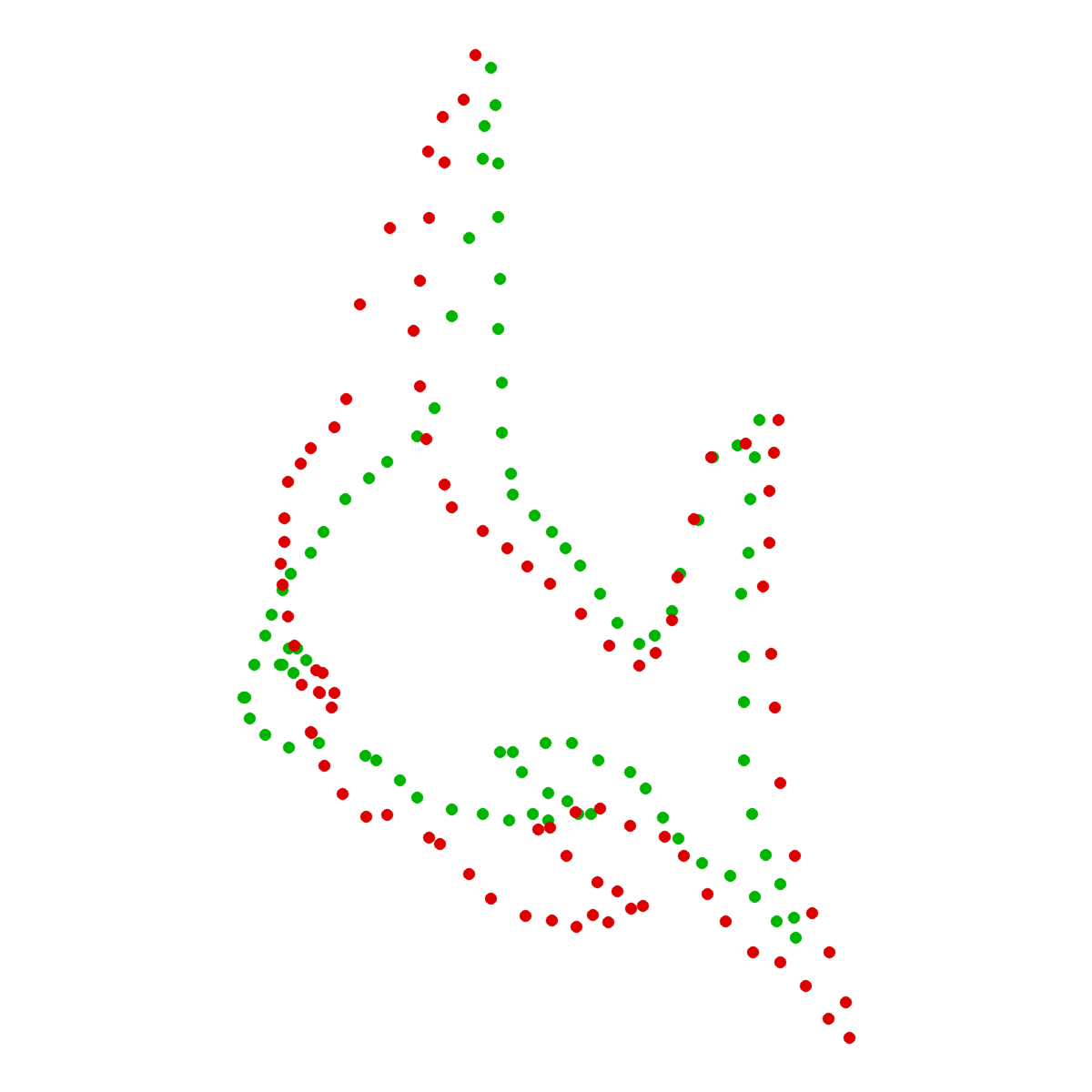}\\[-0.2em]
{\scriptsize (a) Example 1 (Fish, 91 points)}
\end{minipage}
&
\begin{minipage}[t]{0.47\columnwidth}
\centering
\includegraphics[width=\linewidth]{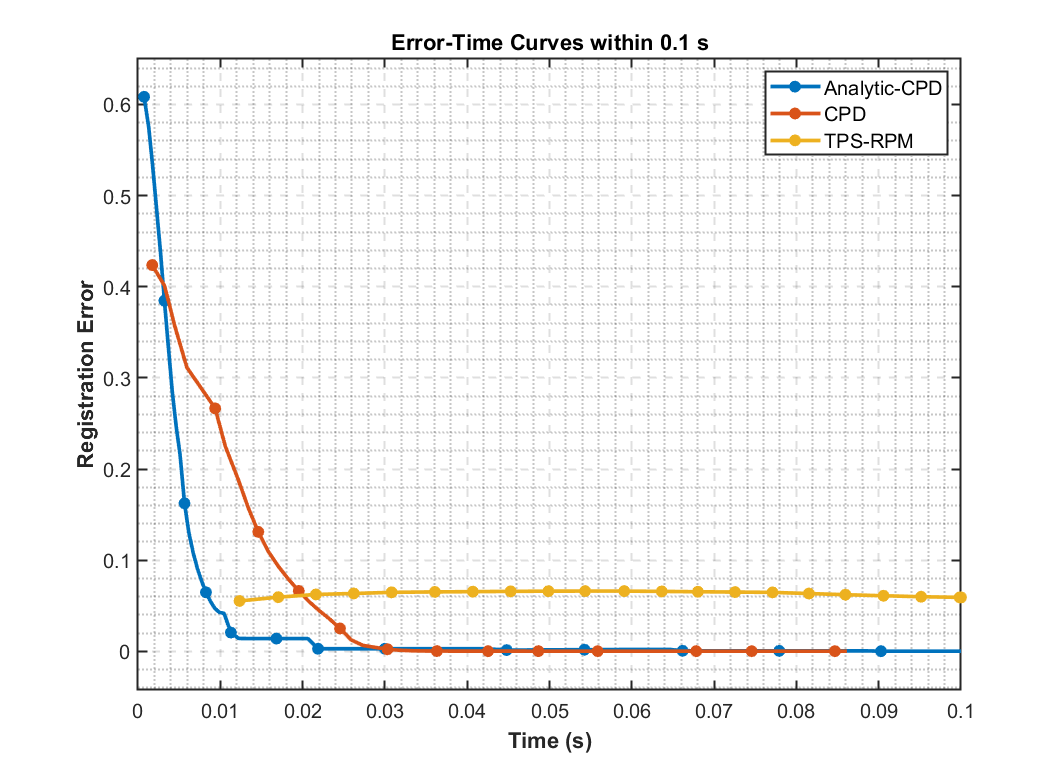}\\[-0.2em]
{\scriptsize (b) Example 1, truncated at 0.1\,s}
\end{minipage}
\\[0.6em]

% ---------------- Group 2 ----------------
\begin{minipage}[t]{0.47\columnwidth}
\centering
\includegraphics[width=\linewidth]{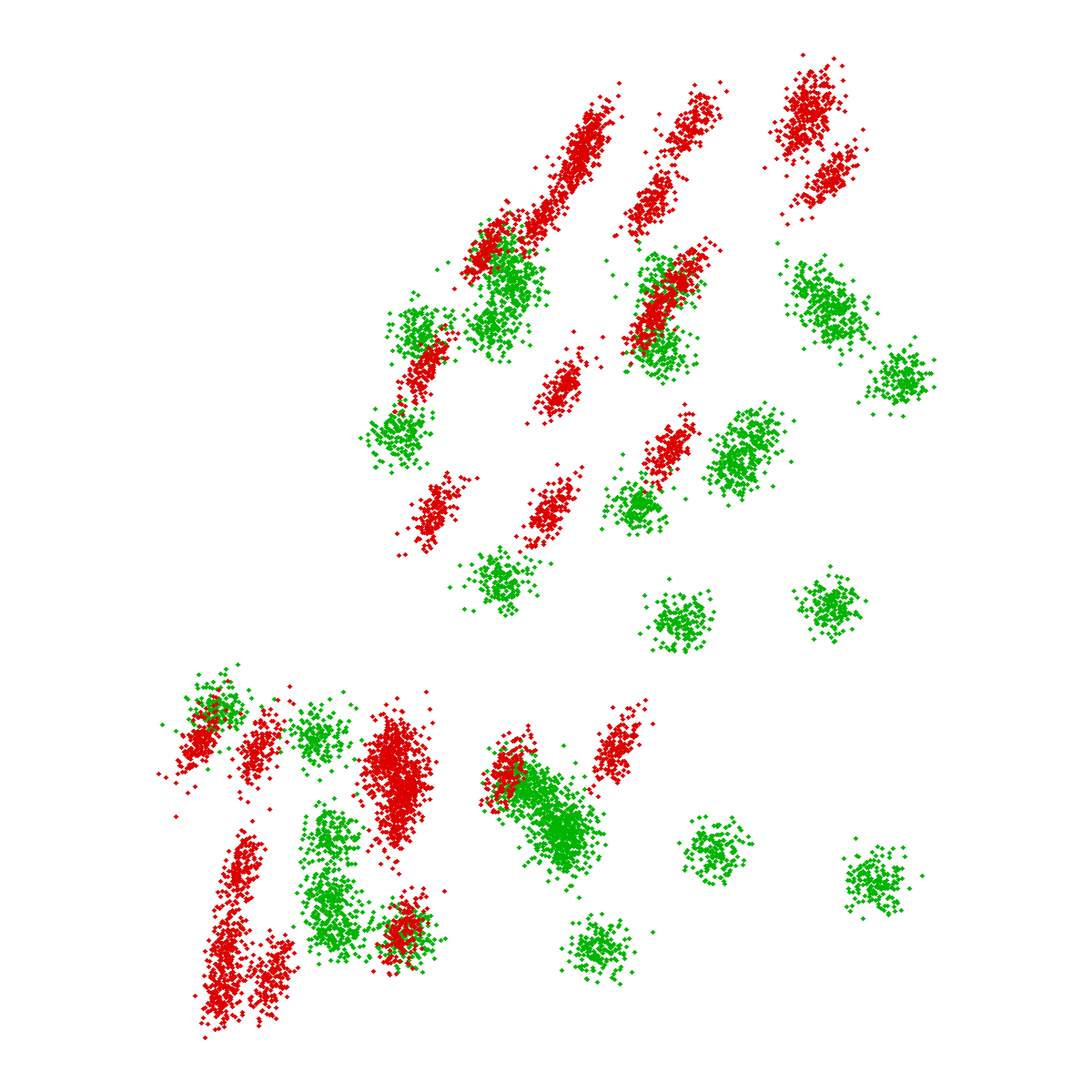}\\[-0.2em]
{\scriptsize (c) Example 2 (Multi-cluster, 5000 points)}
\end{minipage}
&
\begin{minipage}[t]{0.47\columnwidth}
\centering
\includegraphics[width=\linewidth]{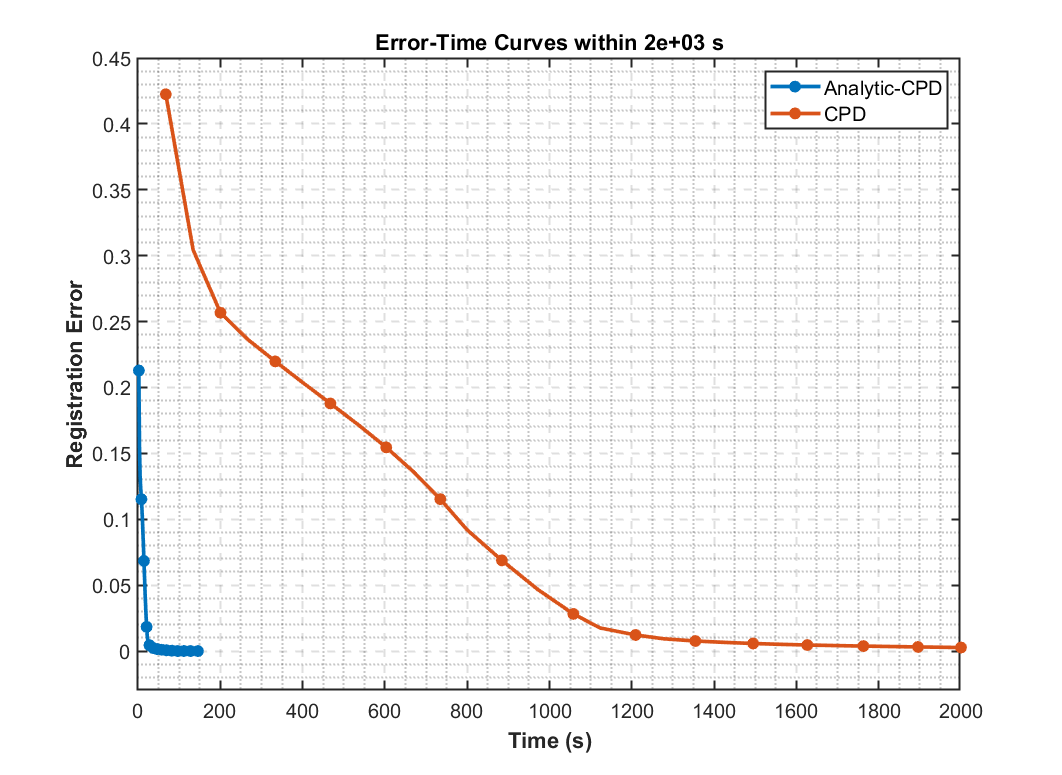}\\[-0.2em]
{\scriptsize (d) Example 2, truncated at 2000\,s}
\end{minipage}
\\

\bottomrule
\end{tabular}

\caption{
Initial configurations and error--time curves for two representative two-dimensional large-deformation examples. The first row shows the fish example with 91 points, whose error--time curve is truncated at \(0.1\)~s.
The second row shows the multi-cluster example with 5000 points, whose curve is truncated at \(2000\)~s.
}
\label{fig:large-deformation-initial-and-error-time}
\end{figure}

Figure~\ref{fig:large-deformation-initial-and-error-time} shows the initial configurations and the corresponding error--time curves. The first example is a fish point set with 91 points, and the second is a multi-cluster point set with 5000 points. For visualization, the error--time curve is truncated at \(0.1\)~s for the fish example and at \(2000\)~s for the multi-cluster example. BCPD is not included in the plotted curves because it is evaluated from the final exported result of its official implementation; its quantitative results are reported in Table~\ref{tab:2d-large-deformation-comparison}.

\begin{table}[t]
\centering
\caption{
Two-dimensional large-deformation comparison.
Errors are measured in the normalized coordinate system.
}
\label{tab:2d-large-deformation-comparison}

\begin{tabular*}{\linewidth}{@{\extracolsep{\fill}}lccc@{}}
\toprule
Example & Method & Final error & Time (s) \\
\midrule
1 (Fish, 91 pts.) & Analytic-CPD & \(4.11\times 10^{-6}\) & \(0.2183\) \\
1 (Fish, 91 pts.) & CPD          & \(1.52\times 10^{-4}\) & \(0.0861\) \\
1 (Fish, 91 pts.) & TPS-RPM      & \(1.72\times 10^{-3}\) & \(1.4828\) \\
1 (Fish, 91 pts.) & BCPD         & \(8.07\times 10^{-6}\) & \(0.3600\) \\
\midrule
2 (Multi-cluster, 5000 pts.) & Analytic-CPD & \(4.62\times 10^{-7}\) & \(145.6275\) \\
2 (Multi-cluster, 5000 pts.) & CPD          & \(8.32\times 10^{-7}\) & \(10790.4560\) \\
2 (Multi-cluster, 5000 pts.) & BCPD         & \(2.27\times 10^{-6}\) & \(122.6910\) \\
\bottomrule
\end{tabular*}
\end{table}

The results show that Analytic-CPD is particularly effective in the large-deformation regime. In the fish example, it achieves the lowest final error, reducing the CPD error from \(1.52\times 10^{-4}\) to \(4.11\times 10^{-6}\). It also improves upon BCPD and substantially outperforms TPS-RPM in accuracy. Although CPD is faster on this small point set, Analytic-CPD remains efficient and is faster than both BCPD and TPS-RPM.

The advantage is more pronounced in the multi-cluster example with 5000 points. Analytic-CPD achieves the lowest final error and reduces the runtime of standard CPD from \(10790.5\)~s to \(145.6\)~s. BCPD is slightly faster in this case, but its final error is higher. These results indicate that the proposed
posterior-guided structured analytic M-step provides a favorable accuracy--time trade-off under large non-affine deformation.

Overall, Analytic-CPD consistently improves the final accuracy over standard CPD in these large-deformation examples. On the larger point set, it also substantially reduces the computational cost. The results support the intended role of Analytic-CPD: when hard correspondences are unreliable, the CPD posterior layer provides robust soft assignment, while the structured analytic
mapping supplies a compact nonlinear deformation model.

\subsection{Three-Dimensional Smooth Non-Analytic Deformation}
\label{subsec:3d-smooth-nonanalytic-deformation}

This subsection provides the main model-mismatch evaluation of Analytic-CPD on three-dimensional point clouds under smooth but non-analytic deformations. Unlike the two-dimensional analytic experiments, the ground-truth deformation used here is not a single finite-order Taylor map, but a smoothly blended
spatially varying field. This setting evaluates whether the structured analytic M-step can approximate general smooth nonlinear deformation in 3D, rather than only recover deformation generated from the same analytic model class.

We use a two-center smoothly blended quadratic field to generate the deformed point clouds. Let \(c_1,c_2\in\mathbb R^3\) be two spatial centers, chosen as the points with the minimum and maximum \(x\)-coordinates of the model. The baseline Taylor coefficient blocks are initialized by drawing their entries independently from
\[
\mathcal U[-\gamma_0,\gamma_0],
\qquad
\gamma_0=0.4.
\]
The first-order diagonal entries are then fixed to identity, while the remaining first-order off-diagonal entries are kept as initialized. This produces a global affine component close to the identity but with nontrivial shear and translation components.

Let \(\bar Q\in\mathbb R^{3\times 6}\) denote the backed-up second-order coefficient block. Two local quadratic templates are formed as
\[
Q_1=\bar Q+\Delta_1,
\qquad
Q_2=\bar Q+\Delta_2,
\]
where
\[
(\Delta_1)_{ij},(\Delta_2)_{ij}
\sim
\mathcal U[-\gamma_Q,\gamma_Q],
\qquad
\gamma_Q=0.2.
\]
Thus, the two local quadratic fields share the same baseline second-order structure but differ by relatively large local perturbations, generating a strong spatially varying nonlinear deformation.

The two quadratic templates are blended by compactly supported \(C^\infty\) bump weights. For
\[
r_i(y)=\|y-c_i\|,
\qquad i=1,2,
\]
we define
\[
b(r;\sigma)
=
\begin{cases}
\exp\!\left(-\dfrac{1}{1-(r/\sigma)^2}\right), & r<\sigma,\\[0.6em]
0, & r\ge \sigma,
\end{cases}
\]
and use the normalized weights
\[
w_i(y)
=
\frac{b(\|y-c_i\|;\sigma)}
     {b(\|y-c_1\|;\sigma)+b(\|y-c_2\|;\sigma)},
\qquad
i=1,2.
\]
In the experiments, \(\sigma\) is chosen according to the distance between the two centers so that the sampled point-cloud domain is covered by the two localized influence regions.

The resulting deformation is
\begin{equation}
\tau(y)
=
Ay+t+
\frac{1}{2}
\left(
w_1(y)Q_1+w_2(y)Q_2
\right)
\phi^{[2]}(y-\mathfrak c),
\label{eq:3d-smooth-nonanalytic-deformation}
\end{equation}
where \(A\) and \(t\) are the first- and zeroth-order components, \(\mathfrak c\) is the Taylor expansion center, and
\[
\phi^{[2]}(u)
=
\left[
u_1^2,\;
2u_1u_2,\;
2u_1u_3,\;
u_2^2,\;
2u_2u_3,\;
u_3^2
\right]^\top .
\]
The factor \(1/2\) makes this convention consistent with the second-order Taylor basis used in the structured analytic mapping model.

The bump function \(b(r;\sigma)\) is \(C^\infty\), but it is not analytic at the boundary \(r=\sigma\): all derivatives vanish at the boundary, while no convergent power series can represent the function across that boundary. Consequently, when \(Q_1\neq Q_2\), the blended field in \eqref{eq:3d-smooth-nonanalytic-deformation} is smooth but not a single global analytic mapping. This provides a heterogeneous non-analytic test bed for evaluating the approximation capability of Analytic-CPD under large three-dimensional deformation.

\subsubsection{Large-deformation examples.}

We next present three representative three-dimensional large-deformation registration examples to illustrate the qualitative behavior of Analytic-CPD in 3D. The first two examples are based on human point-cloud models from SHREC'19, containing 6890 and 10050 points, respectively. These two point clouds are used as source geometries and are deformed by the smooth non-analytic deformation model described above. Therefore, they provide controlled large-deformation examples with known pointwise correspondences, while still using realistic human-body point-cloud geometry. The third example is taken from the MPI-FAUST human dataset~\cite{bogo2014faust} and corresponds to a real articulated human-motion registration scenario. In each example, the fixed point set is shown in red and the moving point set is shown in green.

\begin{figure*}[t]
\centering

% ---------- legend ----------
\begingroup
\setlength{\tabcolsep}{1em}
\begin{tabular}{cc}
\textcolor{red}{\large$\bullet$}\ \ Red = fixed
&
\textcolor{green!70!black}{\large$\bullet$}\ \ Green = moving / moved
\end{tabular}
\endgroup

\vspace{0.8em}

% ---------- main table ----------
\begingroup
\setlength{\tabcolsep}{0.35em}
\renewcommand{\arraystretch}{1.15}

\begin{tabular}{
    >{\centering\arraybackslash}m{0.10\textwidth}
    >{\centering\arraybackslash}m{0.16\textwidth}
    >{\centering\arraybackslash}m{0.16\textwidth}
    >{\centering\arraybackslash}m{0.16\textwidth}
    >{\centering\arraybackslash}m{0.16\textwidth}
    >{\centering\arraybackslash}m{0.16\textwidth}
}
&
{\bfseries Begin}
&
{\bfseries Iter 5}
&
{\bfseries Iter 15}
&
{\bfseries Iter 30}
&
{\bfseries End}
\\[0.5em]

% ---------- Example 1 ----------
\begin{tabular}[c]{@{}c@{}}
\bfseries (1)\\[-0.2em]
\small SHREC'19\\[-0.1em]
\small 6890 pts.
\end{tabular}
&
\includegraphics[width=\linewidth]{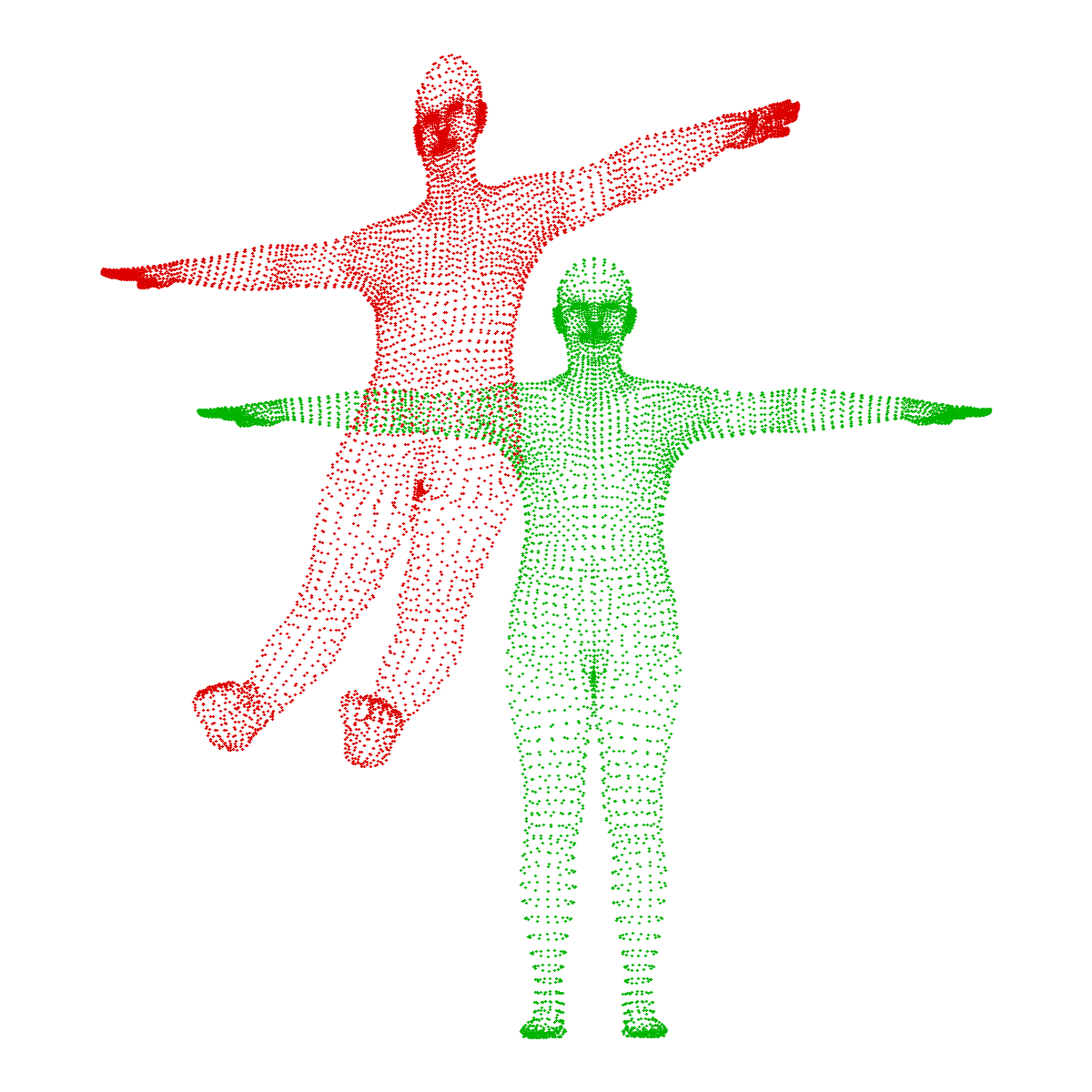}
&
\includegraphics[width=\linewidth]{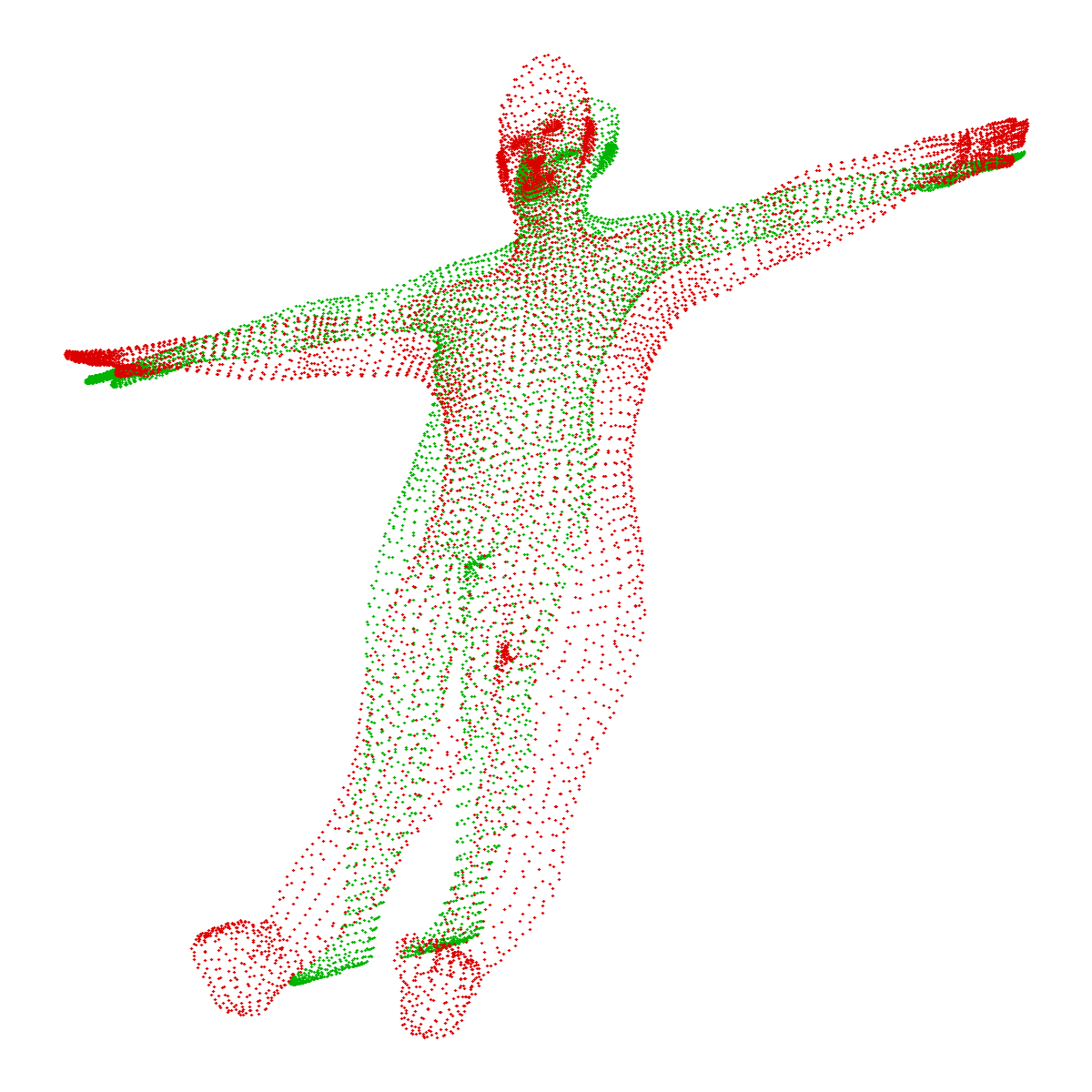}
&
\includegraphics[width=\linewidth]{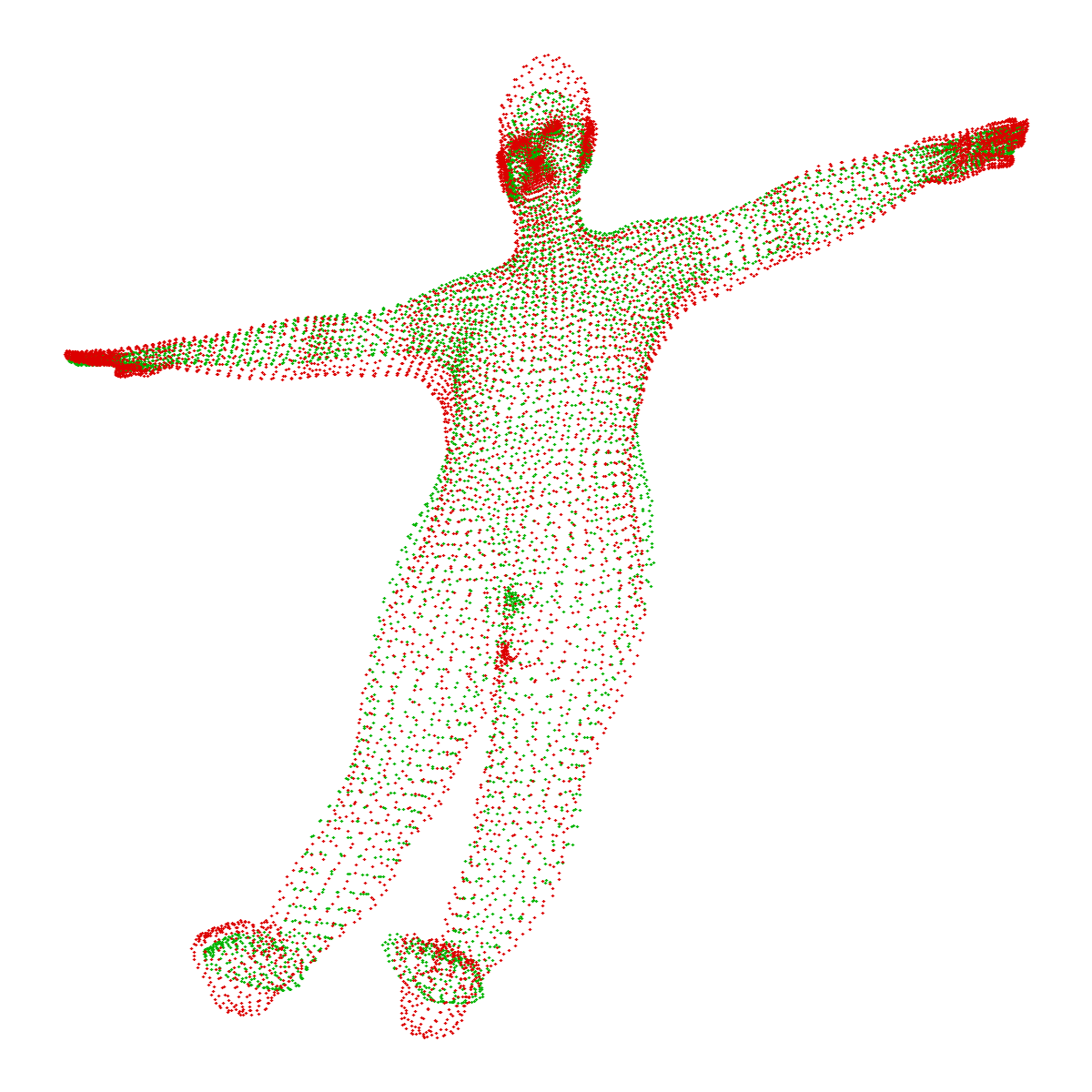}
&
\includegraphics[width=\linewidth]{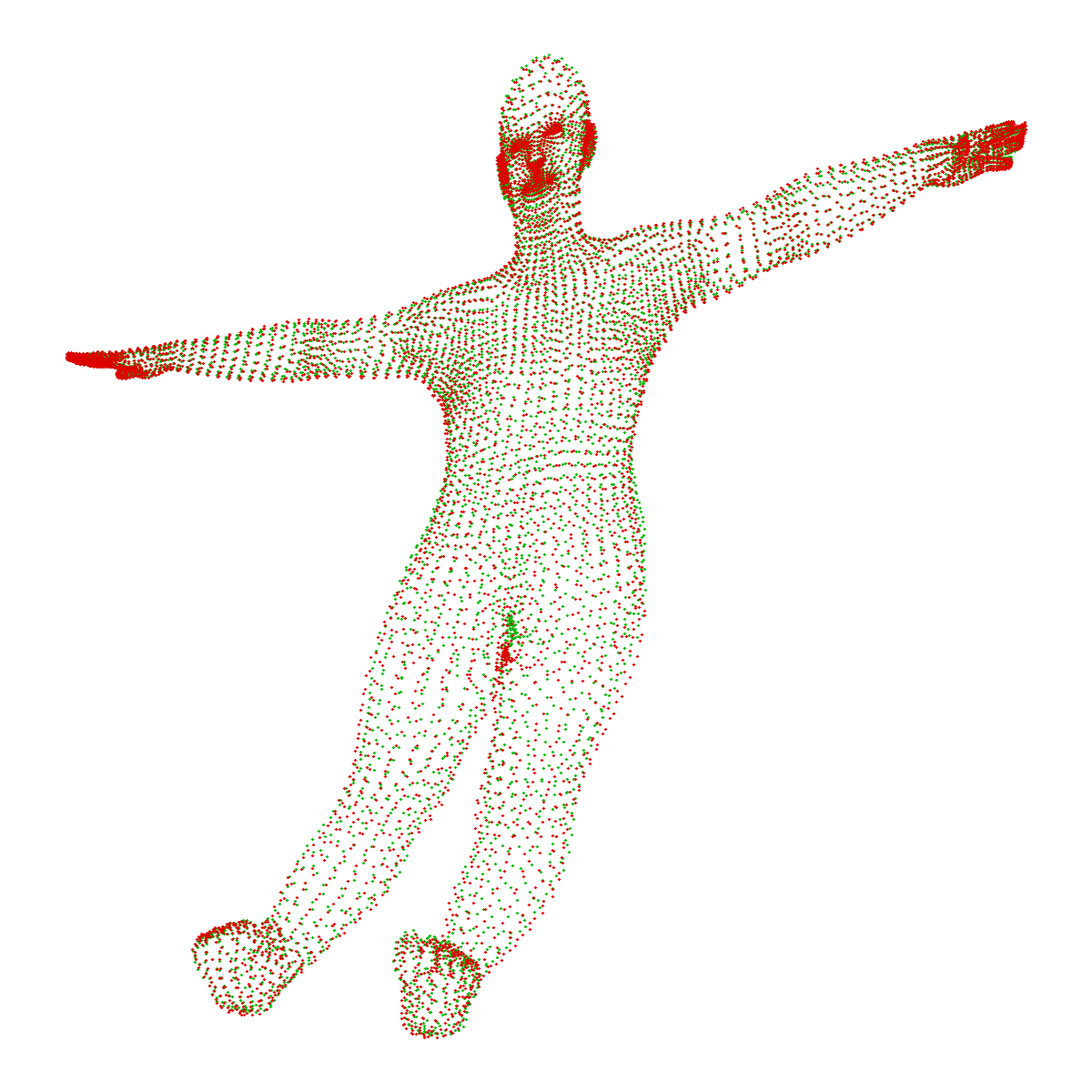}
&
\includegraphics[width=\linewidth]{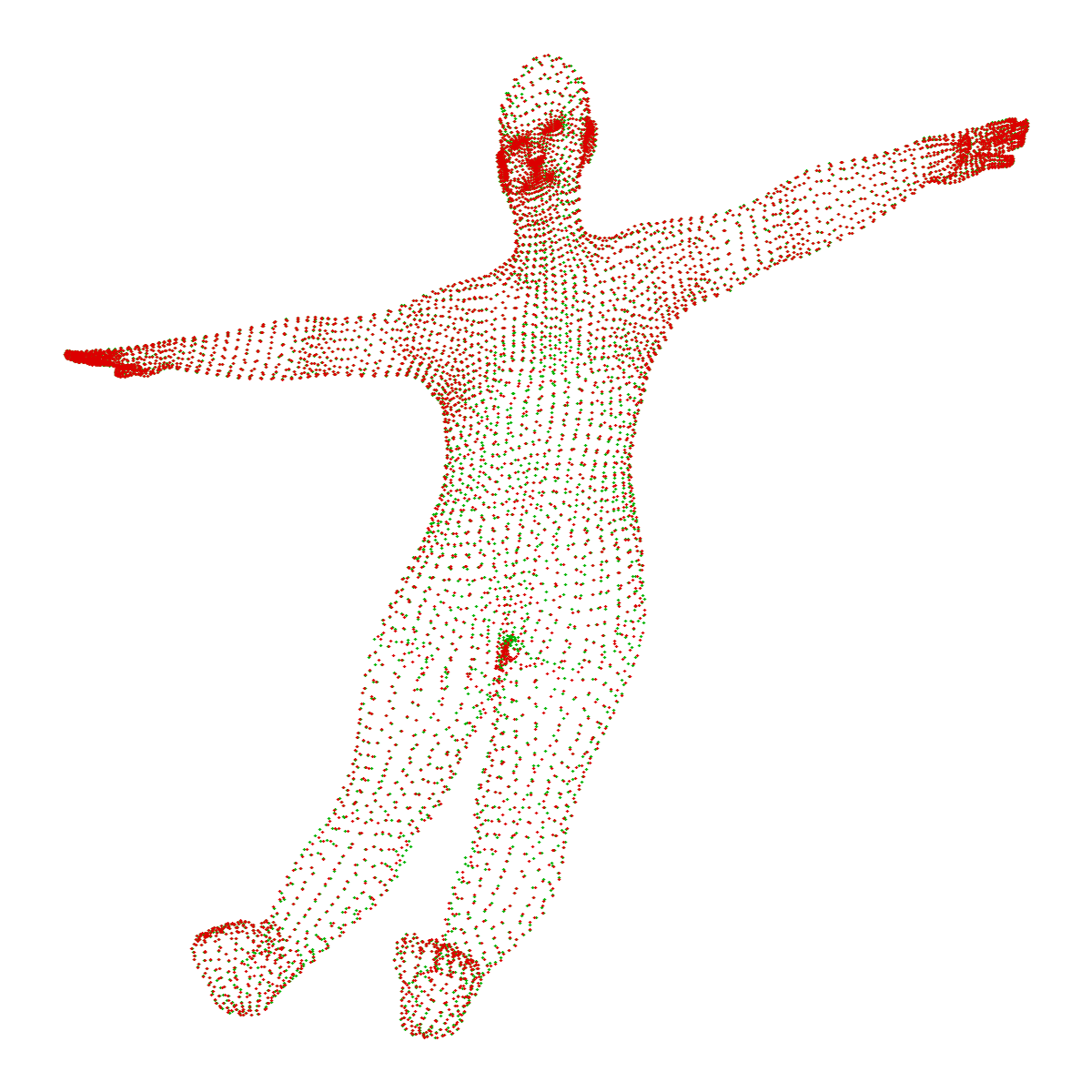}
\\[0.9em]

% ---------- Example 2 ----------
\begin{tabular}[c]{@{}c@{}}
\bfseries (2)\\[-0.2em]
\small SHREC'19\\[-0.1em]
\small 10050 pts.
\end{tabular}
&
\includegraphics[width=\linewidth]{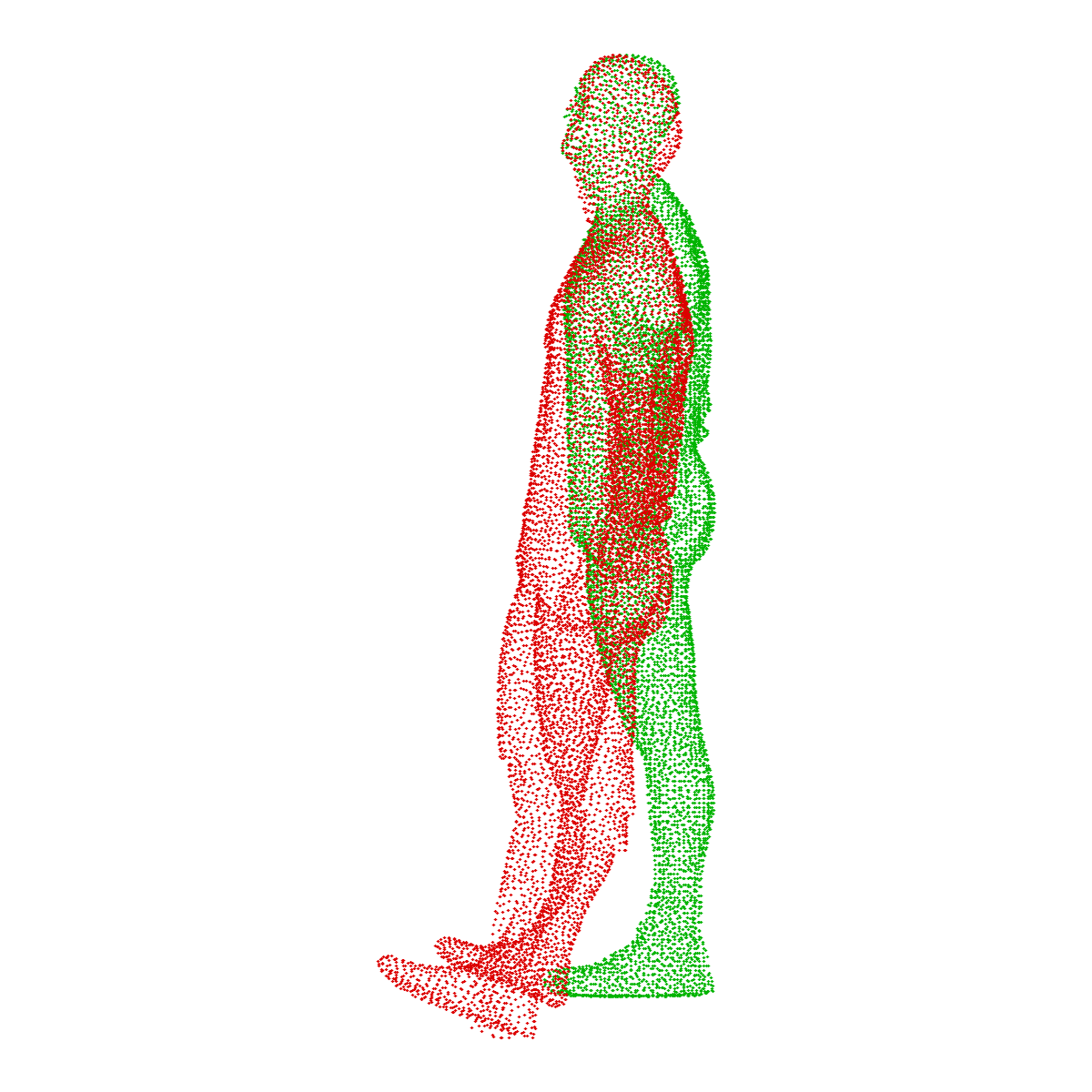}
&
\includegraphics[width=\linewidth]{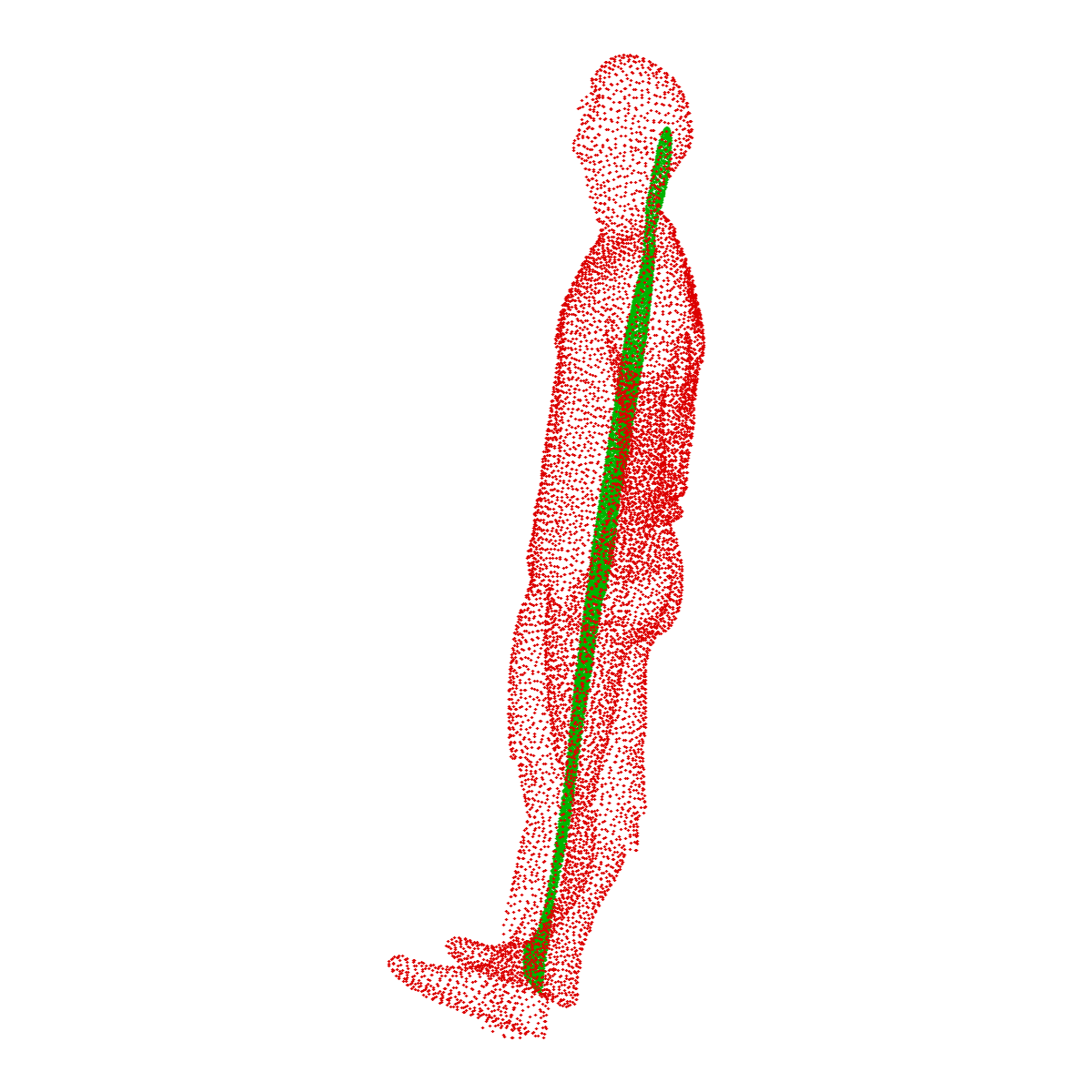}
&
\includegraphics[width=\linewidth]{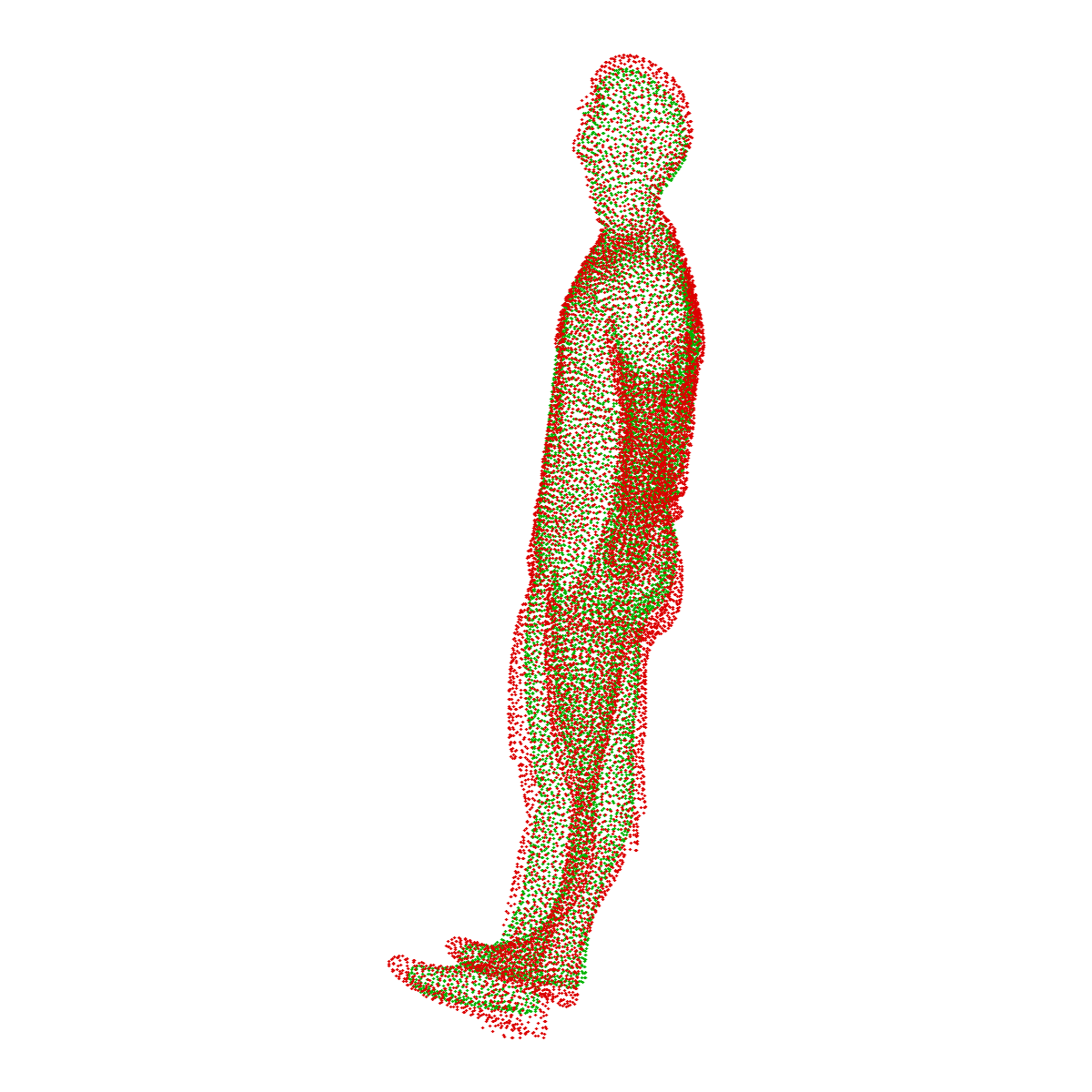}
&
\includegraphics[width=\linewidth]{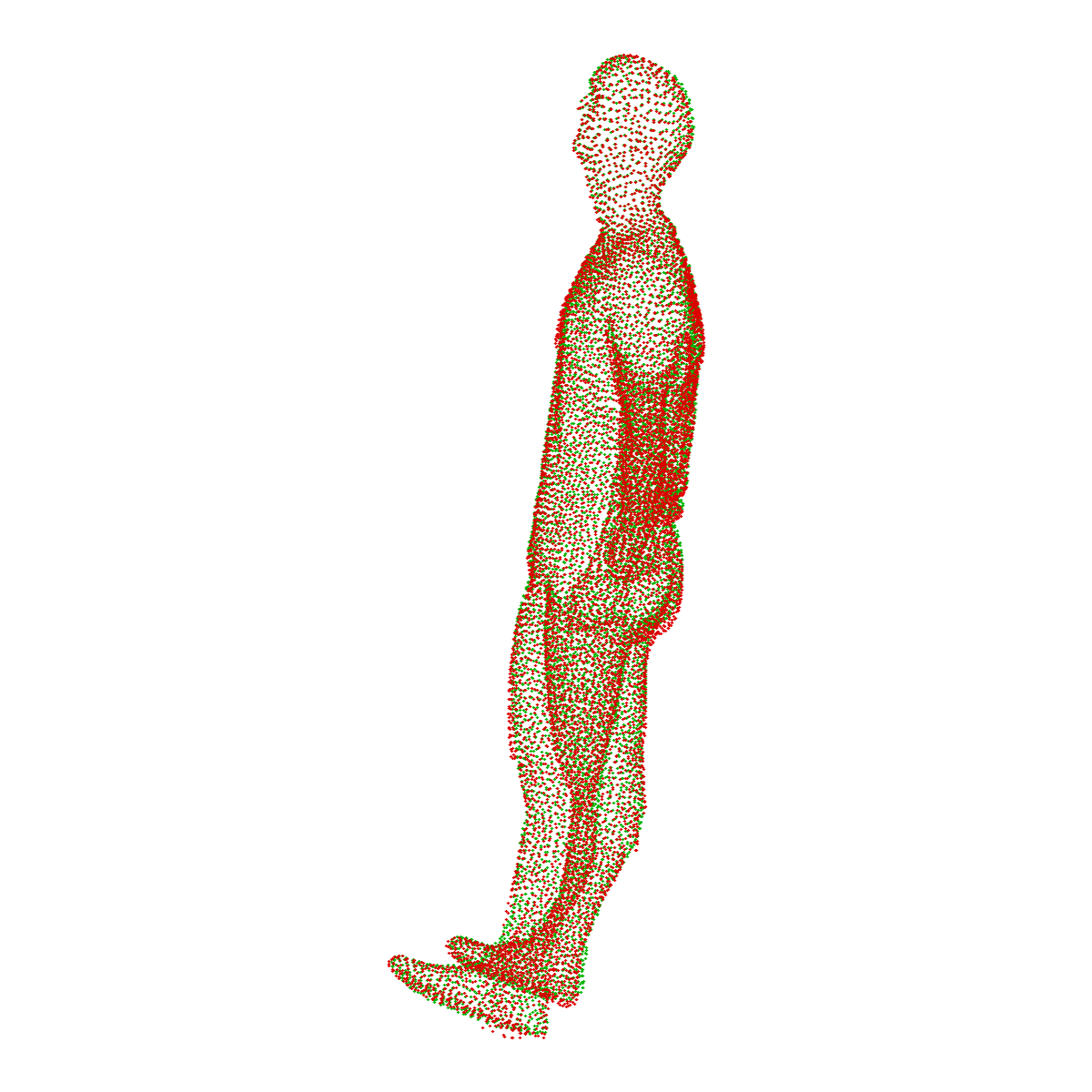}
&
\includegraphics[width=\linewidth]{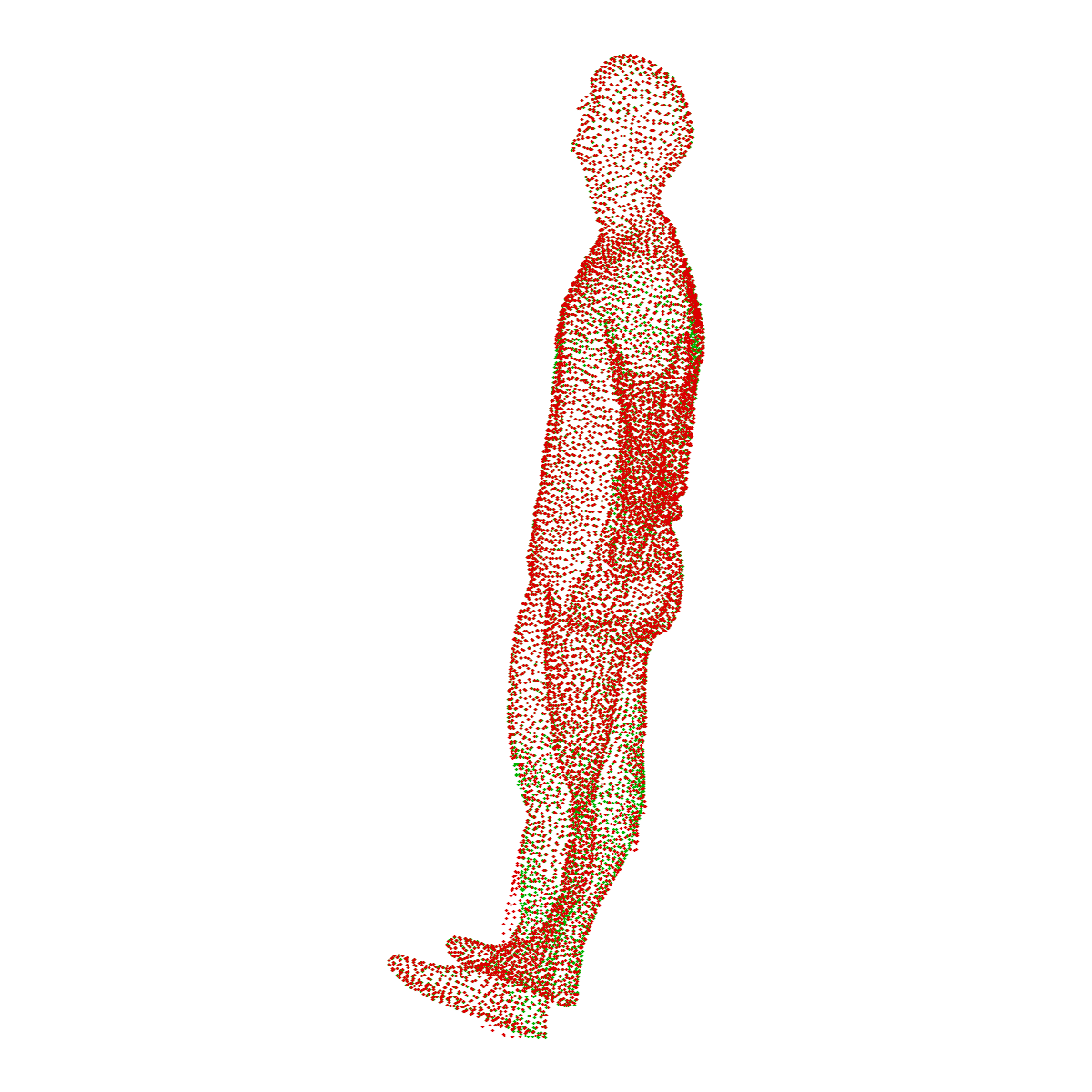}
\\[0.9em]

% ---------- Example 3 : MPI-FAUST ----------
\begin{tabular}[c]{@{}c@{}}
\bfseries (3)\\[-0.2em]
\small MPI-FAUST
\end{tabular}
&
\includegraphics[width=\linewidth]{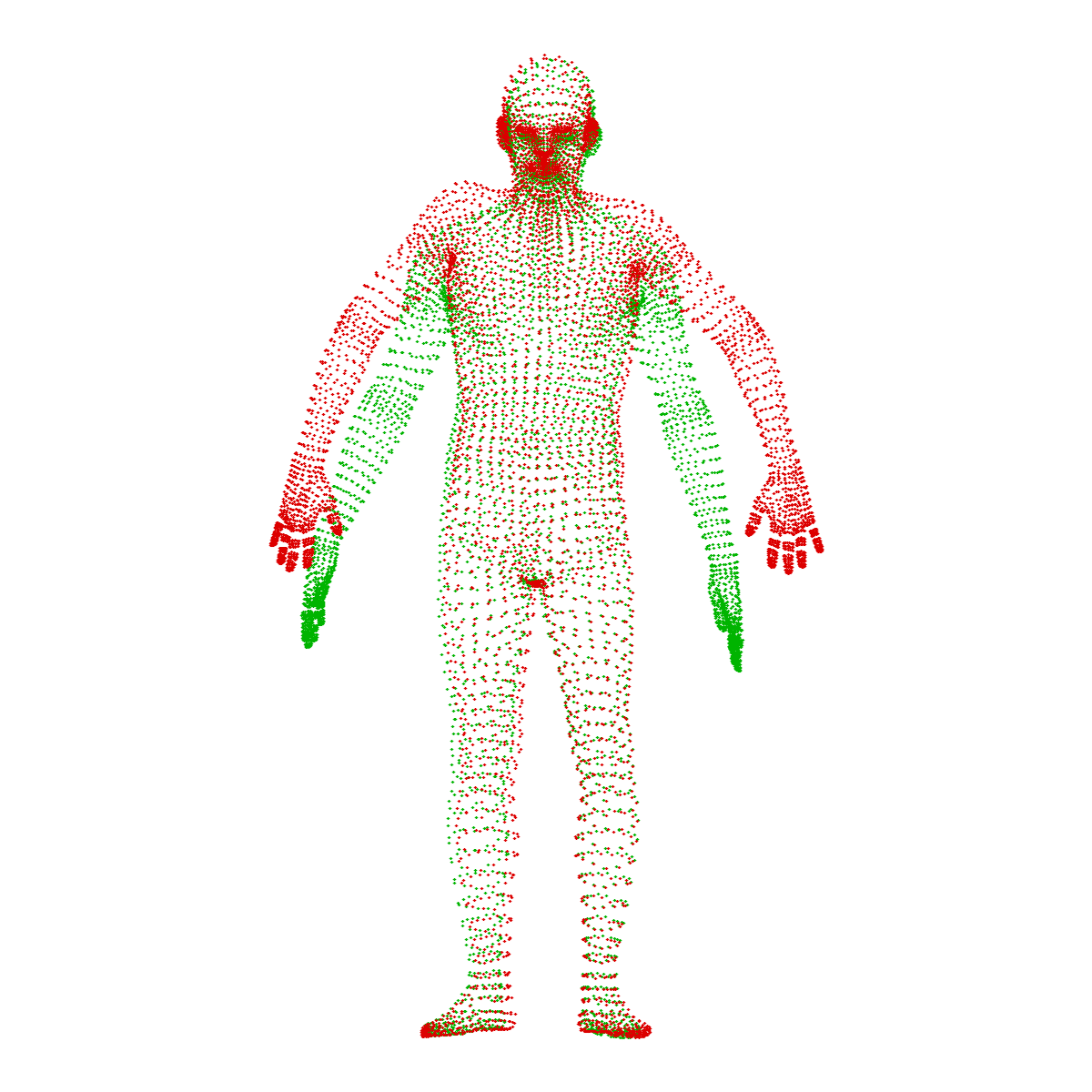}
&
\includegraphics[width=\linewidth]{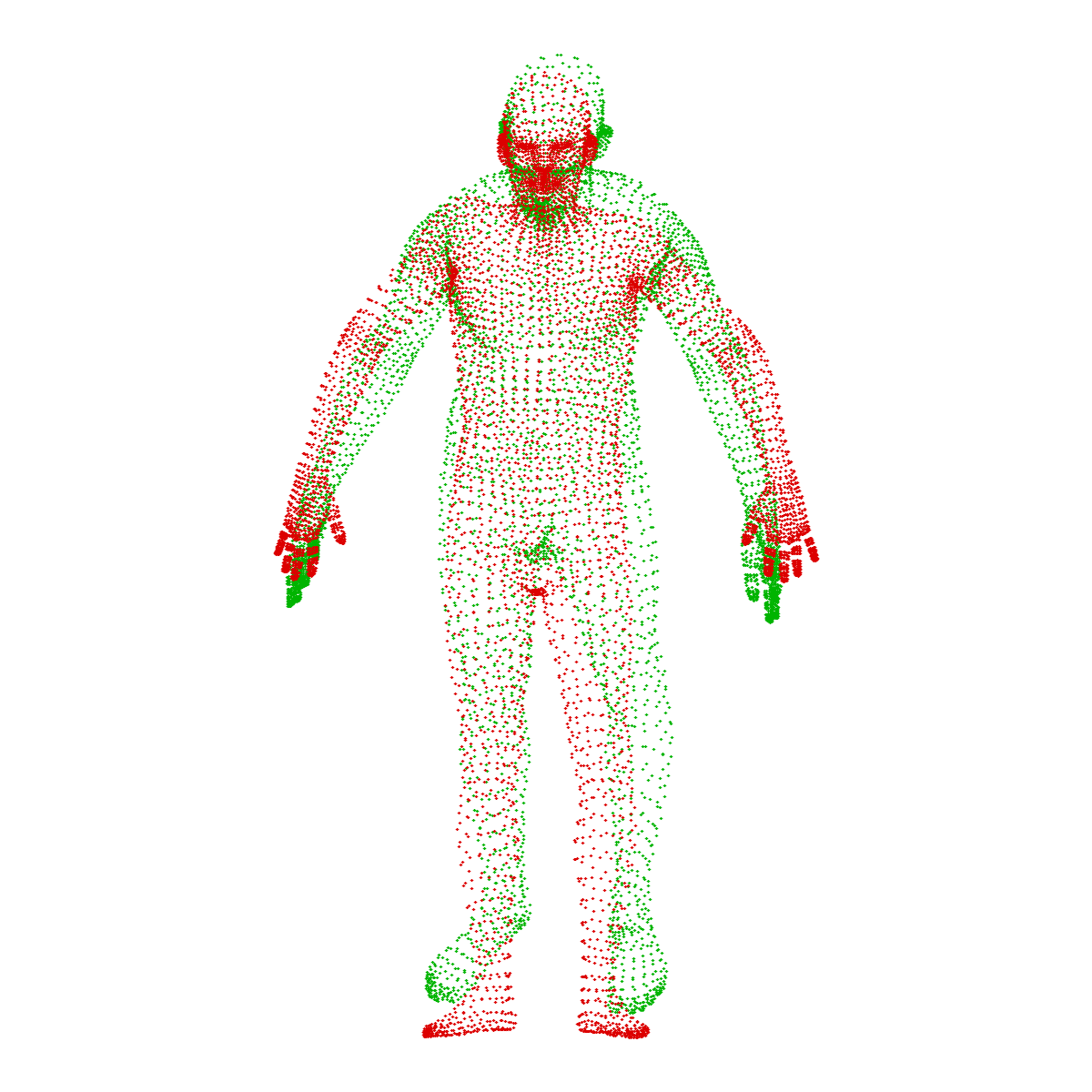}
&
\includegraphics[width=\linewidth]{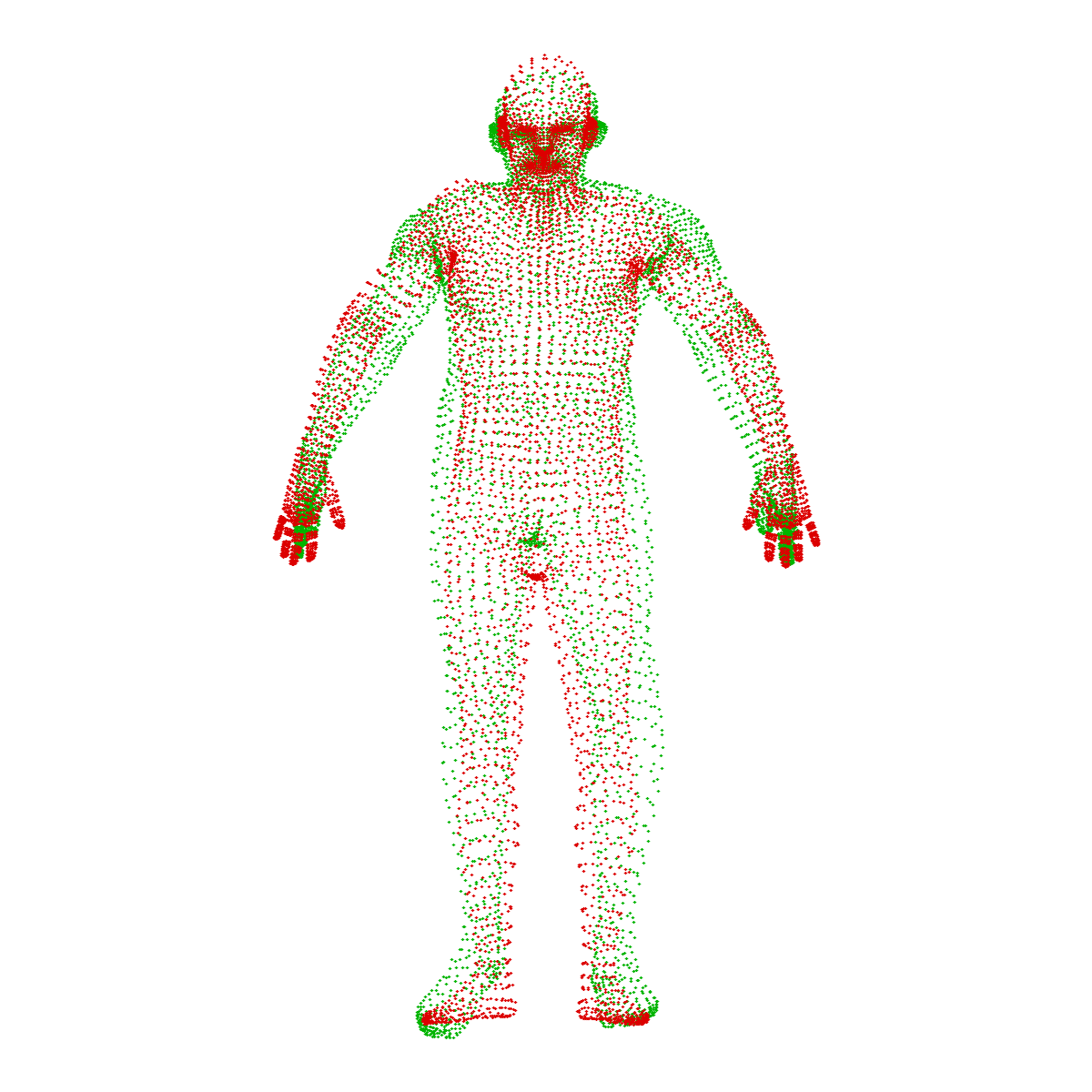}
&
\includegraphics[width=\linewidth]{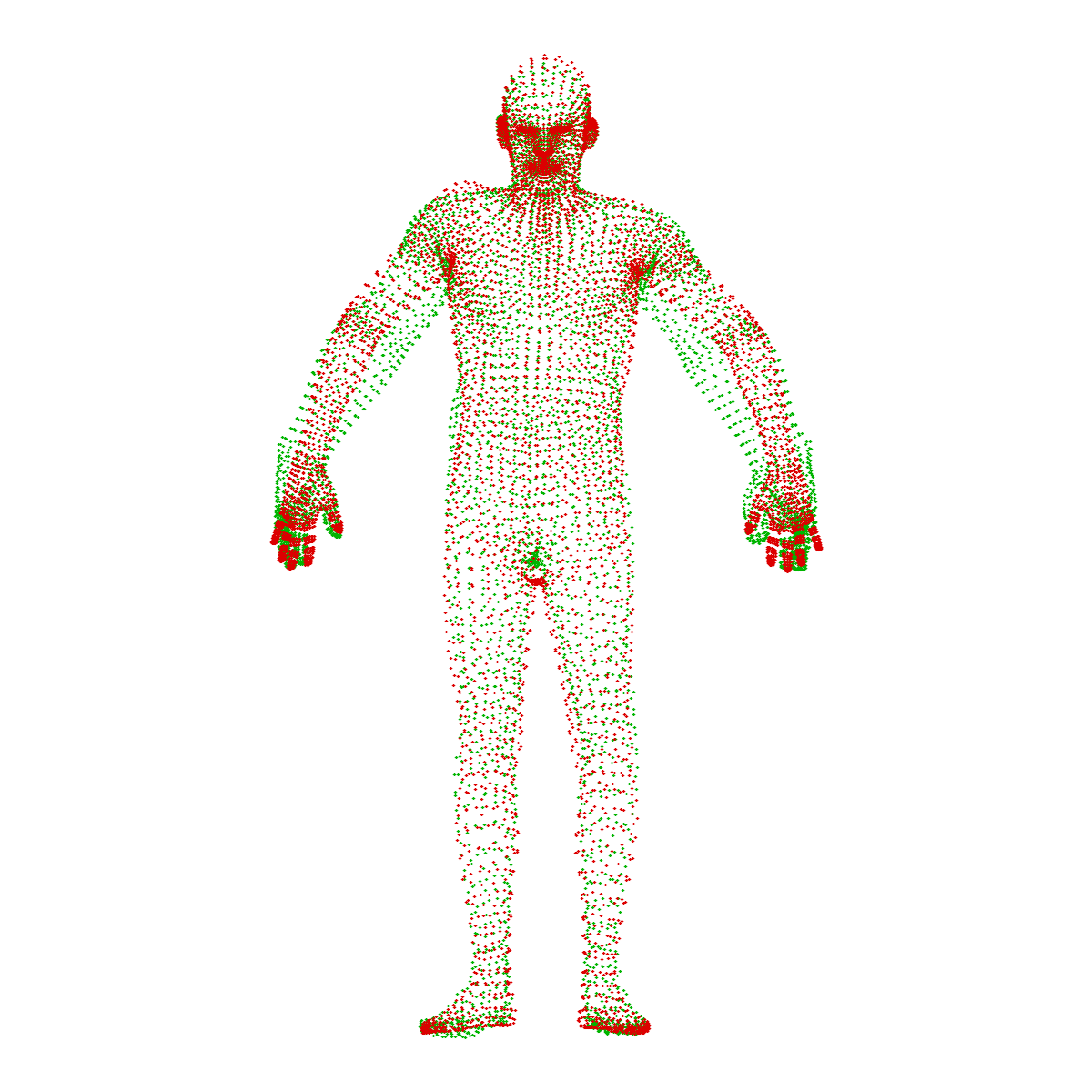}
&
\includegraphics[width=\linewidth]{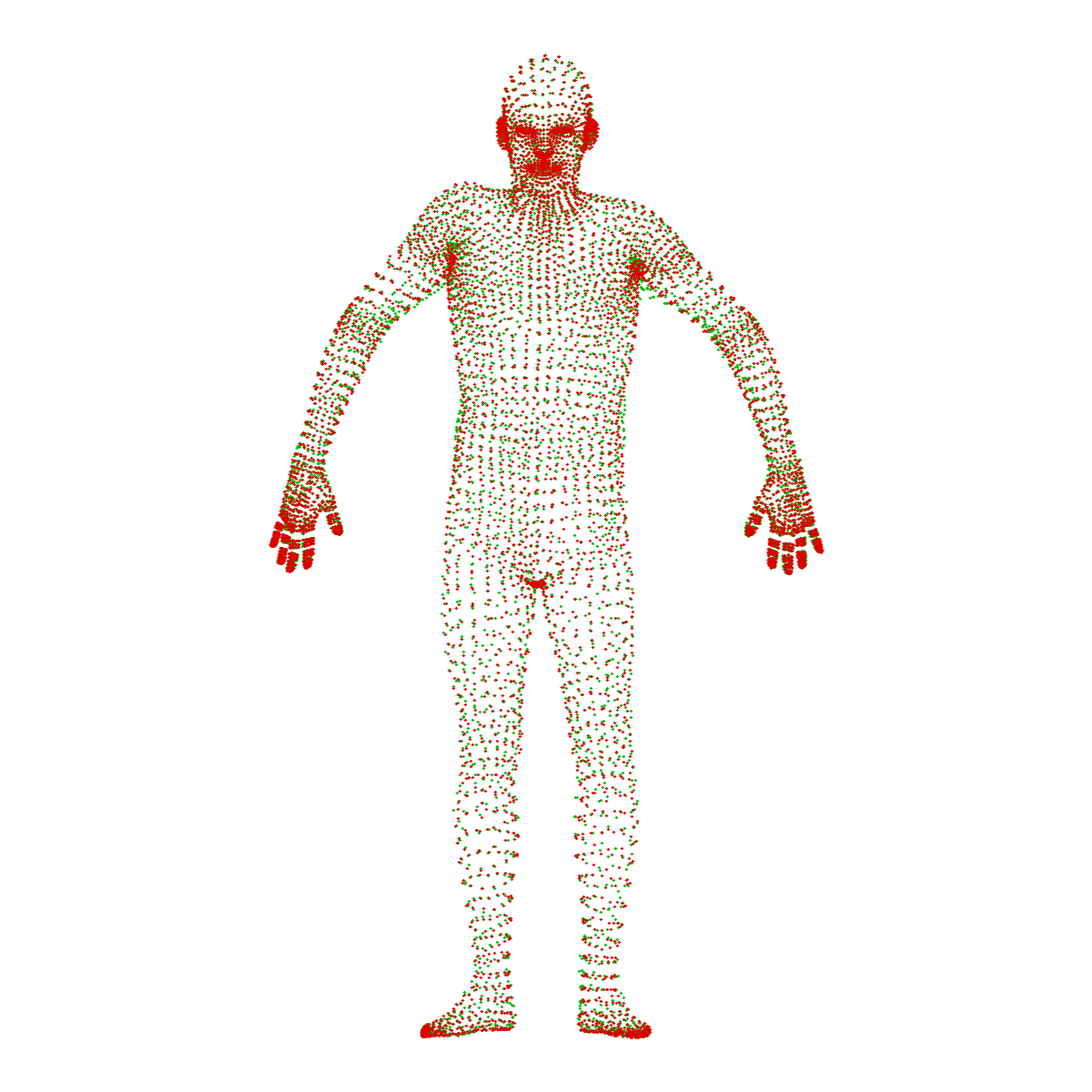}

\end{tabular}
\endgroup

\vspace{0.4em}

\caption{
Three-dimensional large-deformation registration examples using Analytic-CPD.
Red points denote the fixed point set, and green points denote the current moving point set during registration.
For each example, the registration process is shown from the initial configuration (Begin), through intermediate iterations (Iter 5, Iter 15, and Iter 30), to the final registered result (End).
The first two examples use SHREC'19 human point-cloud models with 6890 and 10050 points, respectively, and apply the smooth non-analytic deformation model described above to generate controlled large-deformation registration cases.
The third example is a real articulated human-motion registration case from the MPI-FAUST dataset.
}
\label{fig:3d-large-deformation-human-examples}
\end{figure*}

Figure~\ref{fig:3d-large-deformation-human-examples} shows the registration process from the initial configuration (Begin), through intermediate iterations (Iter 5, Iter 15, and Iter 30), to the final registered result (End). Compared with the two-dimensional analytic-deformation examples, we visualize later intermediate iterations here because three-dimensional large-deformation registration is more challenging. The first two examples involve SHREC'19 human point-cloud models under smooth but non-analytic synthetic deformations. These deformations are globally smooth but are not exactly contained in a single finite-order analytic mapping family. The MPI-FAUST example further introduces real articulated human motion, where local rotations, pose changes, non-uniform surface motion, and sampling effects create a stronger model-mismatch setting than the controlled synthetic cases.

As shown in Fig.~\ref{fig:3d-large-deformation-human-examples}, Analytic-CPD progressively reduces the geometric discrepancy between the moving and fixed point sets in all three examples. The intermediate snapshots show that the method first corrects the dominant global mismatch and then gradually refines the nonlinear deformation. In the final results, the transformed moving point set closely overlaps the fixed point set. These examples demonstrate that the proposed posterior-guided structured analytic mapping framework is not restricted to planar cases or to synthetic analytic deformations. It remains effective for moderately large three-dimensional human point clouds, including SHREC'19-based smooth non-analytic deformation examples and real articulated human-motion examples from MPI-FAUST.

\subsubsection{Comparative experiments}
\label{subsubsec:3d-comparative-experiments}

We next provide quantitative comparisons for three-dimensional registration. The evaluation is organized from representative examples to broader statistics. First, we use large-deformation examples with error--time curves to compare the convergence behavior of Analytic-CPD with CPD and ClusterReg. Second, we conduct a controlled multi-model and multi-seed statistical experiment under smooth non-analytic deformations, which is the main quantitative test of the intended target regime of Analytic-CPD. Third, we report additional statistics on registered FAUST human-shape pairs to examine the behavior of the method on
non-synthetic articulated human data.

\textbf{Deformation comparison.}

We next compare Analytic-CPD with CPD, ClusterReg, and BCPD on three representative three-dimensional registration tasks. The first two examples are generated by the smooth \(C^\infty\) but non-analytic deformation model described above, and the third example is a human-motion registration case from the FAUST dataset. These cases evaluate whether Analytic-CPD remains effective beyond the model-matched finite analytic setting.

\begin{figure}[t]
\centering
\renewcommand{\arraystretch}{1.10}
\setlength{\tabcolsep}{0pt}

\newcommand{\figcell}[2]{%
\begin{minipage}[t]{0.47\columnwidth}
\centering
\includegraphics[width=\linewidth]{#1}\\[-0.25em]
{\scriptsize #2}
\end{minipage}
}

\begin{tabular}{@{}c@{\hspace{0.02\columnwidth}}c@{}}
\toprule
\textbf{\small Initial configuration}
&
\textbf{\small Error--time curves}
\\
\midrule

% ---------------- Group 1 ----------------
\figcell{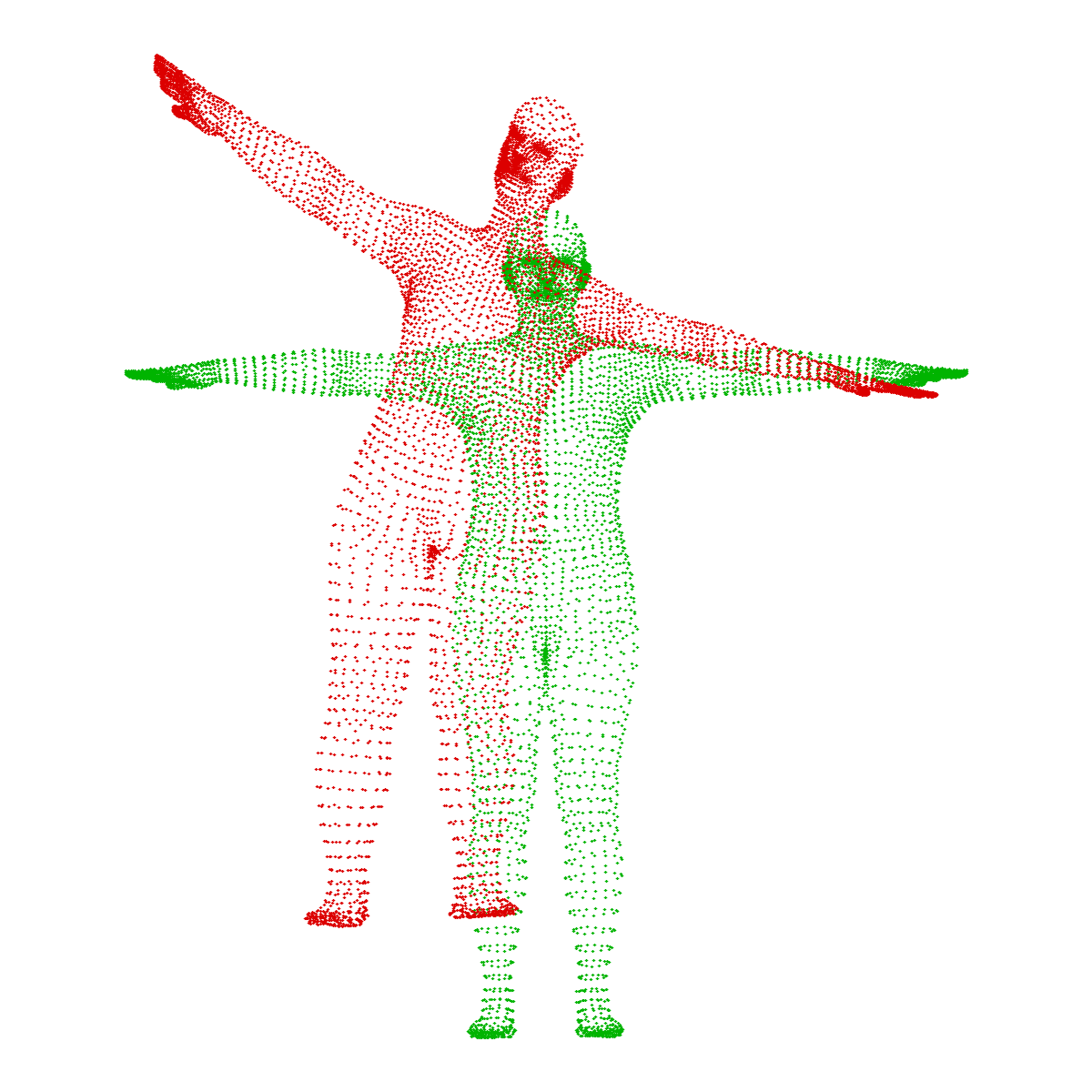}
{(a) Initial position of Example 1\\
(SHREC'19 human point cloud, 6890 points)}
&
\figcell{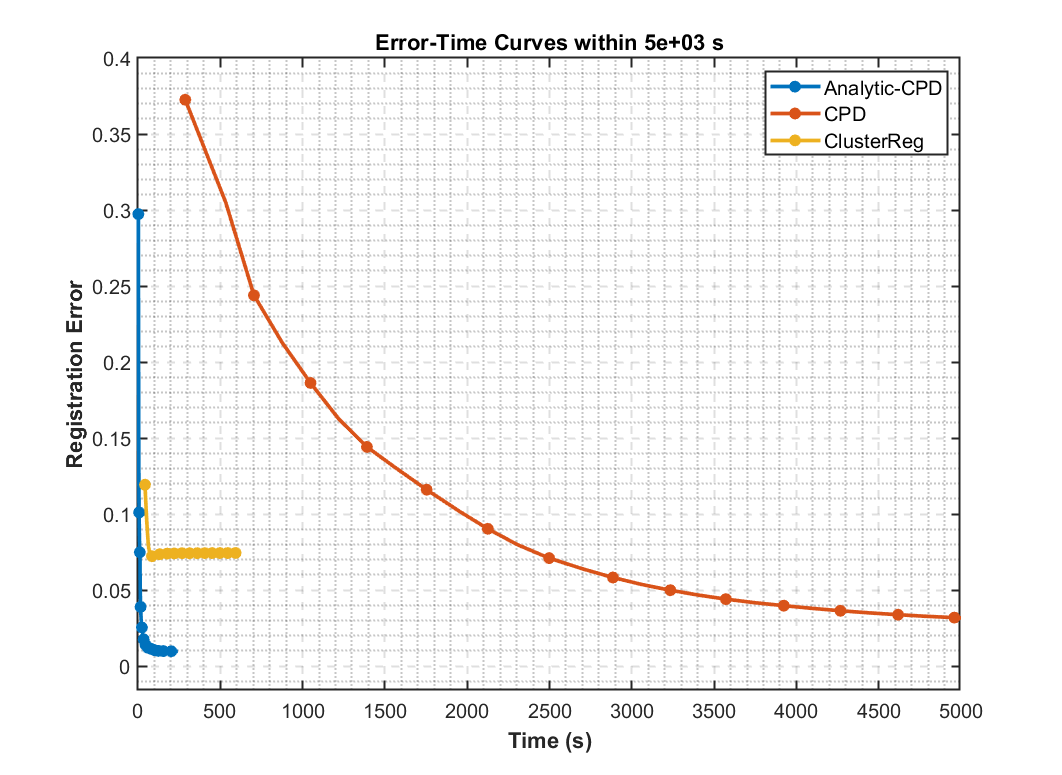}
{(b) Error--time comparison of Example 1\\
(Analytic-CPD, CPD, and ClusterReg)}
\\[0.6em]

% ---------------- Group 2 ----------------
\figcell{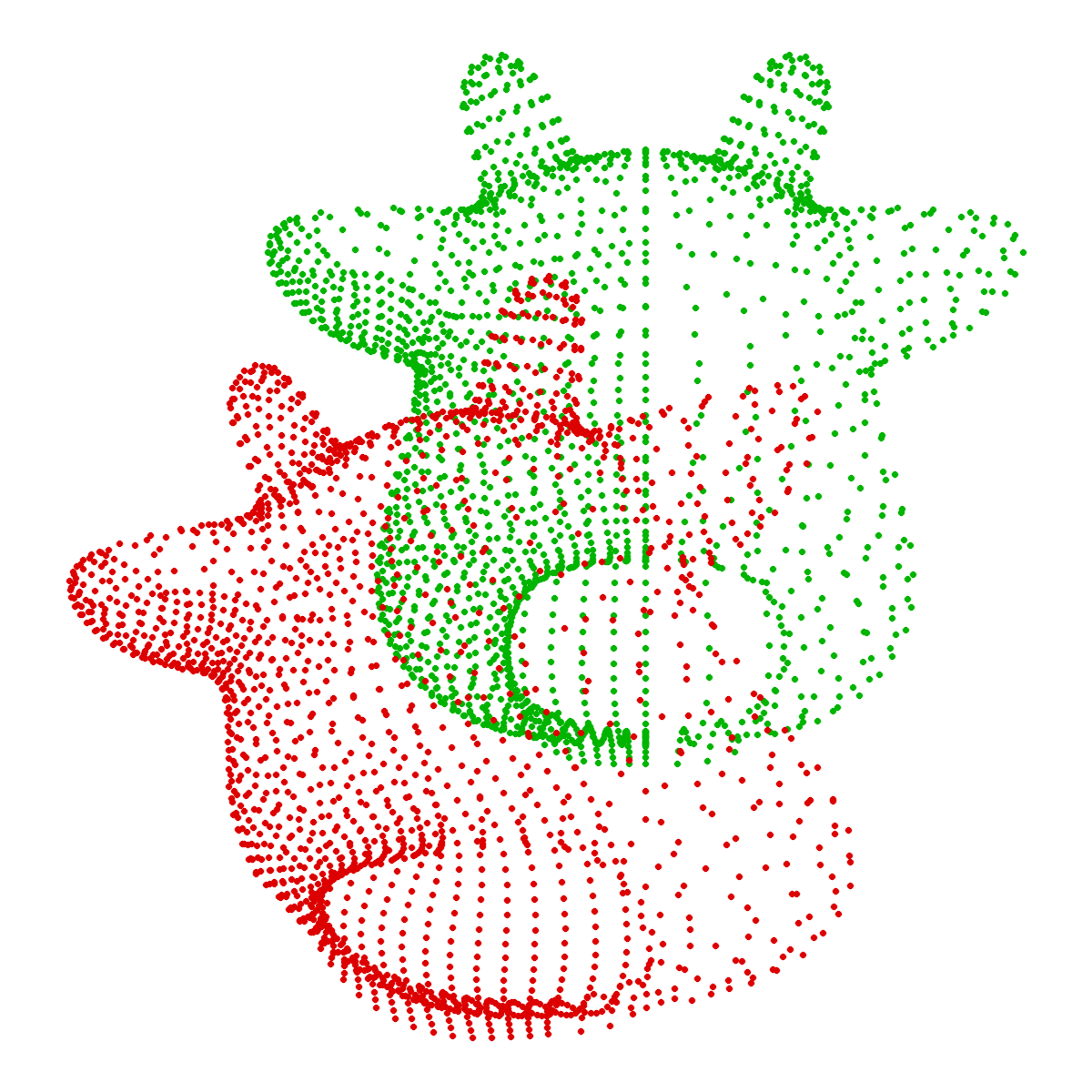}
{(c) Initial position of Example 2\\
(Cowhead point cloud, 2036 points)}
&
\figcell{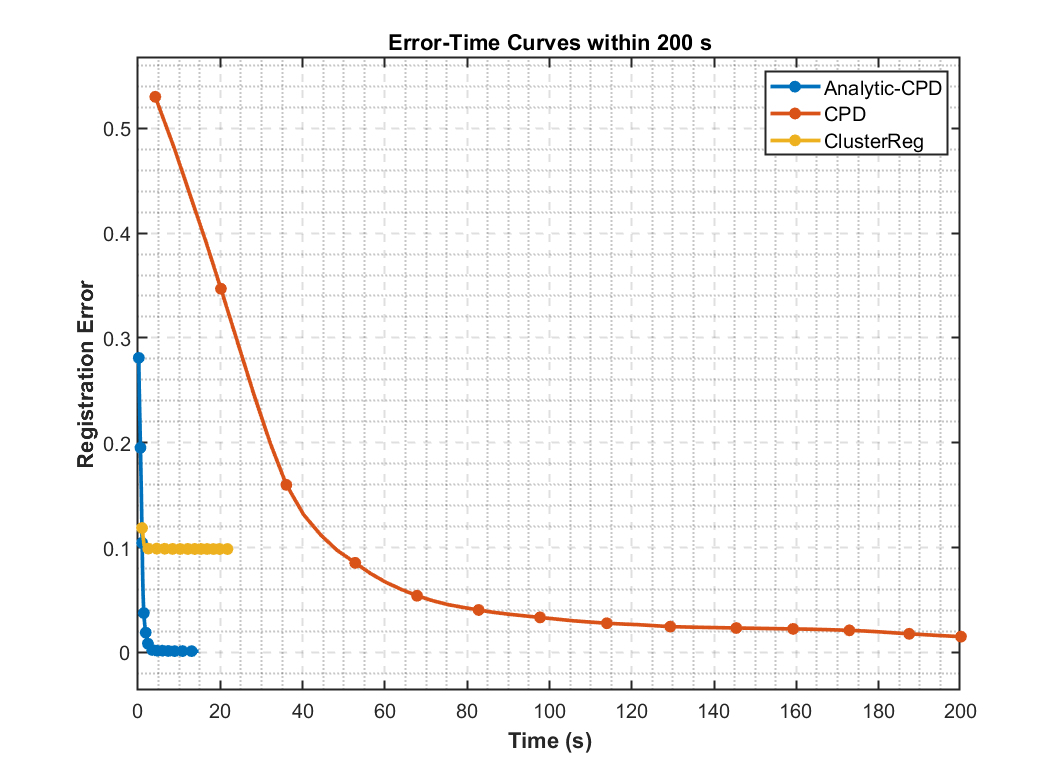}
{(d) Error--time comparison of Example 2\\
(Analytic-CPD, CPD, and ClusterReg)}
\\[0.6em]

% ---------------- Group 3 ----------------
\figcell{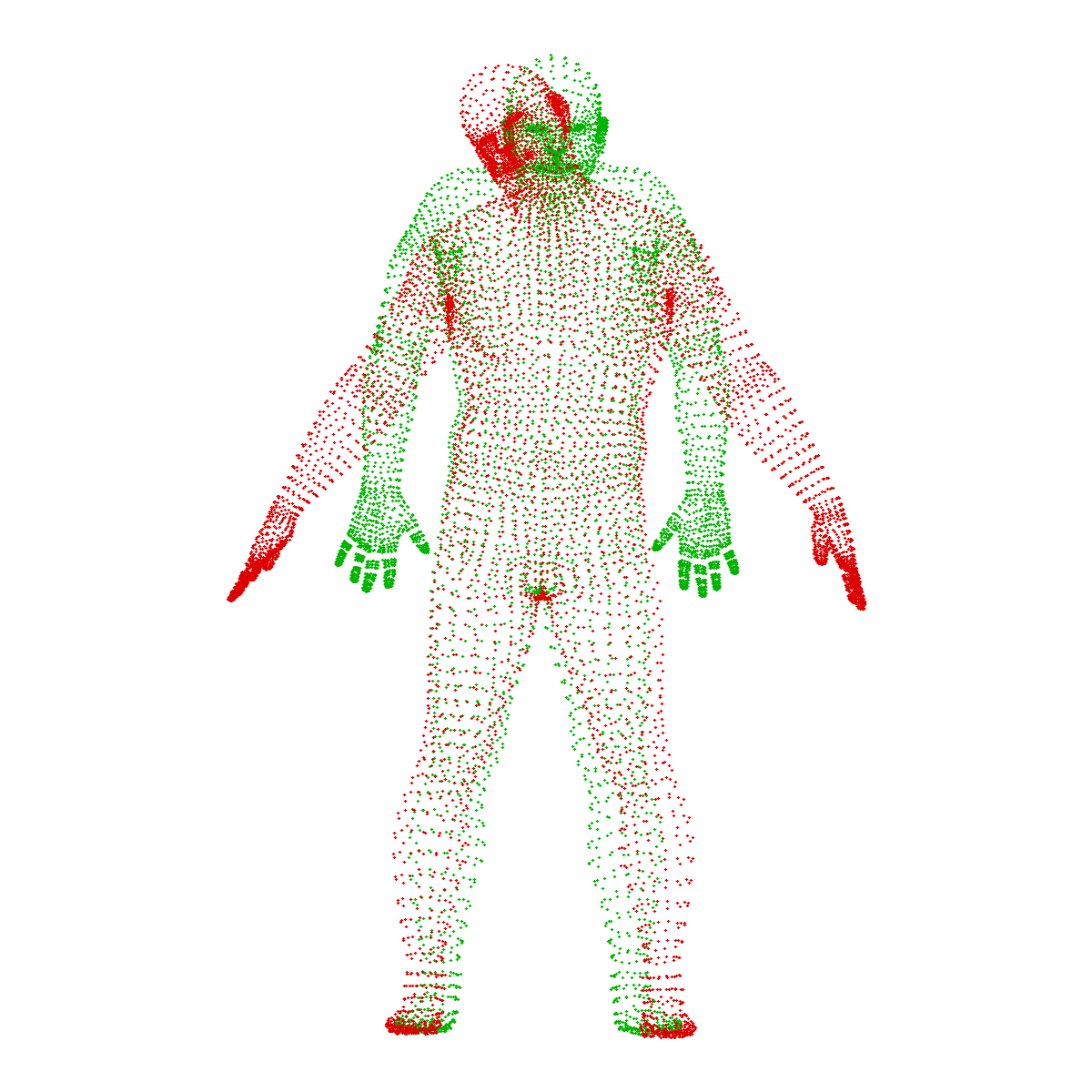}
{(e) Initial position of Example 3\\
(FAUST human motion, 6890 points)}
&
\figcell{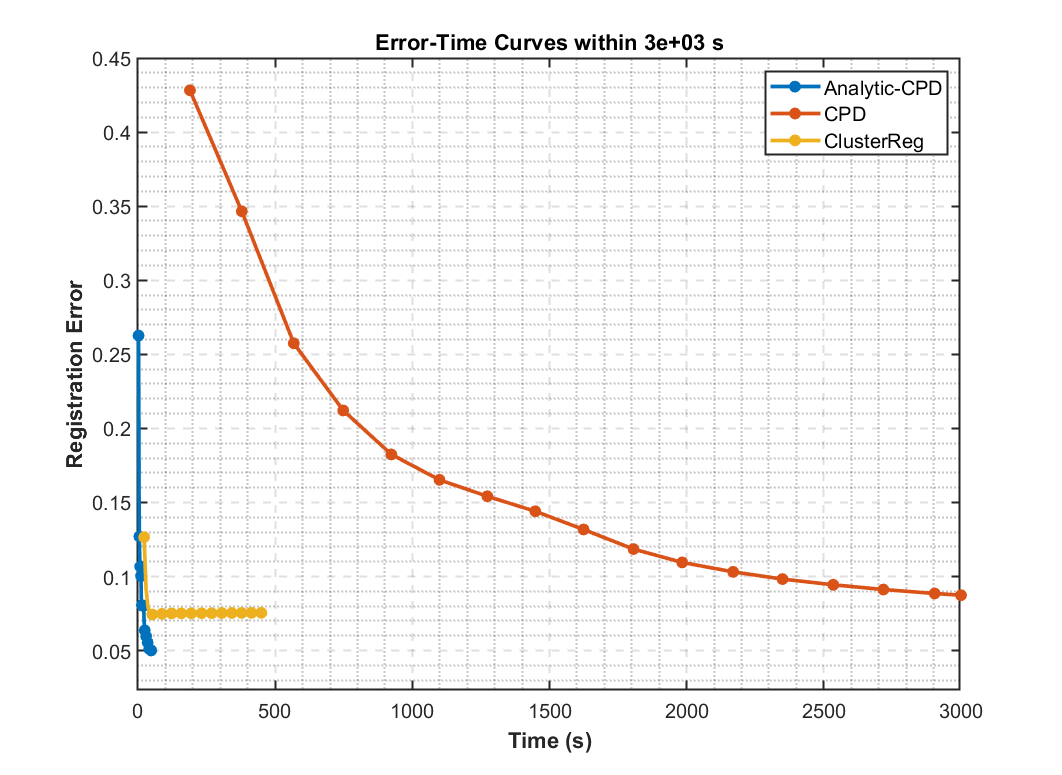}
{(f) Error--time comparison of Example 3\\
(Analytic-CPD, CPD, and ClusterReg)}
\\

\bottomrule
\end{tabular}

\caption{
Three-dimensional registration under large smooth deformation and articulated human motion. The left column shows the initial relative positions of the moving and fixed point clouds, and the right column shows the corresponding error--time curves for Analytic-CPD, CPD, and ClusterReg. BCPD is evaluated from its final exported result and is reported in Table~\ref{tab:3d-deformation-comparison}. The first two rows show smooth non-analytic deformation examples, and the third row shows a FAUST human-motion registration example.
}
\label{fig:3d-initial-and-error-time-comparison}
\end{figure}

Figure~\ref{fig:3d-initial-and-error-time-comparison} shows the initial configurations and the corresponding error--time curves for Analytic-CPD, CPD, and ClusterReg. BCPD is evaluated using its official command-line implementation; since its per-iteration external RMSE curve is not directly available, only its final error and runtime are reported in Table~\ref{tab:3d-deformation-comparison}. All final errors are recomputed from the registered point sets using the same external RMSE metric.

\begin{table}[t]
\centering
\caption{
Three-dimensional large-deformation and human-motion registration comparison.
Errors are measured in the normalized coordinate system.
}
\label{tab:3d-deformation-comparison}

\begin{tabular*}{\linewidth}{@{\extracolsep{\fill}}lcc@{}}
\toprule
Method & Final error & Time (s) \\
\midrule

\multicolumn{3}{@{}l}{\textbf{Example 1: SHREC'19 human, 6890 points}} \\
Analytic-CPD & \(9.67\times 10^{-3}\) & \(242.09\) \\
CPD          & \(1.70\times 10^{-2}\) & \(23581.80\) \\
ClusterReg   & \(7.44\times 10^{-2}\) & \(604.10\) \\
BCPD         & \(2.70\times 10^{-2}\) & \(220.00\)  \\

\midrule
\multicolumn{3}{@{}l}{\textbf{Example 2: Cowhead, 2036 points}} \\
Analytic-CPD & \(1.15\times 10^{-3}\) & \(14.72\) \\
CPD          & \(1.37\times 10^{-2}\) & \(265.44\) \\
ClusterReg   & \(9.86\times 10^{-2}\) & \(22.41\) \\
BCPD         & \(3.39\times 10^{-2}\) & \(11.94\) \\

\midrule
\multicolumn{3}{@{}l}{\textbf{Example 3: FAUST human motion, 6890 points}} \\
Analytic-CPD & \(5.02\times 10^{-2}\) & \(53.43\) \\
CPD          & \(7.72\times 10^{-2}\) & \(13128.39\) \\
ClusterReg   & \(7.55\times 10^{-2}\) & \(458.49\) \\
BCPD         & \(8.03\times 10^{-2}\) & \(232.04\) \\

\bottomrule
\end{tabular*}
\end{table}

Analytic-CPD achieves the lowest final error in all three examples. In Example~1, it reduces the CPD error from \(1.70\times 10^{-2}\) to \(9.67\times 10^{-3}\), while reducing the runtime from \(23581.8\)~s to \(242.1\)~s. BCPD is slightly faster in this official-implementation reference, but its final error is noticeably higher. ClusterReg is also less accurate on
this example.

In Example~2, Analytic-CPD reaches a final error of \(1.15\times 10^{-3}\), which is about one order of magnitude lower than CPD and clearly lower than both BCPD and ClusterReg. BCPD and ClusterReg are faster on this smaller point cloud, but their residuals are substantially larger. In the FAUST example,
where the deformation is no longer generated by the controlled smooth non-analytic model, Analytic-CPD still obtains the lowest final error, \(5.02\times 10^{-2}\), and is much faster than CPD, ClusterReg, and BCPD.

Overall, these experiments indicate that the proposed posterior-guided analytic M-step is not limited to exactly analytic ground-truth deformations. Even when the deformation lies outside the finite Taylor family, Analytic-CPD provides an effective smooth finite-dimensional approximation. Compared with standard CPD,
it consistently improves final accuracy and greatly reduces runtime in the two larger examples. Compared with BCPD and ClusterReg, it provides better final registration accuracy while maintaining competitive runtime.

\subsubsection{Multi-model and multi-seed statistics under smooth non-analytic deformations}
\label{subsubsec:multi-model-multi-seed}

To evaluate whether the advantage of Analytic-CPD generalizes beyond individual examples, we conduct a multi-model and multi-seed experiment under smooth non-analytic deformations. This experiment serves as the main quantitative evidence for the intended large smooth-deformation regime of Analytic-CPD.
Because the deformation mechanism is controlled and pointwise correspondences are available, final errors can be evaluated against clean deformed targets. Compared with articulated human-motion benchmarks, this setting more directly isolates the ability of a registration method to recover large smooth
deformations under model mismatch.

\begin{figure}[t]
\centering

% ---------- legend ----------
\begingroup
\setlength{\tabcolsep}{0.6em}
\begin{tabular}{cc}
\textcolor{red}{\large$\bullet$}\ \ Red = fixed
&
\textcolor{green!70!black}{\large$\bullet$}\ \ Green = moving
\end{tabular}
\endgroup

\vspace{0.5em}

% ---------- local commands ----------
\begingroup
\def\initcell#1{%
\includegraphics[width=\linewidth]{#1}%
}
\def\rowlabel#1{%
\scriptsize\bfseries\shortstack[c]{#1}%
}

% ---------- main table ----------
\setlength{\tabcolsep}{1.0pt}
\renewcommand{\arraystretch}{1.05}

\begin{tabular}{
    @{}
    >{\centering\arraybackslash}m{0.155\linewidth}
    *{5}{>{\centering\arraybackslash}m{0.150\linewidth}}
    @{}
}
\toprule
\scriptsize\shortstack[c]{Dataset\\Seed}
&
\scriptsize Seed 1
&
\scriptsize Seed 2
&
\scriptsize Seed 3
&
\scriptsize Seed 4
&
\scriptsize Seed 5
\\
\midrule

\rowlabel{FAUST\\tr\_reg\_020}
&
\initcell{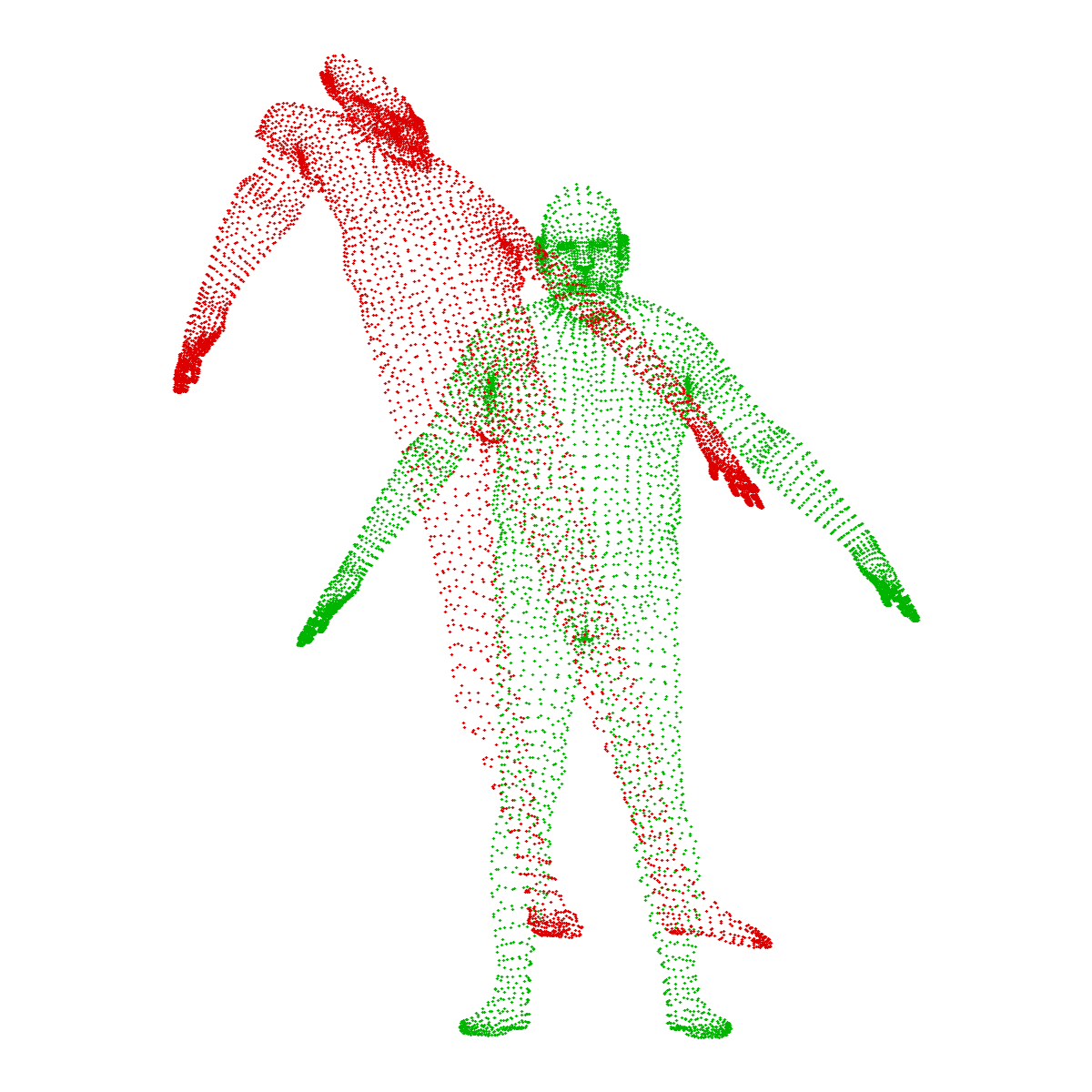}
&
\initcell{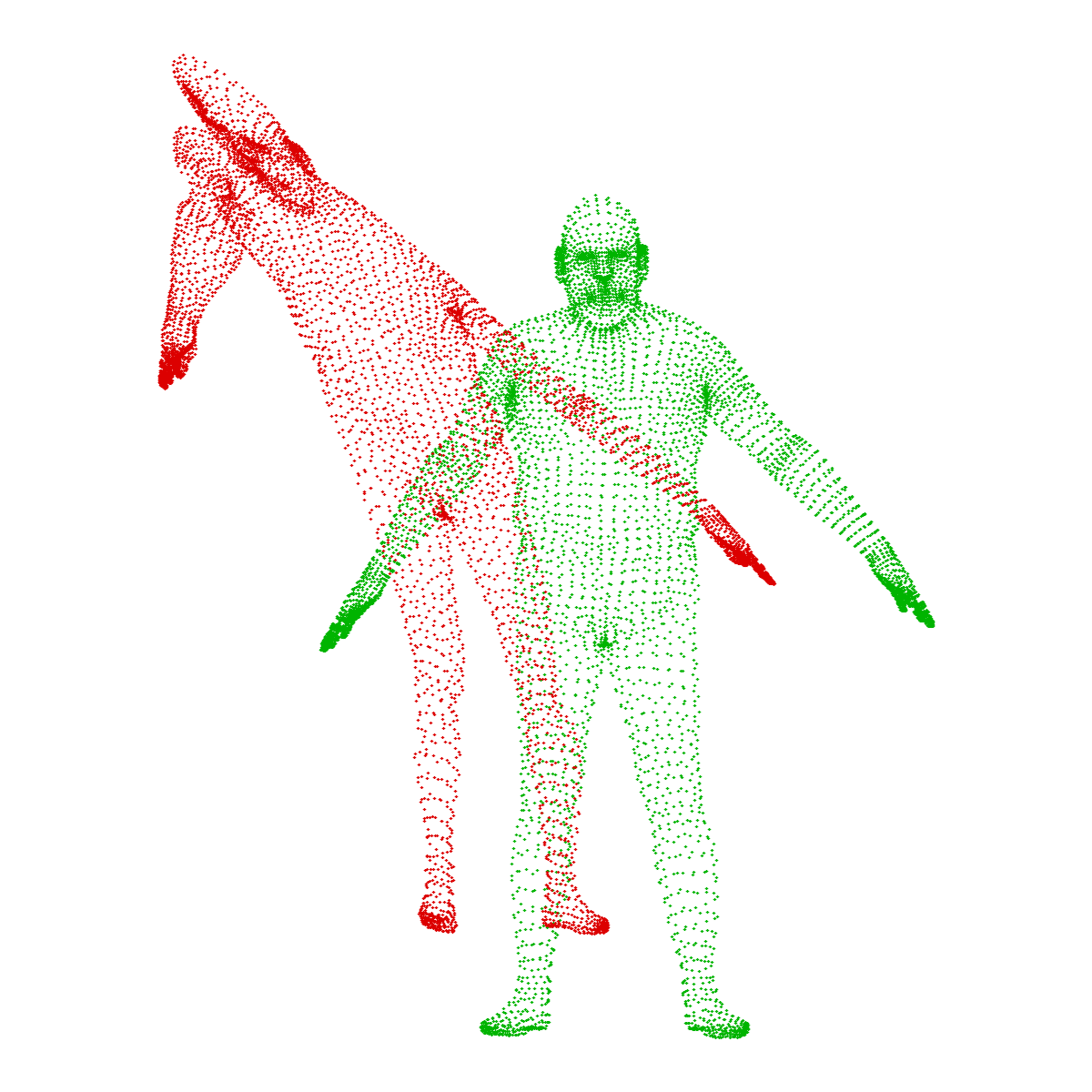}
&
\initcell{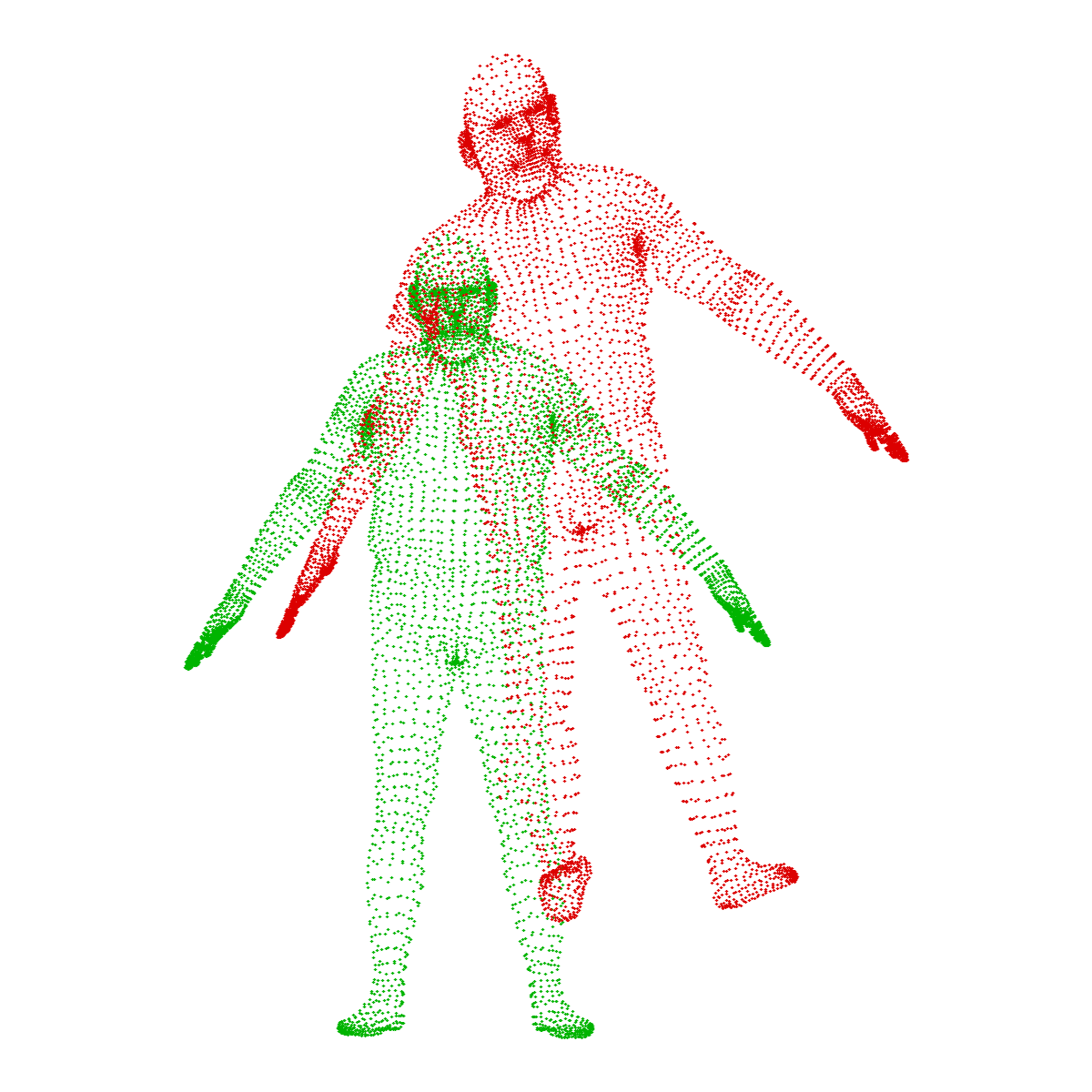}
&
\initcell{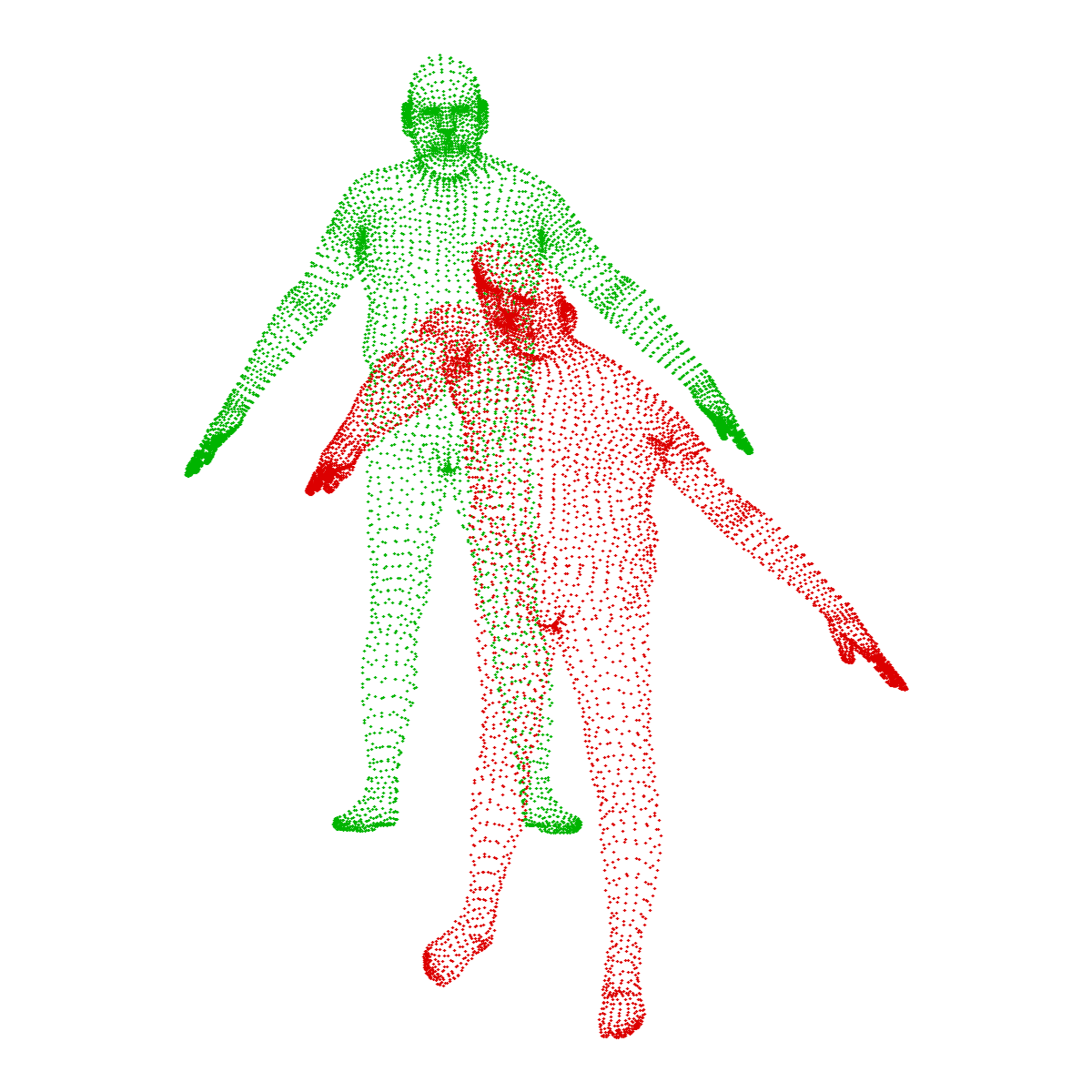}
&
\initcell{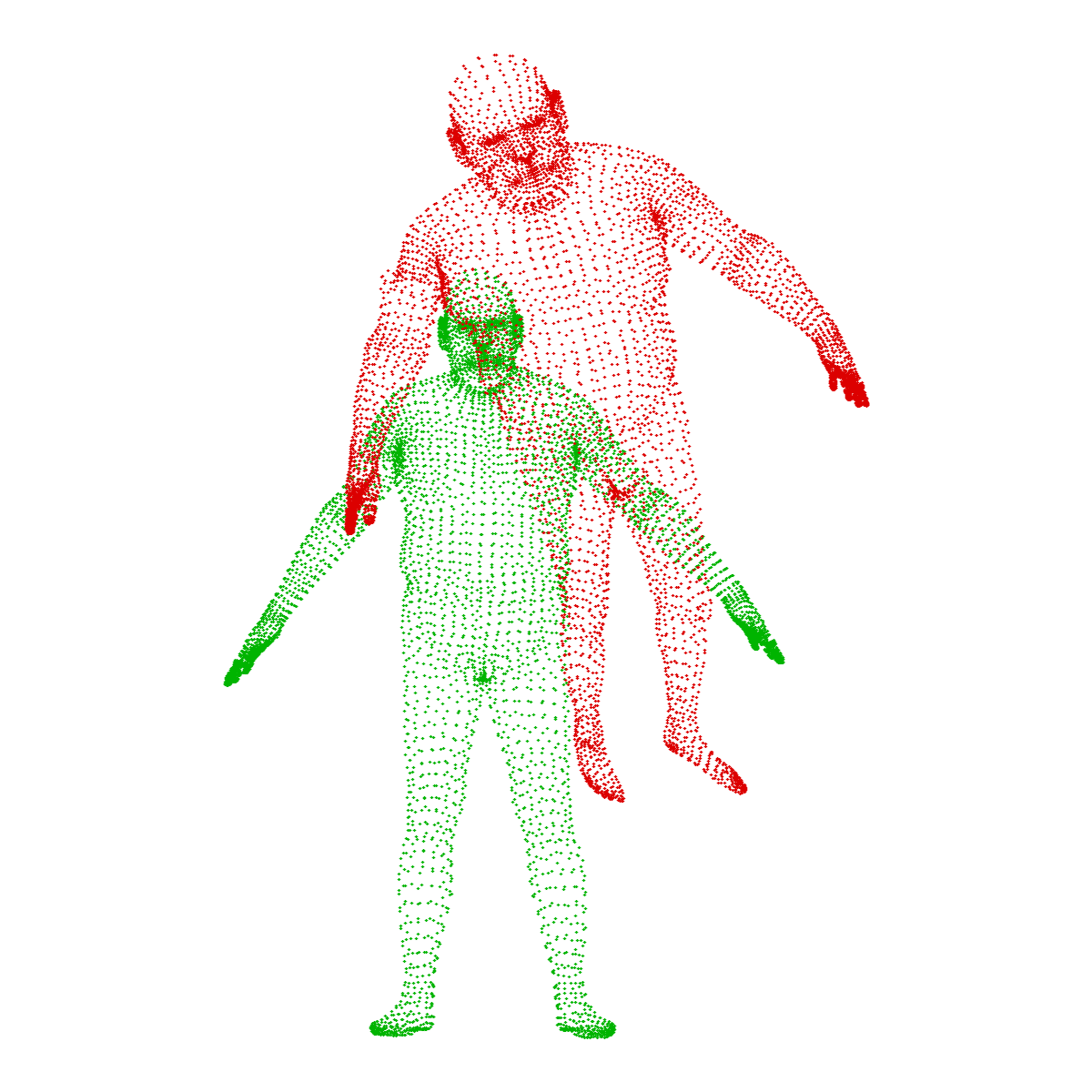}
\\[0.35em]

\rowlabel{FAUST\\tr\_reg\_038}
&
\initcell{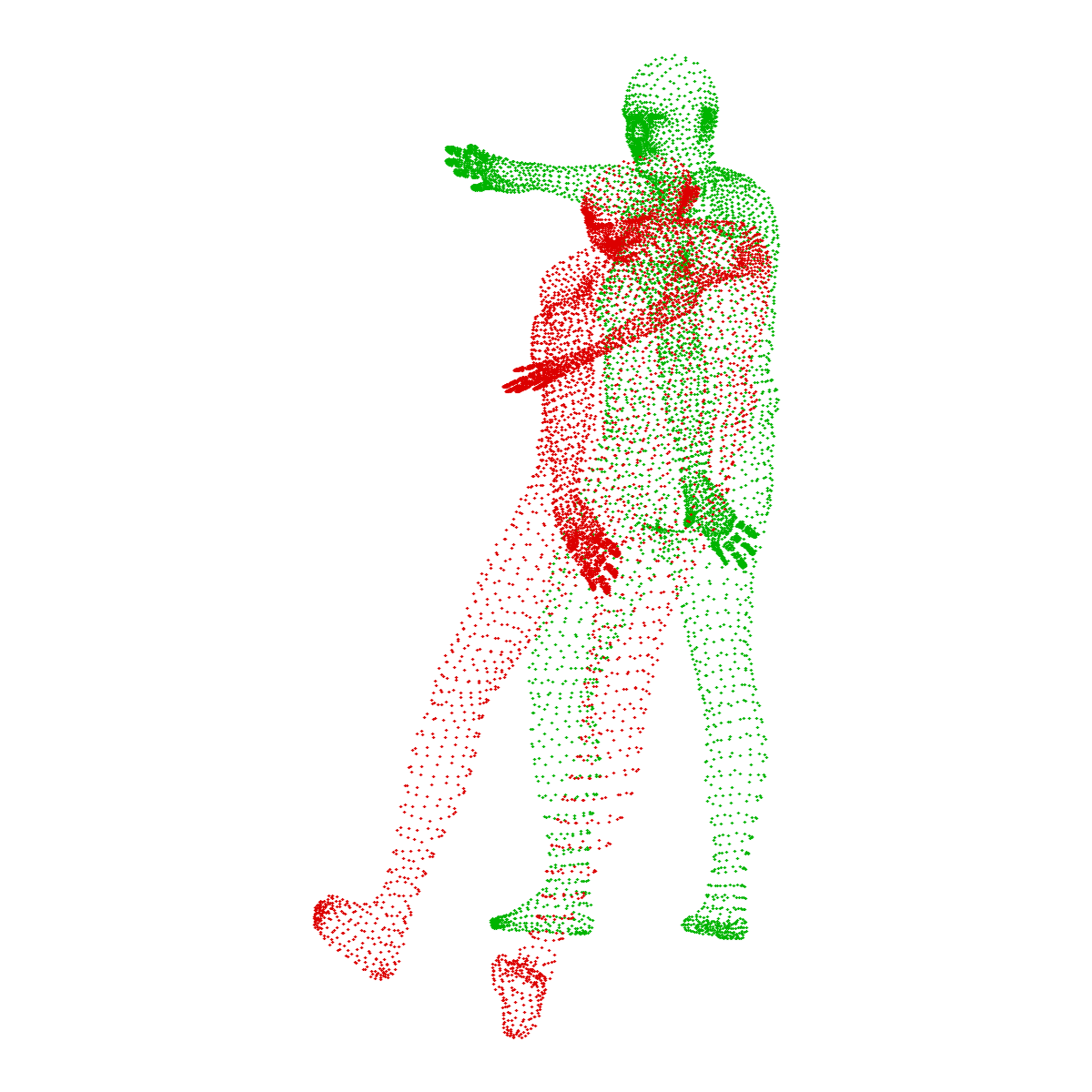}
&
\initcell{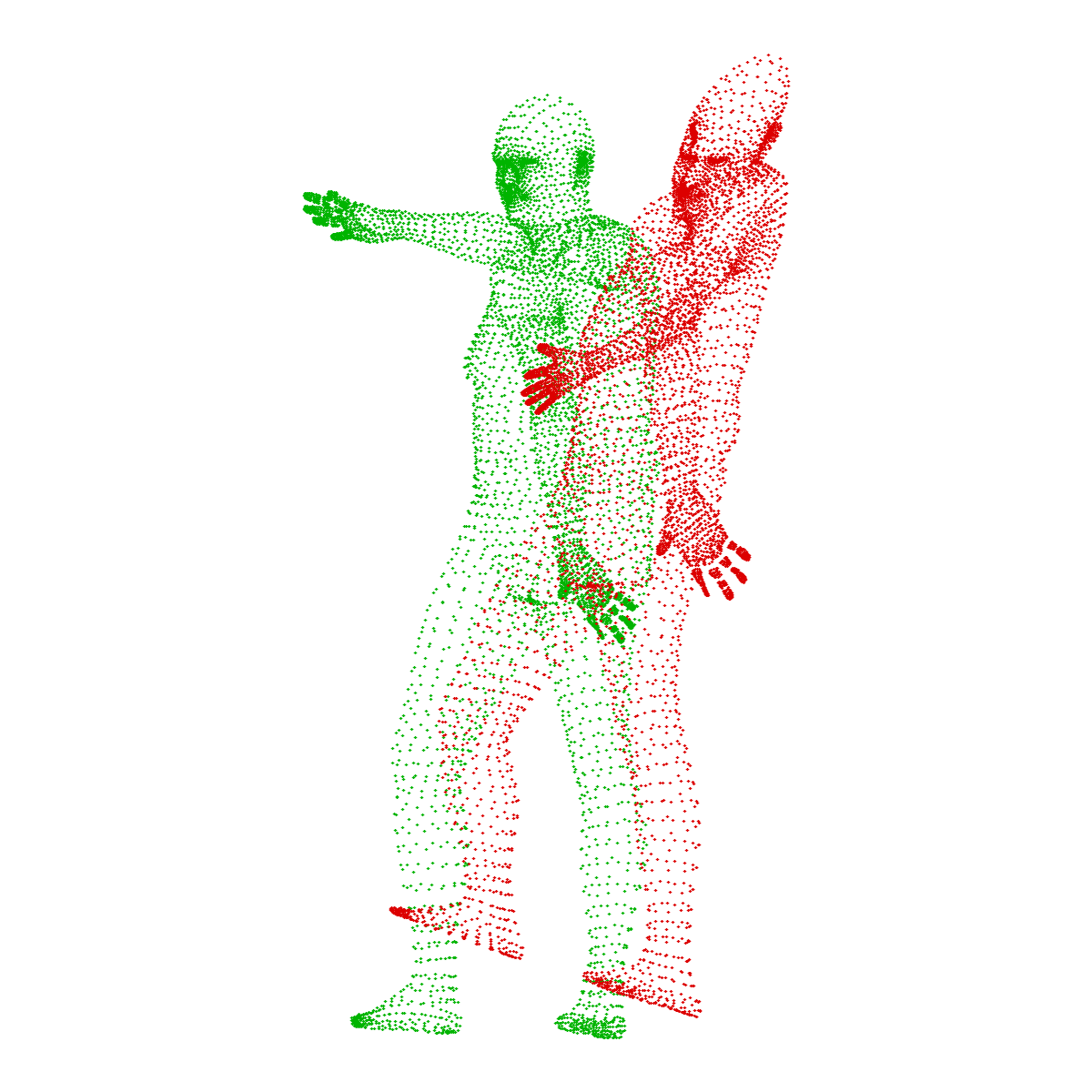}
&
\initcell{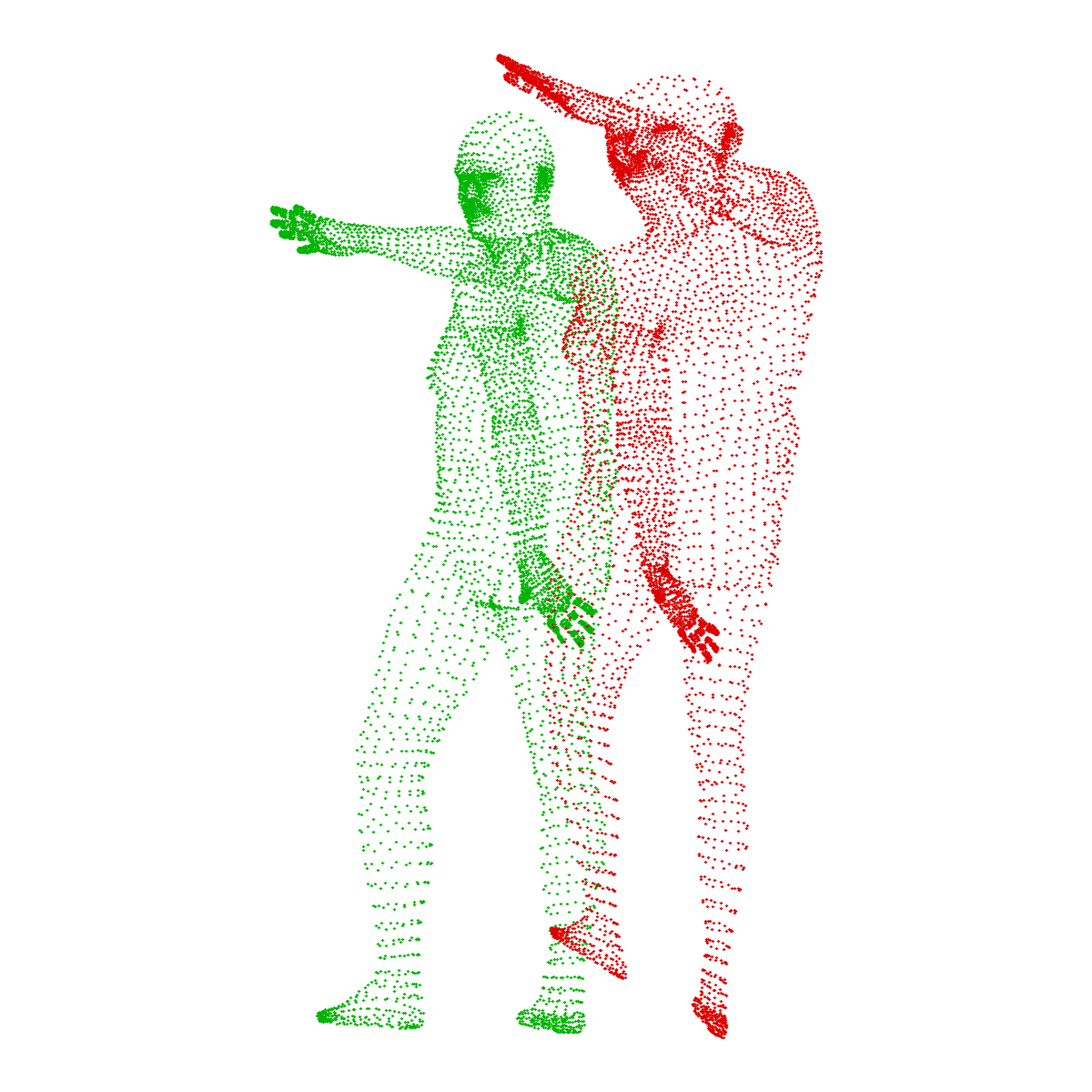}
&
\initcell{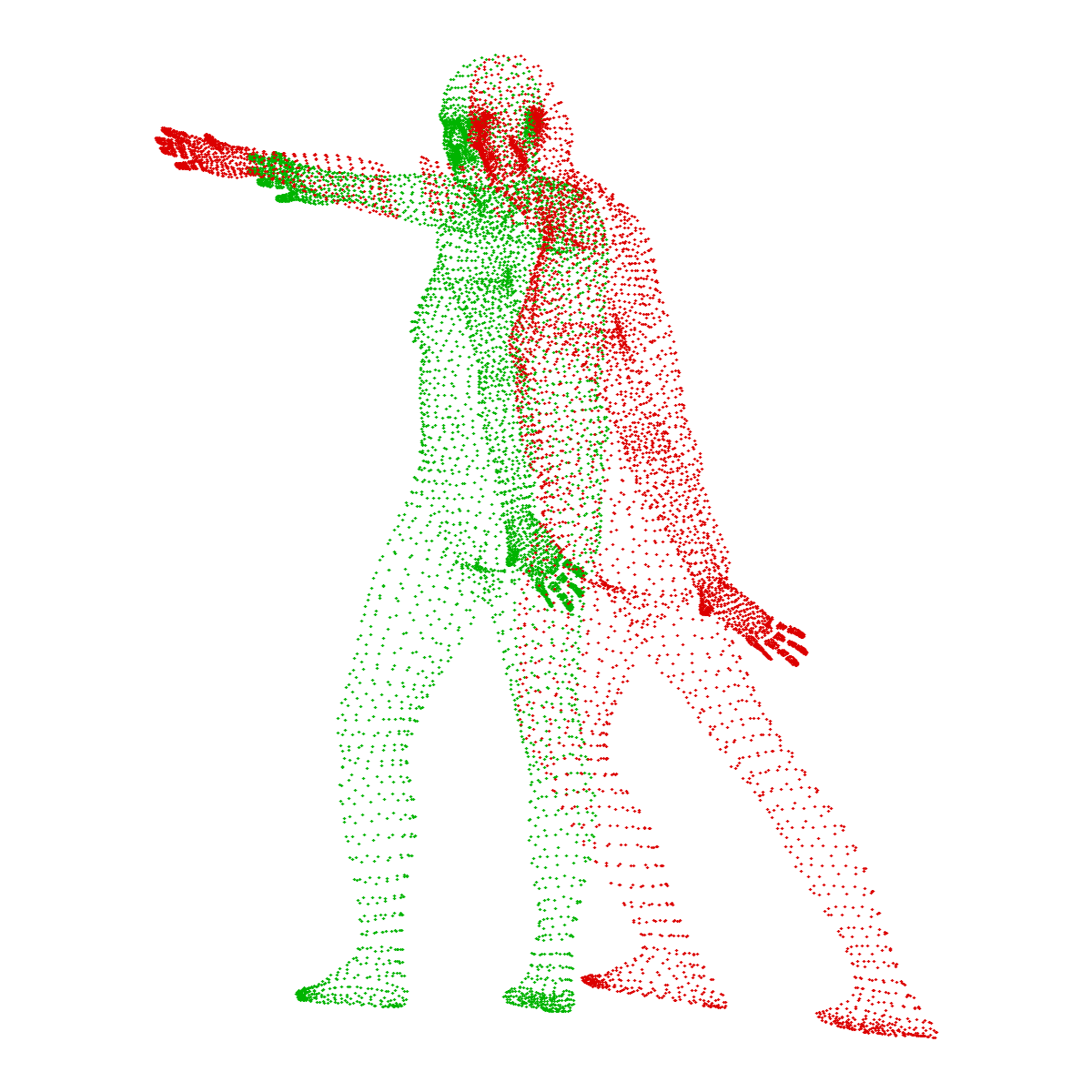}
&
\initcell{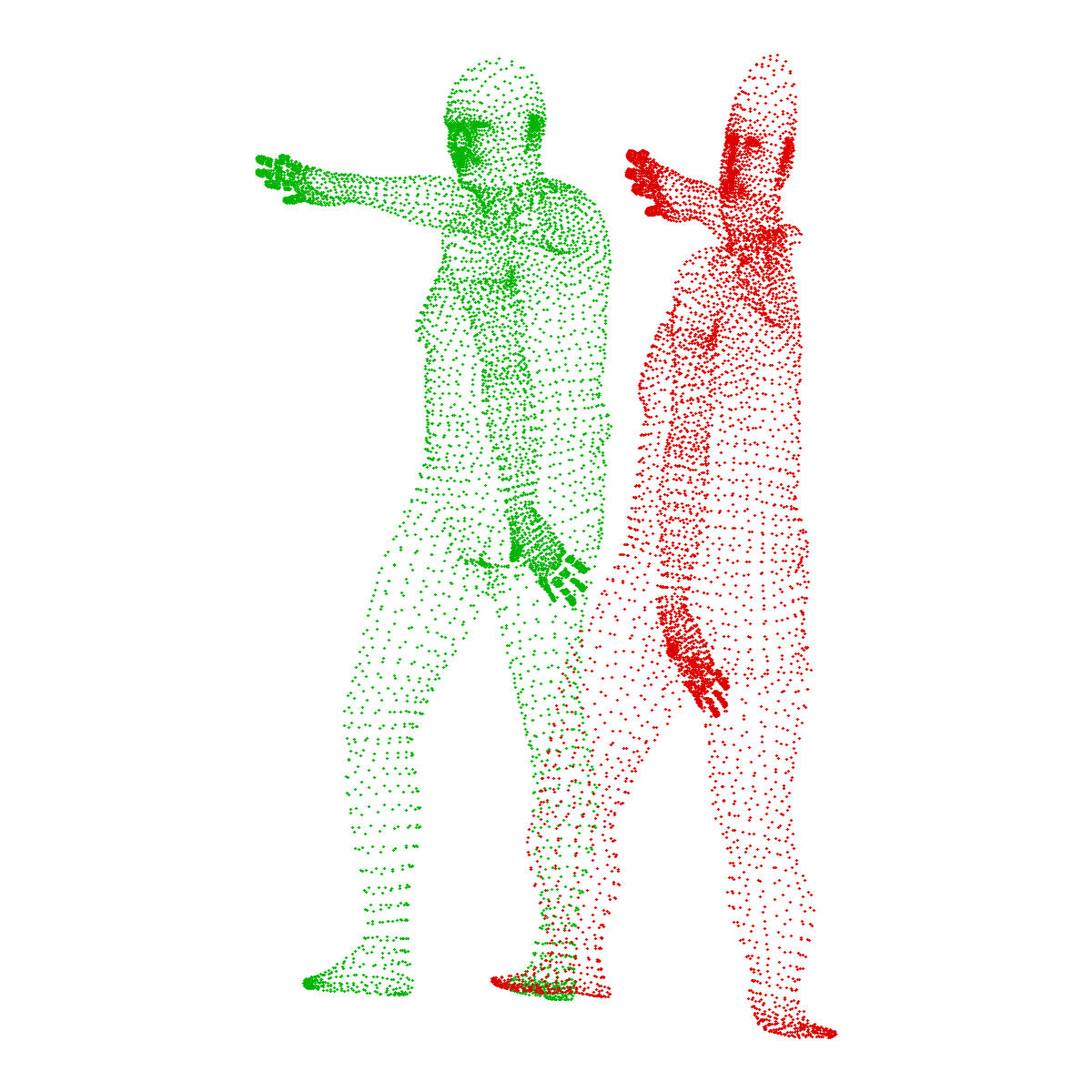}
\\[0.35em]

\rowlabel{Stanford\\Bunny}
&
\initcell{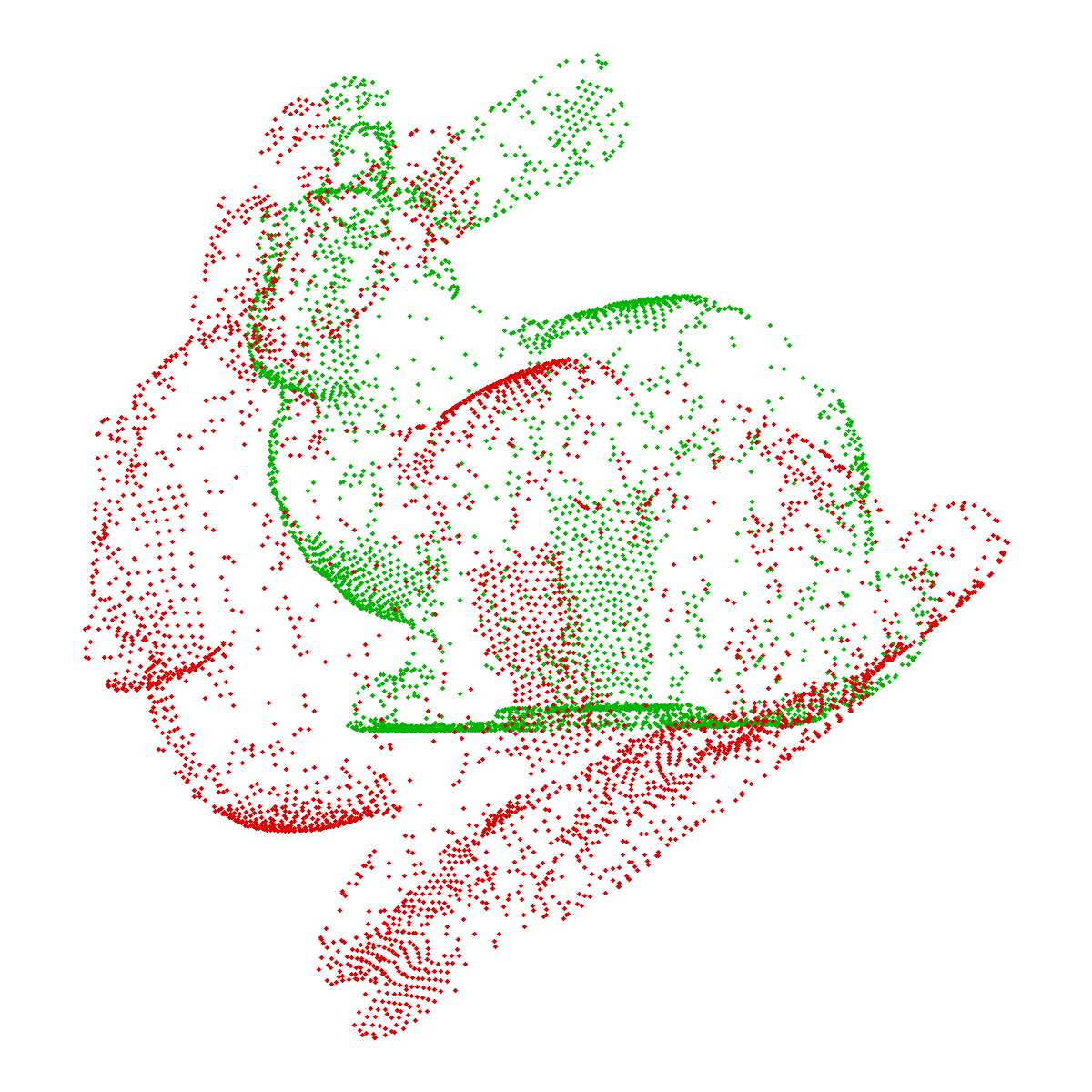}
&
\initcell{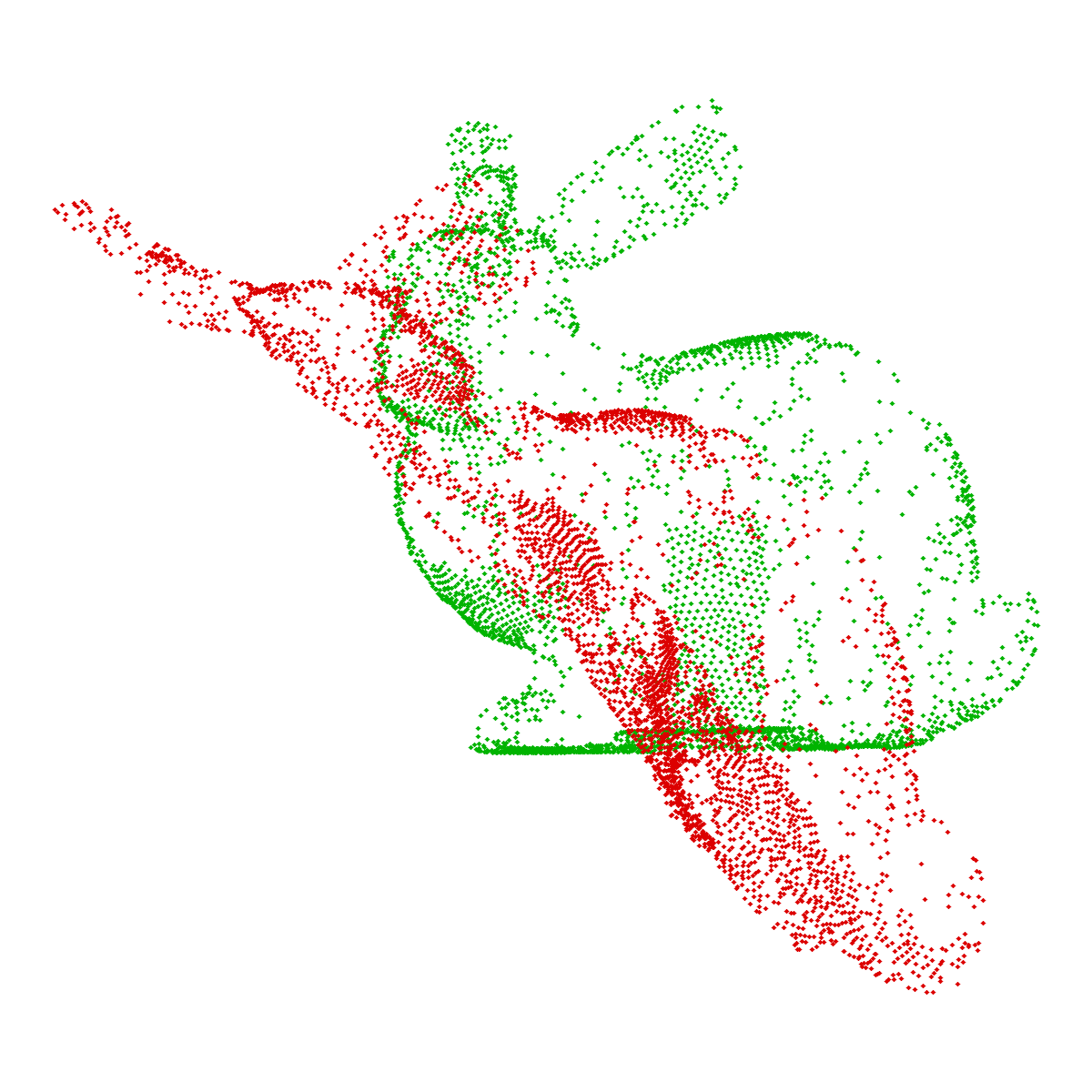}
&
\initcell{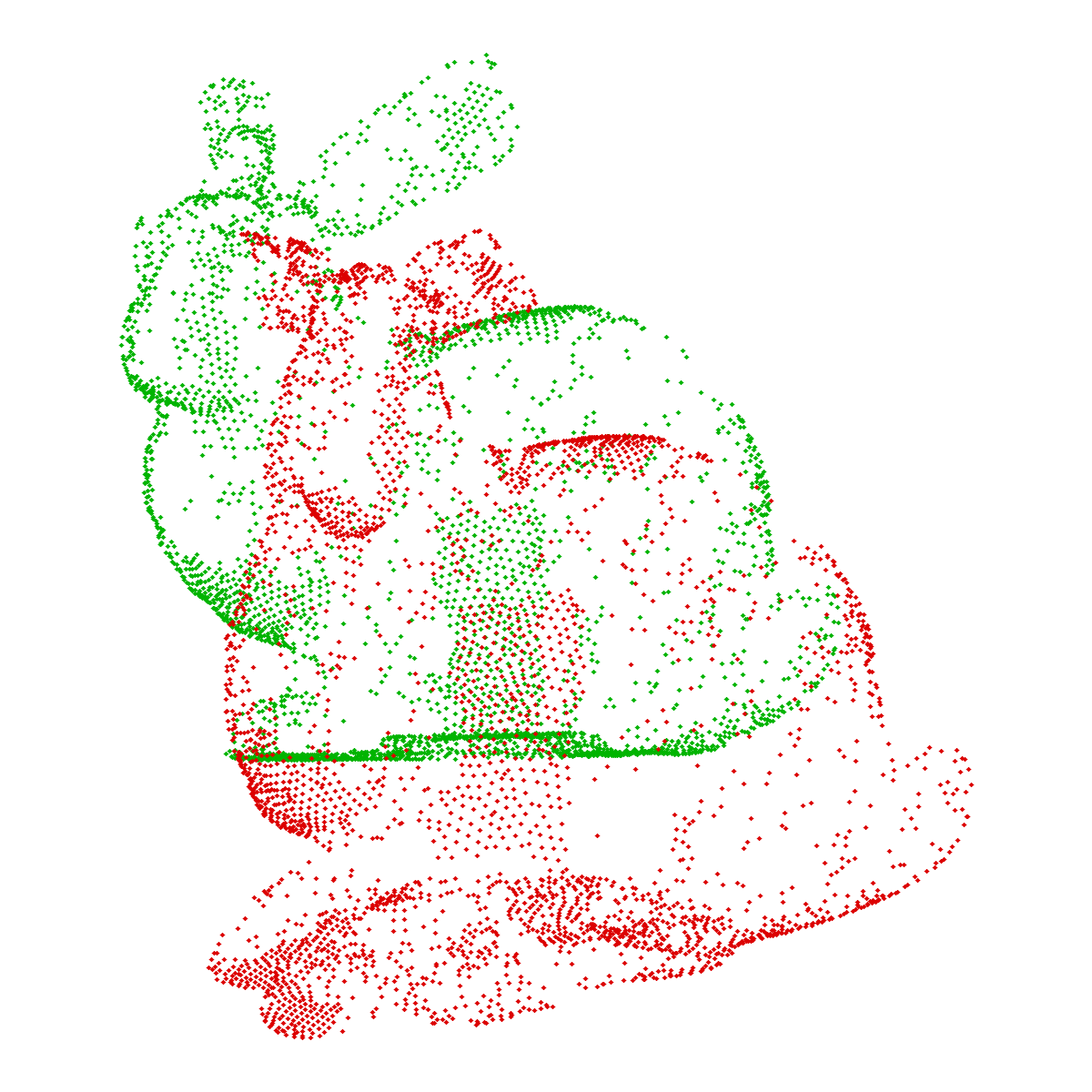}
&
\initcell{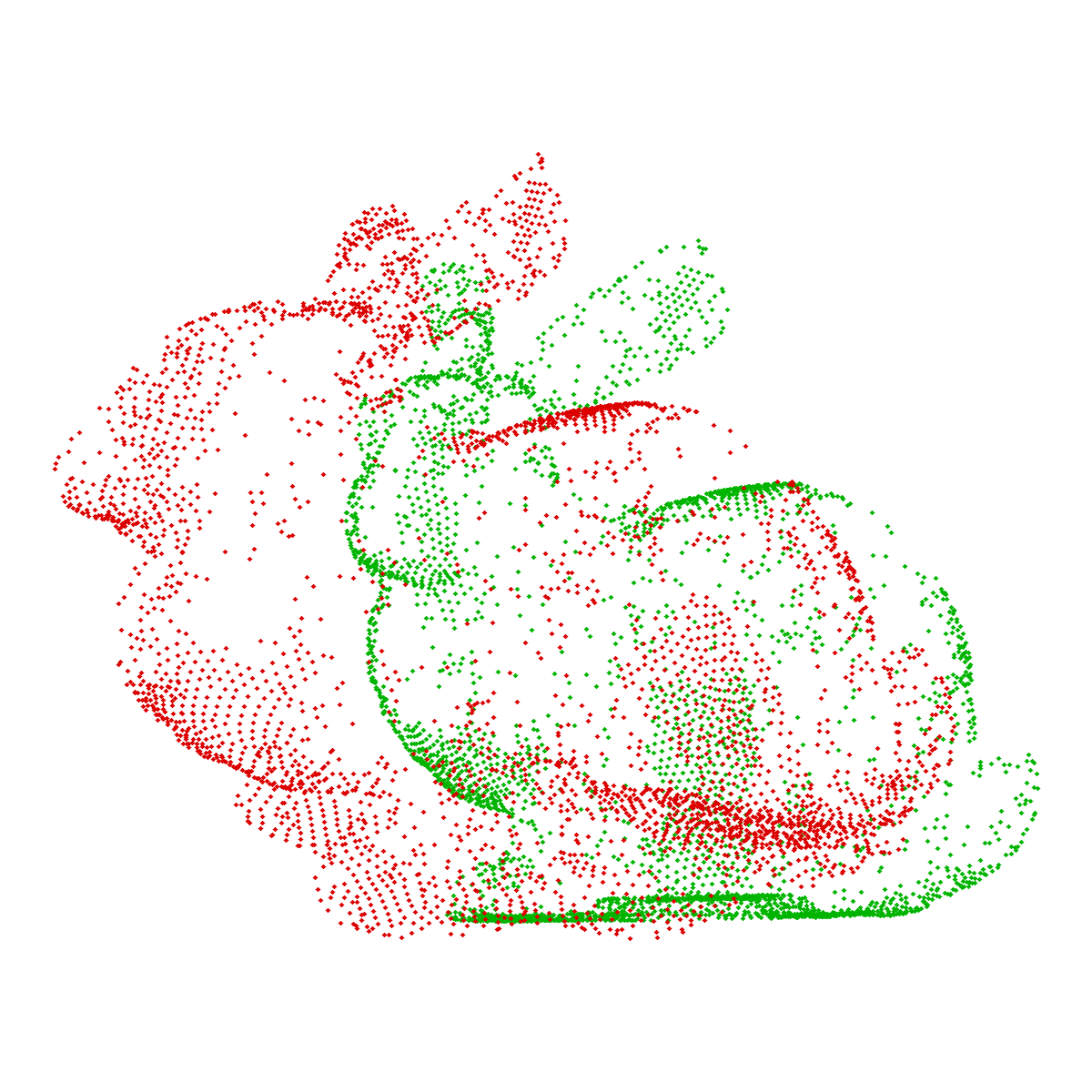}
&
\initcell{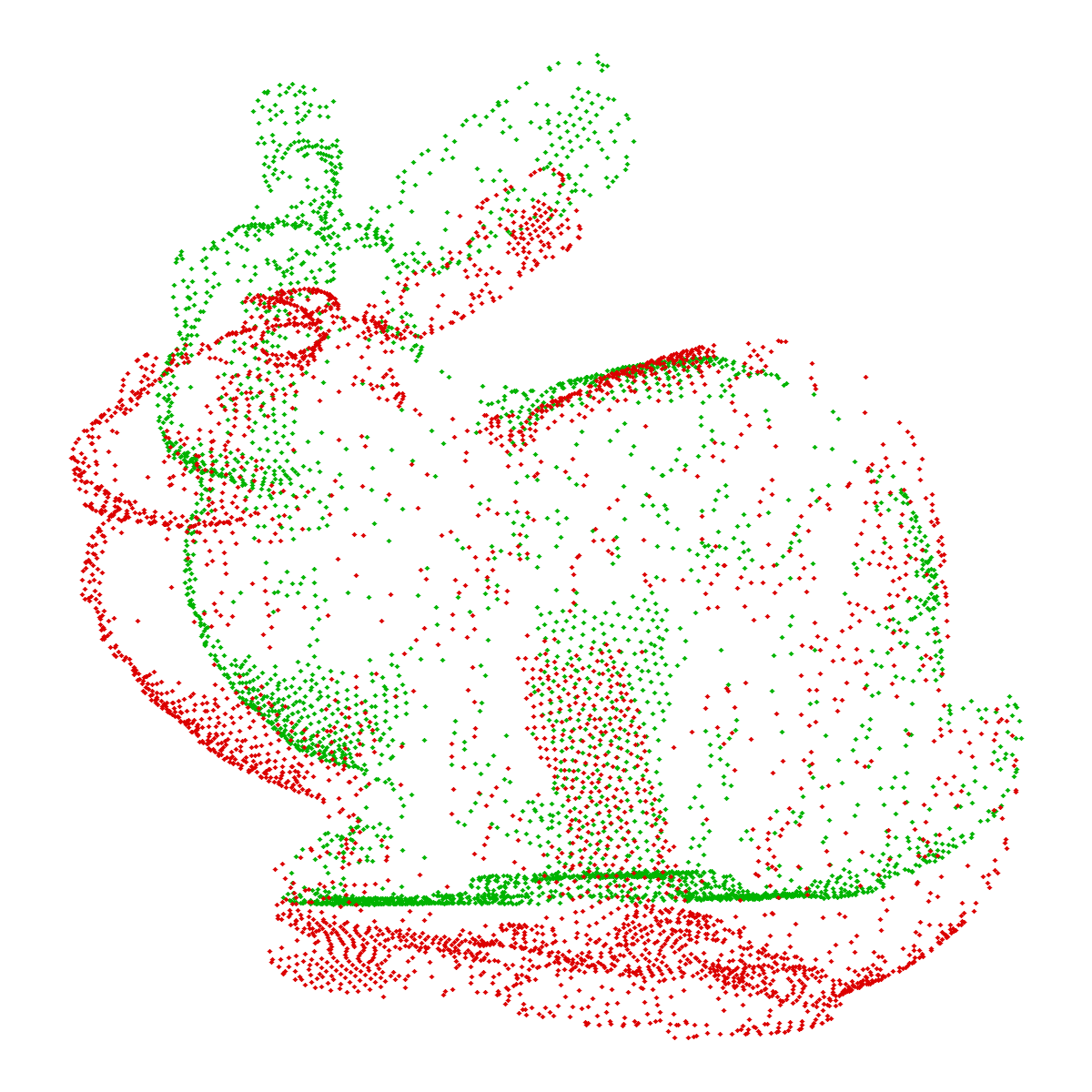}
\\[0.35em]

\rowlabel{MedShapeNet\\Liver\\8000 pts.}
&
\initcell{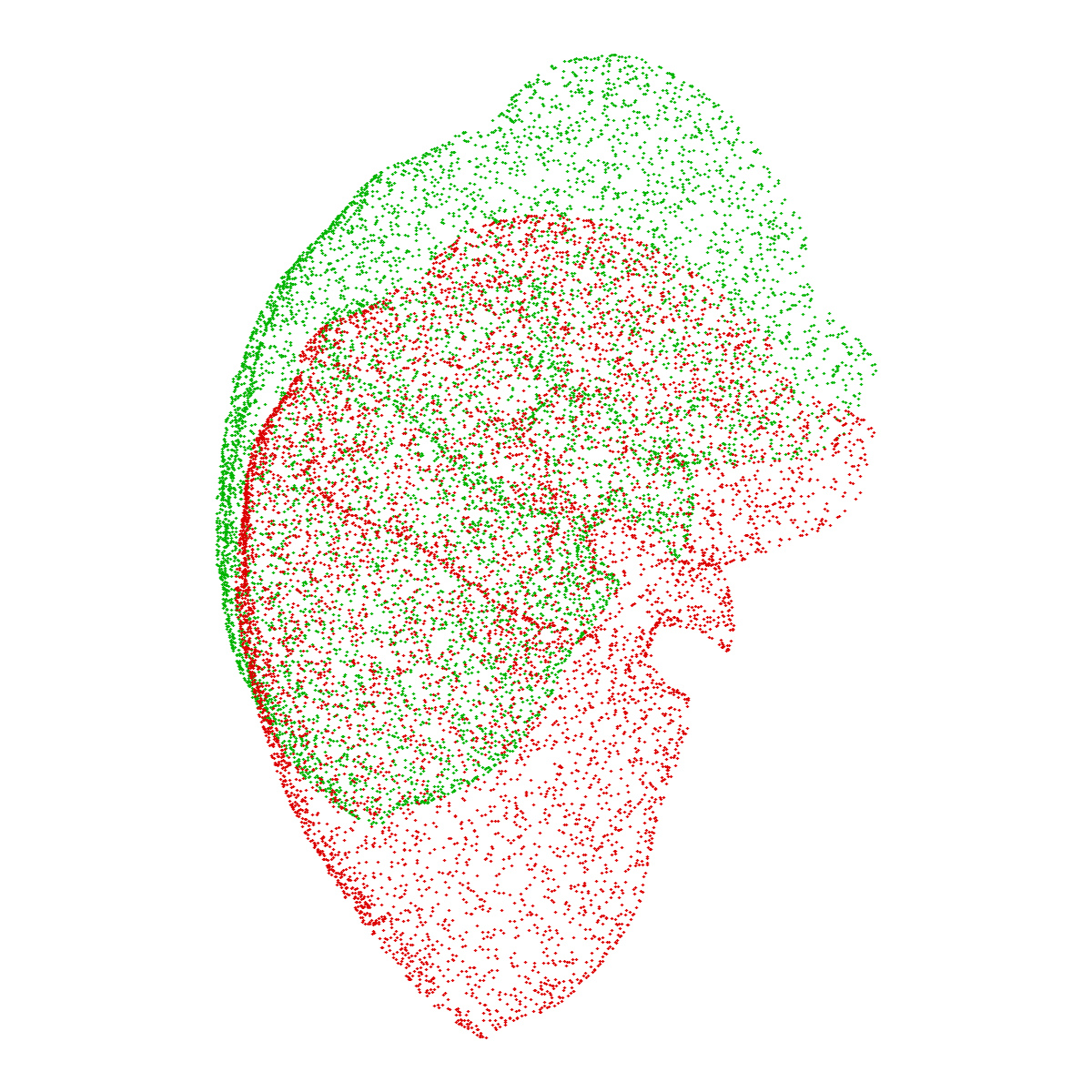}
&
\initcell{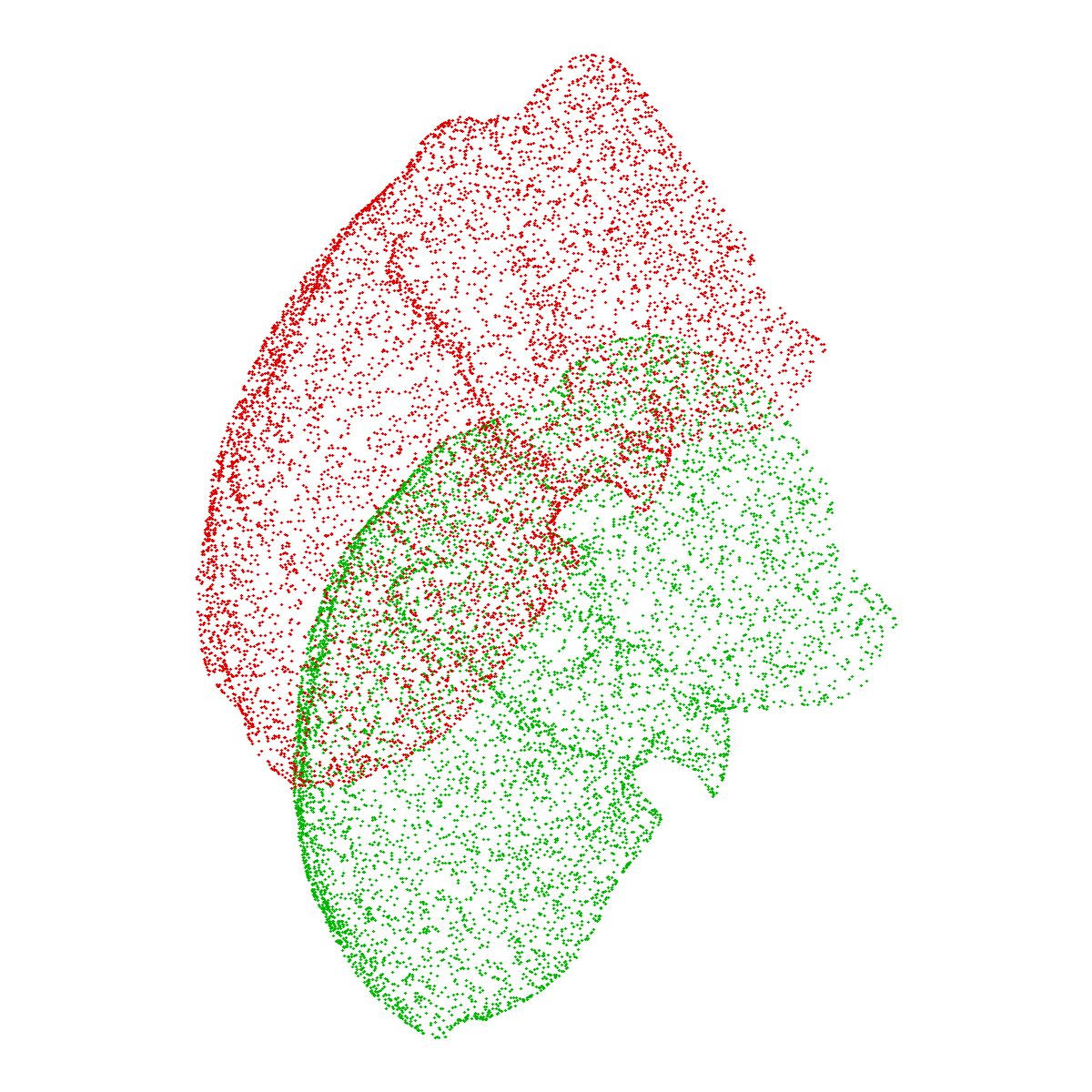}
&
\initcell{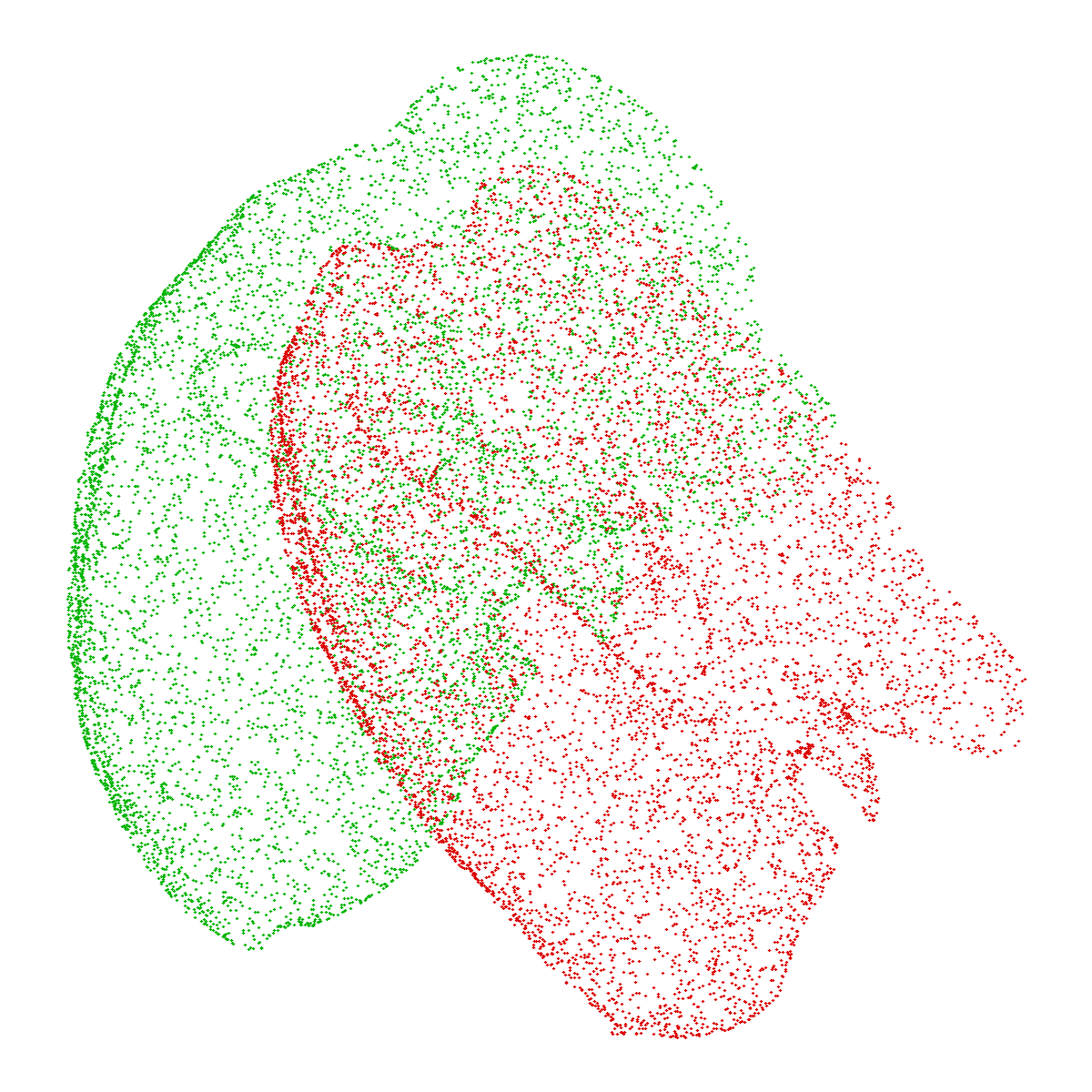}
&
\initcell{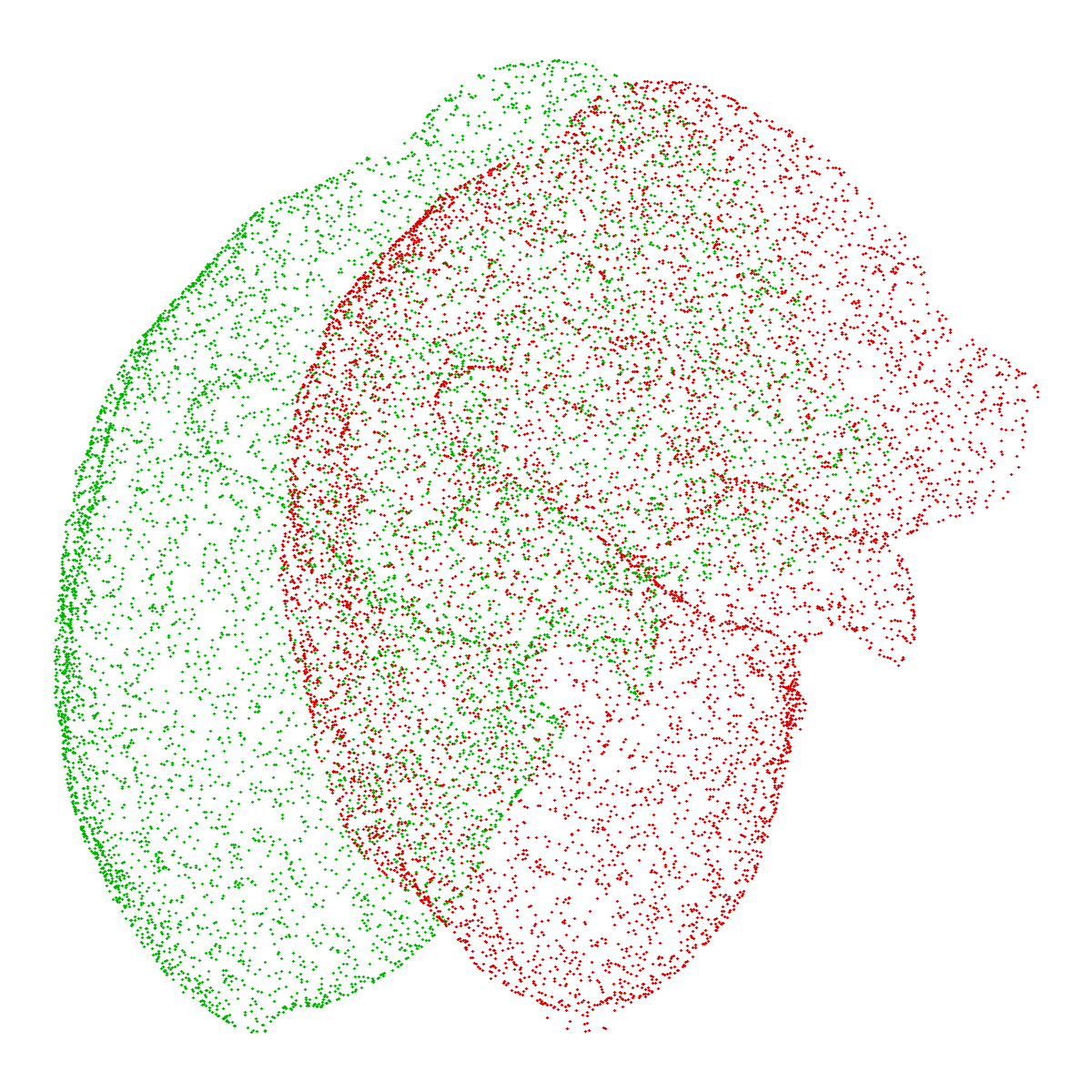}
&
\initcell{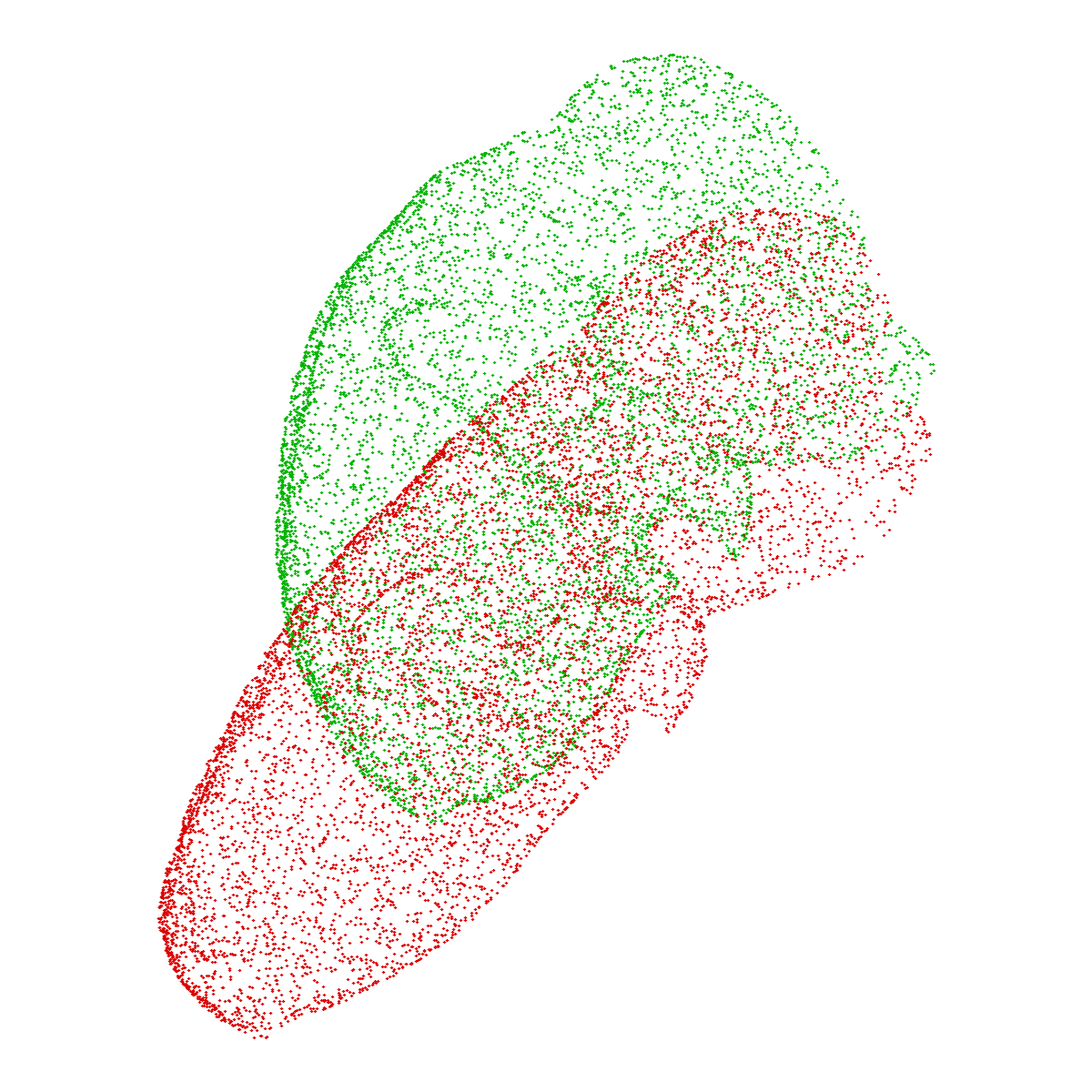}
\\[0.35em]

\rowlabel{MedShapeNet\\Kidney\\6000 pts.}
&
\initcell{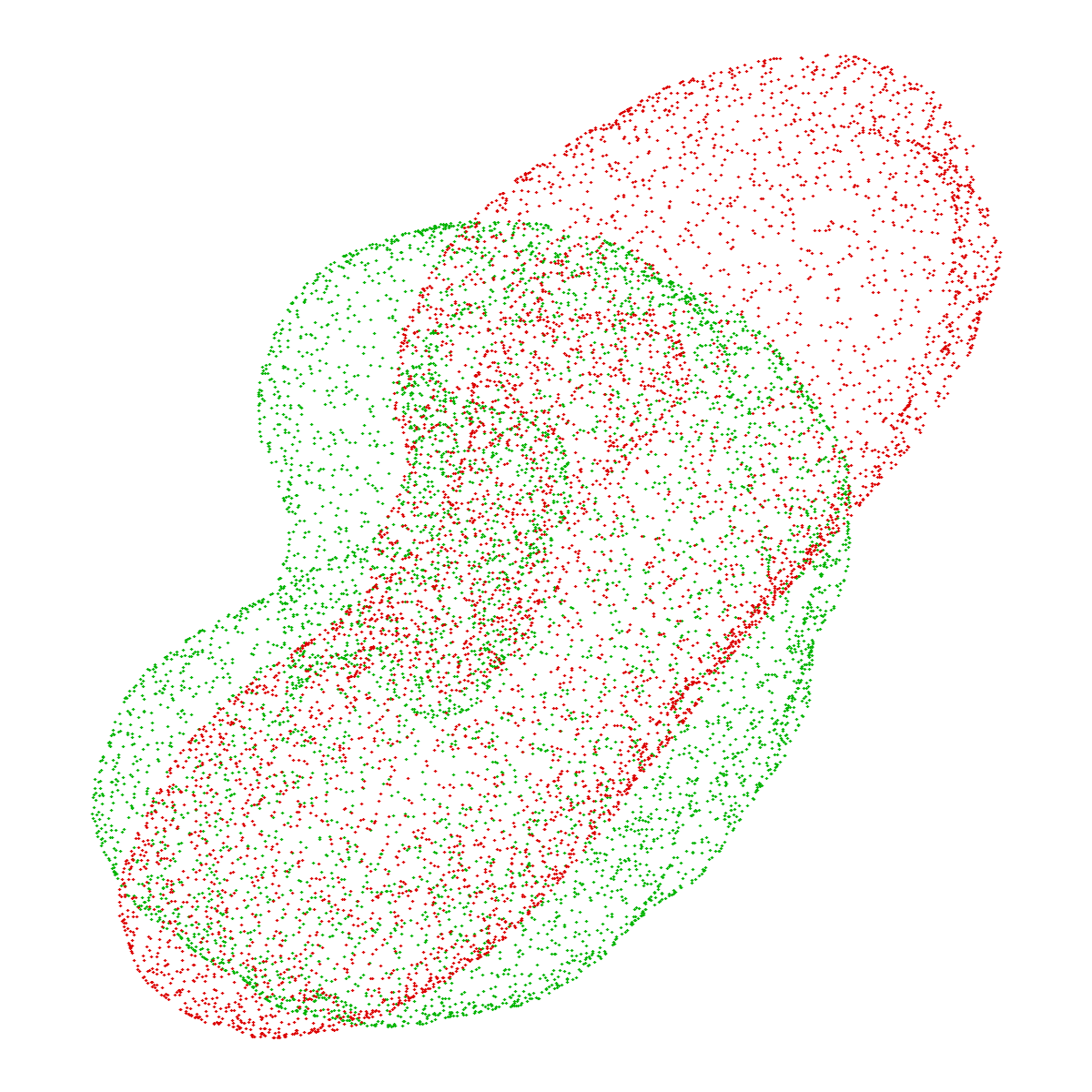}
&
\initcell{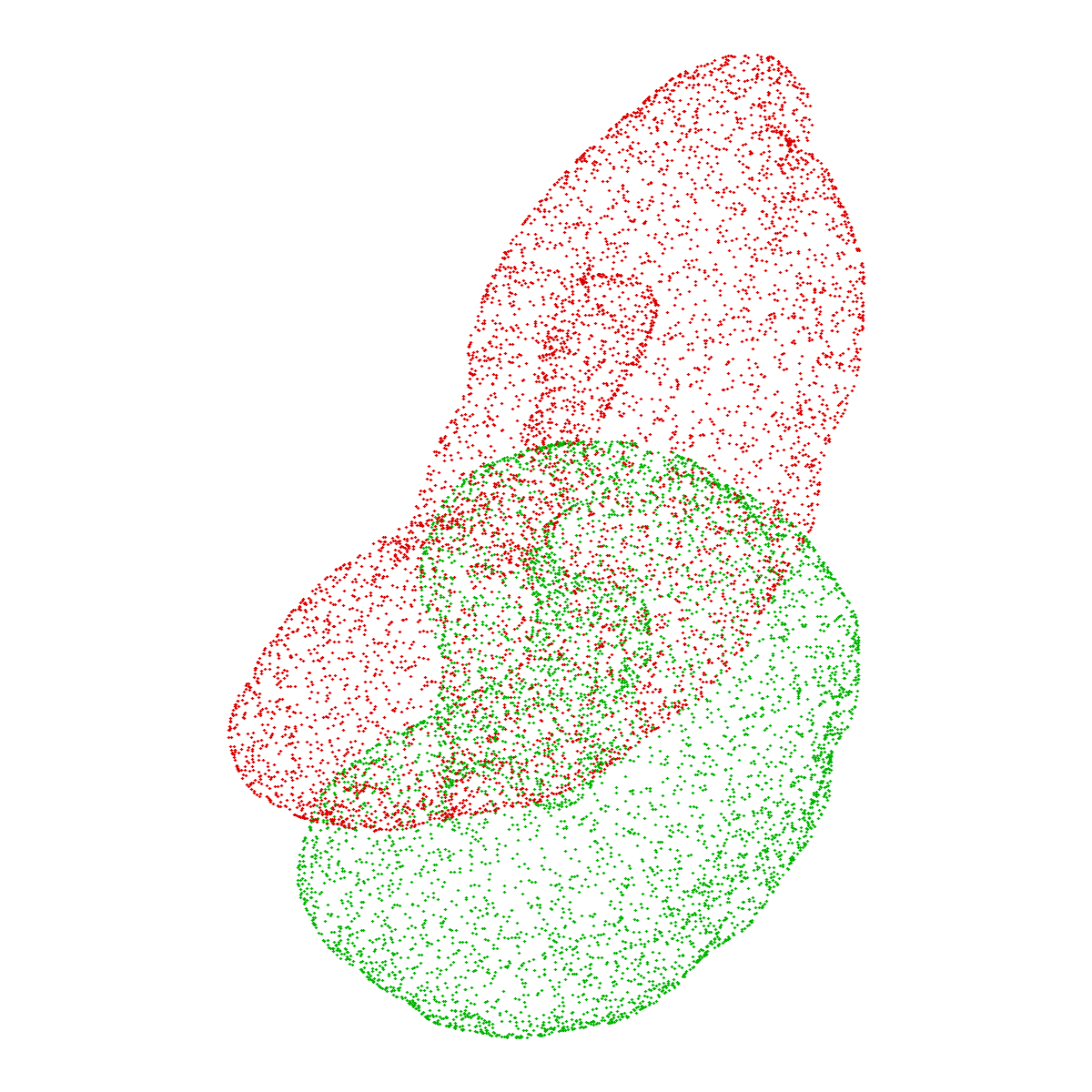}
&
\initcell{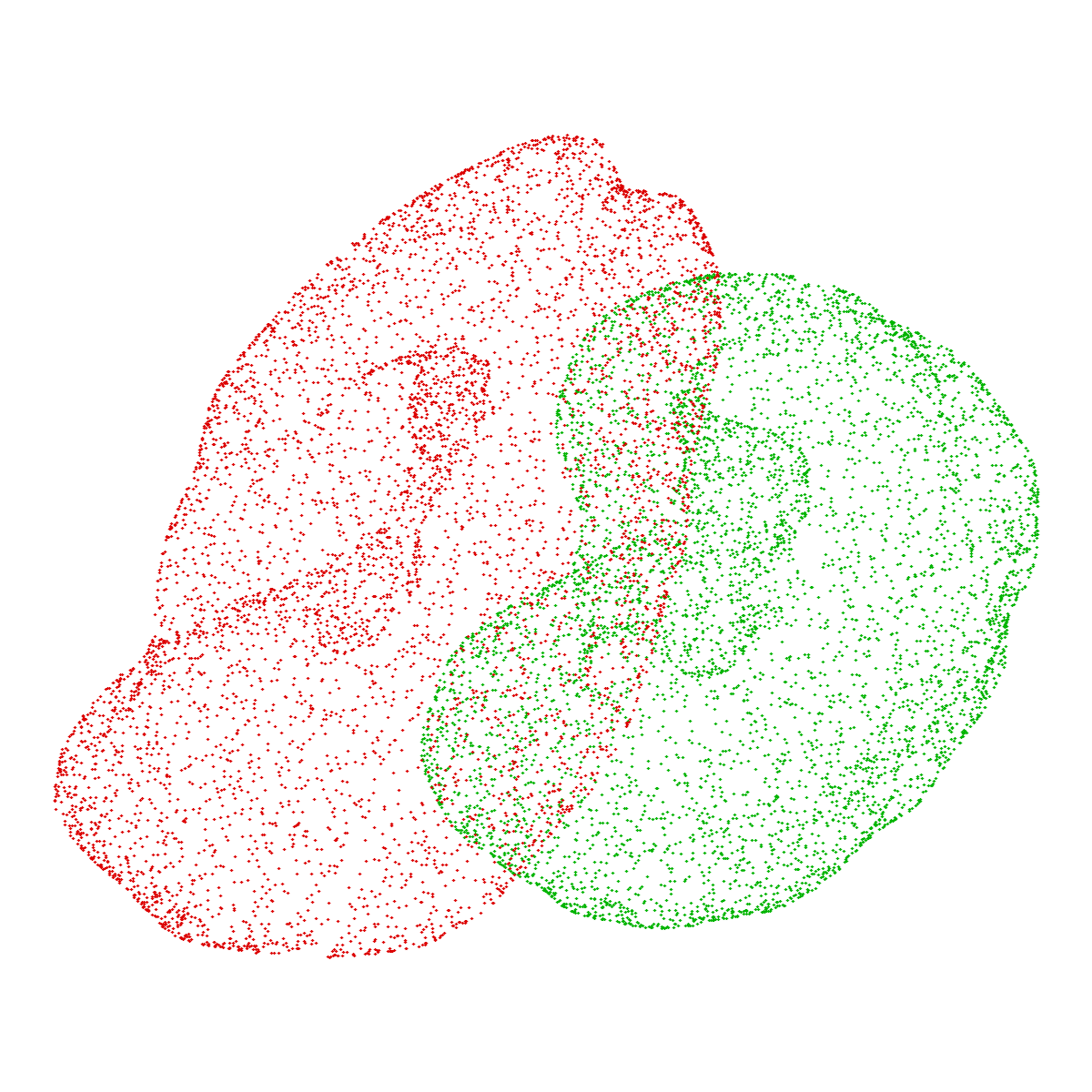}
&
\initcell{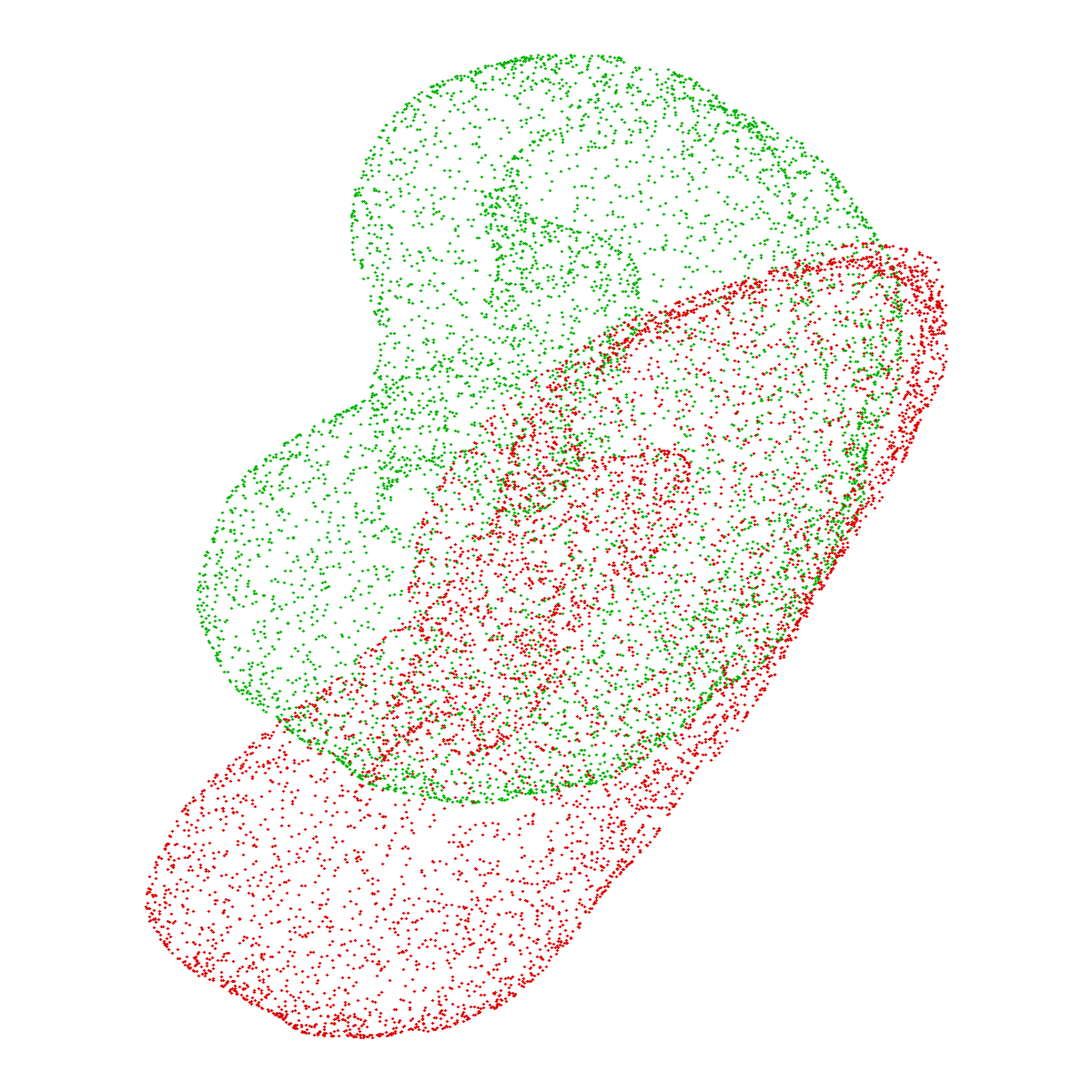}
&
\initcell{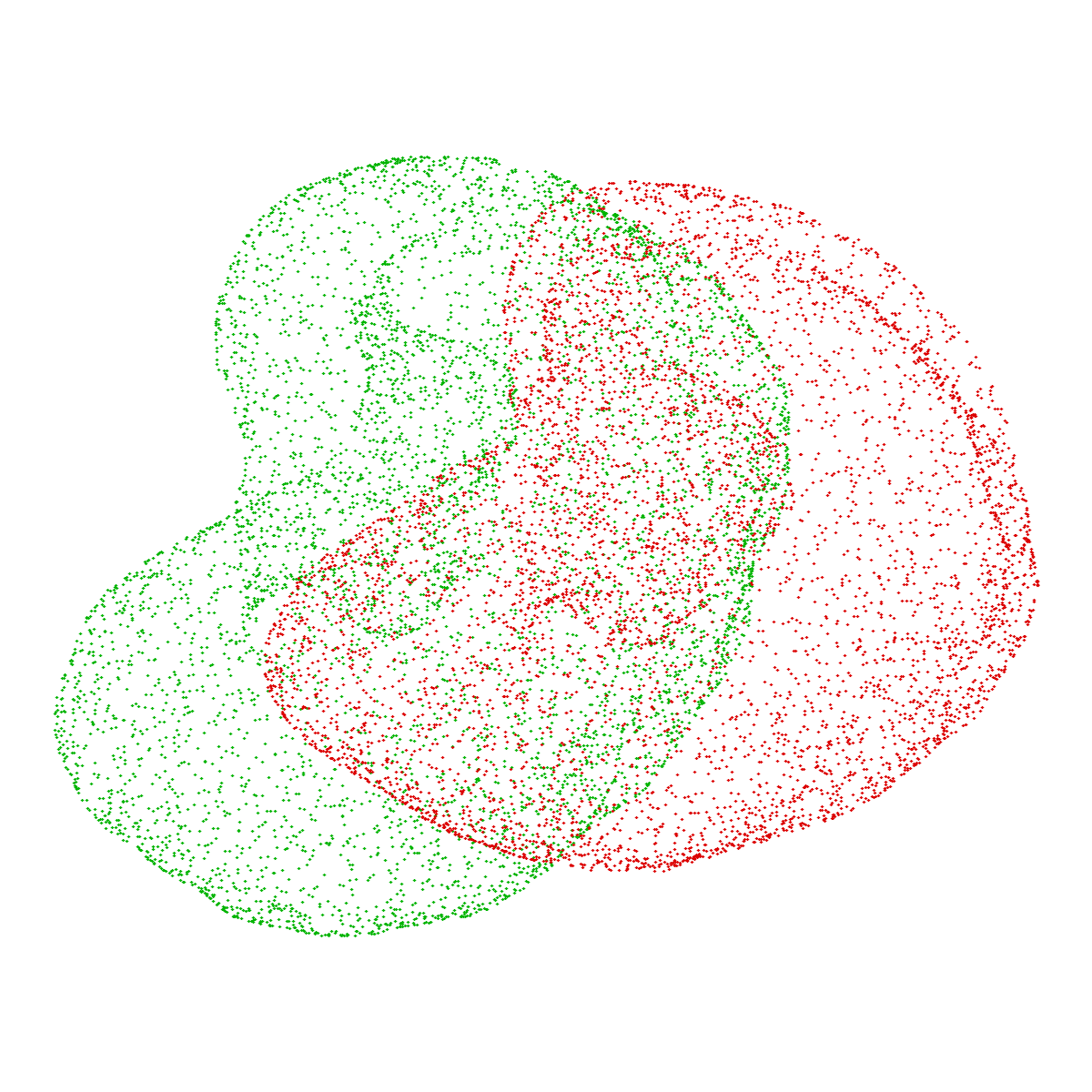}
\\

\bottomrule
\end{tabular}
\endgroup

\vspace{0.2em}

\caption{
Initial configurations for the multi-model and multi-seed experiment under smooth non-analytic deformations. Each row corresponds to one source model, and each column corresponds to one random deformation seed. Red points denote the fixed point set, and green points denote the initial moving point set before
registration.
}
\label{fig:multi-model-multi-seed-initial-configurations}
\end{figure}

We use five representative three-dimensional point-cloud models as source geometries: FAUST tr\_reg\_020 and FAUST tr\_reg\_038~\cite{bogo2014faust}, a diluted Stanford Bunny point cloud~\cite{turk1994zippered}, and MedShapeNet liver and kidney point clouds~\cite{li2025medshapenet}, with 8000 and 6000
sampled points, respectively. For each source geometry, five random deformation seeds are generated using the same \(C^\infty\) but non-analytic deformation mechanism described above, resulting in 25 registration cases in total. Thus,
the experiment combines realistic three-dimensional source geometries with controlled and reproducible smooth non-analytic deformation fields. The initial configurations are shown in
Fig.~\ref{fig:multi-model-multi-seed-initial-configurations}. All methods are evaluated on the same deformation instances, and all final errors are recomputed from the registered point sets using the same external pointwise RMSE metric in the normalized coordinate system.

Table~\ref{tab:multi-model-multi-seed-statistics} reports the final registration errors and running times. For each model and each method, results are reported as mean \(\pm\) standard deviation over five random seeds.

\begin{table}[t]
\centering
\caption{
Multi-model and multi-seed registration statistics under smooth non-analytic deformations. Errors and running times are reported as mean \(\pm\) standard deviation over five random seeds. Lower final error and lower time are better.
}
\label{tab:multi-model-multi-seed-statistics}

\begin{tabular*}{\columnwidth}{@{\extracolsep{\fill}}lcc@{}}
\toprule
Method & Final error & Time (s) \\
\midrule

\multicolumn{3}{@{}l}{\textbf{FAUST tr\_reg\_020}} \\
Analytic-CPD
& \(\mathbf{0.00751 \pm 0.00229}\)
& \(\mathbf{180.73 \pm 9.96}\) \\
ClusterReg
& \(0.08239 \pm 0.00695\)
& \(471.46 \pm 20.58\) \\
BCPD
& \(0.04380 \pm 0.02419\)
& \(286.61 \pm 52.85\) \\

\midrule
\multicolumn{3}{@{}l}{\textbf{FAUST tr\_reg\_038}} \\
Analytic-CPD
& \(\mathbf{0.00874 \pm 0.00374}\)
& \(\mathbf{189.39 \pm 11.03}\) \\
ClusterReg
& \(0.07500 \pm 0.01437\)
& \(480.55 \pm 12.98\) \\
BCPD
& \(0.06400 \pm 0.04098\)
& \(278.06 \pm 96.61\) \\

\midrule
\multicolumn{3}{@{}l}{\textbf{Stanford Bunny, diluted version, 3523 points}} \\
Analytic-CPD
& \(\mathbf{0.02660 \pm 0.02256}\)
& \(\mathbf{51.18 \pm 1.63}\) \\
ClusterReg
& \(0.20920 \pm 0.09263\)
& \(92.24 \pm 3.52\) \\
BCPD
& \(0.05920 \pm 0.03510\)
& \(55.54 \pm 26.87\) \\

\midrule
\multicolumn{3}{@{}l}{\textbf{MedShapeNet Liver, 8000 points}} \\
Analytic-CPD
& \(\mathbf{0.03618 \pm 0.01690}\)
& \(\mathbf{211.19 \pm 7.06}\) \\
ClusterReg
& \(0.07798 \pm 0.00883\)
& \(761.25 \pm 45.34\) \\
BCPD
& \(0.05200 \pm 0.01523\)
& \(495.60 \pm 114.36\) \\

\midrule
\multicolumn{3}{@{}l}{\textbf{MedShapeNet Kidney, 6000 points}} \\
Analytic-CPD
& \(\mathbf{0.03270 \pm 0.00608}\)
& \(\mathbf{132.72 \pm 13.83}\) \\
ClusterReg
& \(0.06690 \pm 0.01623\)
& \(335.44 \pm 21.25\) \\
BCPD
& \(0.04840 \pm 0.01754\)
& \(245.33 \pm 20.13\) \\

\bottomrule
\end{tabular*}
\end{table}

As shown in Table~\ref{tab:multi-model-multi-seed-statistics}, Analytic-CPD achieves the lowest average final error on all five models. On the two FAUST point clouds, its average errors are \(7.51\times 10^{-3}\) and \(8.74\times 10^{-3}\), respectively, substantially lower than those of ClusterReg and BCPD. On the diluted Stanford Bunny and the two MedShapeNet organ point clouds, Analytic-CPD also obtains the lowest final errors,
indicating that the method is not restricted to human-body geometry and can handle different three-dimensional shape structures.

Analytic-CPD also reports the lowest average runtime in all five groups under the runtime protocol used in this experiment. This observation is consistent with the intended design of the method: although Analytic-CPD retains the CPD-style posterior layer, its deformation update is governed by a compact structured analytic least-squares problem rather than a point-indexed kernel
displacement system. Since BCPD and ClusterReg are evaluated using their official implementations, the runtime values should be interpreted as official-implementation reference results rather than strict raw-code complexity comparisons.

Overall, this multi-model and multi-seed experiment confirms the main target regime of Analytic-CPD. The tested deformations are smooth but non-analytic and therefore do not lie exactly in a single finite-order Taylor family. Nevertheless, Analytic-CPD consistently provides accurate finite-dimensional smooth
approximations across different shapes and random seeds. This controlled setting complements the articulated human-motion examples by isolating the smooth deformation approximation ability of the registration model under known correspondences and reproducible deformation instances.

\textbf{Small-to-moderate deformation statistics on FAUST registered human shapes.}
\label{subsubsec:faust-small-statistics}

To complement the controlled smooth non-analytic deformation experiment, we further evaluate the methods on nine registered FAUST human-shape pairs with small-to-moderate pose and shape variations. 
This experiment is not intended to replace large-deformation or partial-observation benchmarks. 
Instead, it provides a non-synthetic validation setting in which the deformation patterns are not generated by our analytic or bump-based deformation model.

The initial configurations of the nine FAUST pairs are shown in Fig.~\ref{fig:faust-nine-initial-configurations}. 
For each pair, the red point cloud denotes the fixed model, while the green point cloud denotes the moving model. 
Since the FAUST registered models provide pointwise correspondence, all final errors are recomputed from the registered point sets using the same external pointwise RMSE metric in the normalized coordinate system. 
The initial RMSE values range from \(8.9\times 10^{-2}\) to \(3.10\times 10^{-1}\), covering small-to-moderate deformation magnitudes.

Table~\ref{tab:faust-small-statistics} reports the average and median results over the nine pairs. 
BCPD obtains the lowest mean RMSE by a small margin. 
Analytic-CPD, however, achieves the best RMSE in four out of the nine cases and remains very close to BCPD on average, while being substantially faster. 
In particular, the mean RMSE values of Analytic-CPD and BCPD are \(5.13\times 10^{-2}\) and \(5.01\times 10^{-2}\), respectively, whereas the mean runtime is reduced from \(243.71\) s to \(75.69\) s. 
Measured by the per-case runtime ratio, Analytic-CPD is on average \(5.71\times\) faster than BCPD, with a median speedup of \(5.13\times\).

These results suggest that the proposed compact analytic M-step provides a favorable accuracy--efficiency trade-off on non-synthetic registered human-shape data. 
Although Analytic-CPD is not uniformly more accurate than BCPD on this FAUST subset, it achieves comparable accuracy with a much smaller computational cost. 
This supports the use of the structured analytic deformation model as an efficient alternative to kernel-based non-rigid CPD variants when the deformation is topology-preserving and admits a compact low-dimensional representation.

\begin{table}[t]
\centering
\caption{
Small-to-moderate deformation statistics on nine FAUST registered human-shape
pairs. All final errors are recomputed using the same external pointwise RMSE
metric in the normalized coordinate system. The best value in each column is
highlighted in bold.
}
\label{tab:faust-small-statistics}
\scriptsize
\setlength{\tabcolsep}{3.5pt}
\renewcommand{\arraystretch}{1.08}
\begin{tabular}{lccccc}
\toprule
Method 
& \begin{tabular}{c}Mean\\RMSE\end{tabular}
& \begin{tabular}{c}Median\\RMSE\end{tabular}
& \begin{tabular}{c}Mean\\time (s)\end{tabular}
& \begin{tabular}{c}Median\\time (s)\end{tabular}
& \begin{tabular}{c}Best\\cases\end{tabular} \\
\midrule
Analytic-CPD 
& \(5.13{\times}10^{-2}\) 
& \(4.20{\times}10^{-2}\) 
& \(\mathbf{75.69}\) 
& \(\mathbf{40.24}\) 
& \(4/9\) \\
BCPD 
& \(\mathbf{5.01{\times}10^{-2}}\) 
& \(\mathbf{3.40{\times}10^{-2}}\) 
& \(243.71\) 
& \(240.56\) 
& \(\mathbf{5/9}\) \\
ClusterReg 
& \(6.48{\times}10^{-2}\) 
& \(4.90{\times}10^{-2}\) 
& \(457.88\) 
& \(455.92\) 
& \(0/9\) \\
\bottomrule
\end{tabular}
\end{table}

\begin{figure}[t]
\centering

\begingroup
\setlength{\tabcolsep}{2pt}
\renewcommand{\arraystretch}{1.03}

\begin{tabular}{
    @{}
    >{\centering\arraybackslash}m{0.32\linewidth}
    >{\centering\arraybackslash}m{0.32\linewidth}
    >{\centering\arraybackslash}m{0.32\linewidth}
    @{}
}
\includegraphics[width=\linewidth]{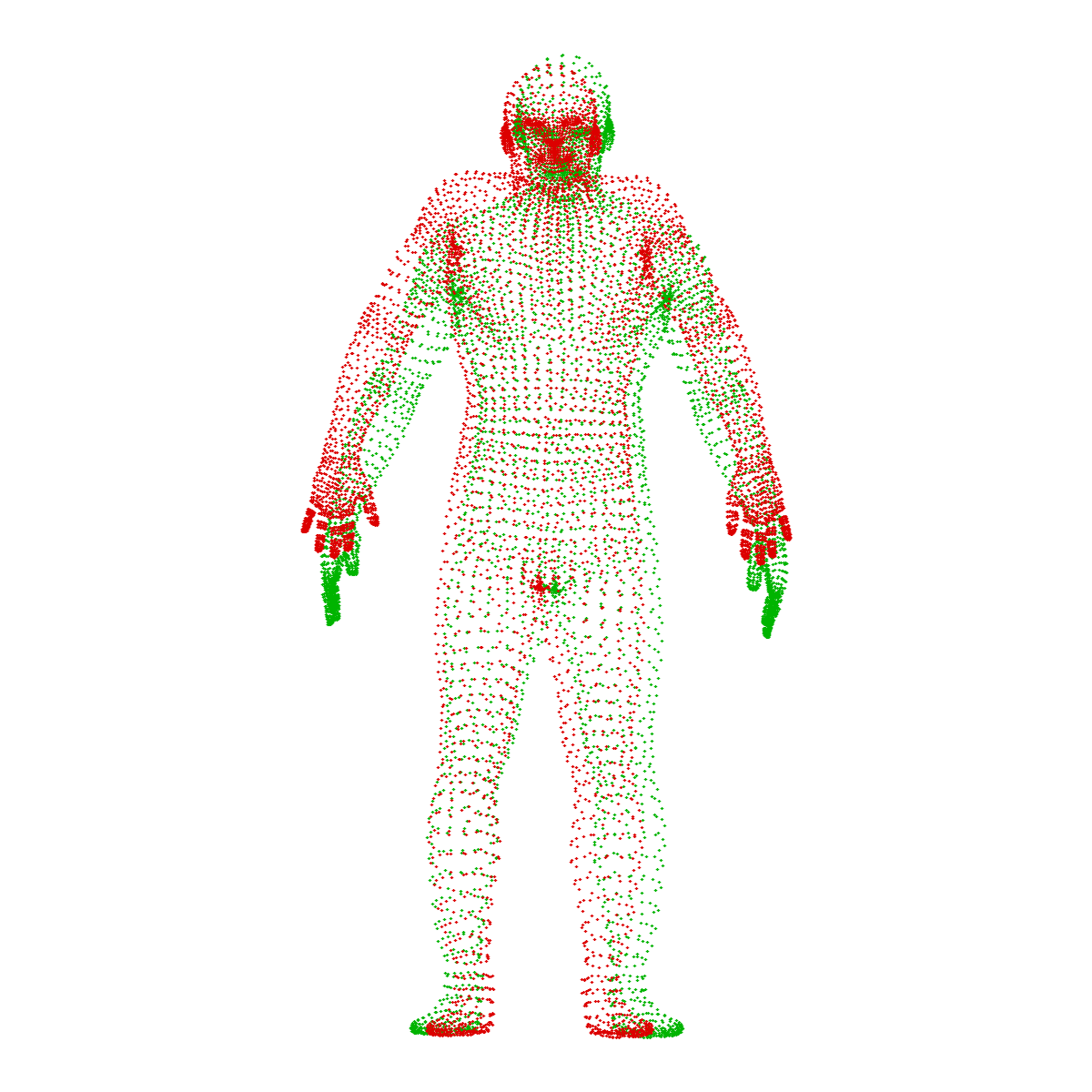}
&
\includegraphics[width=\linewidth]{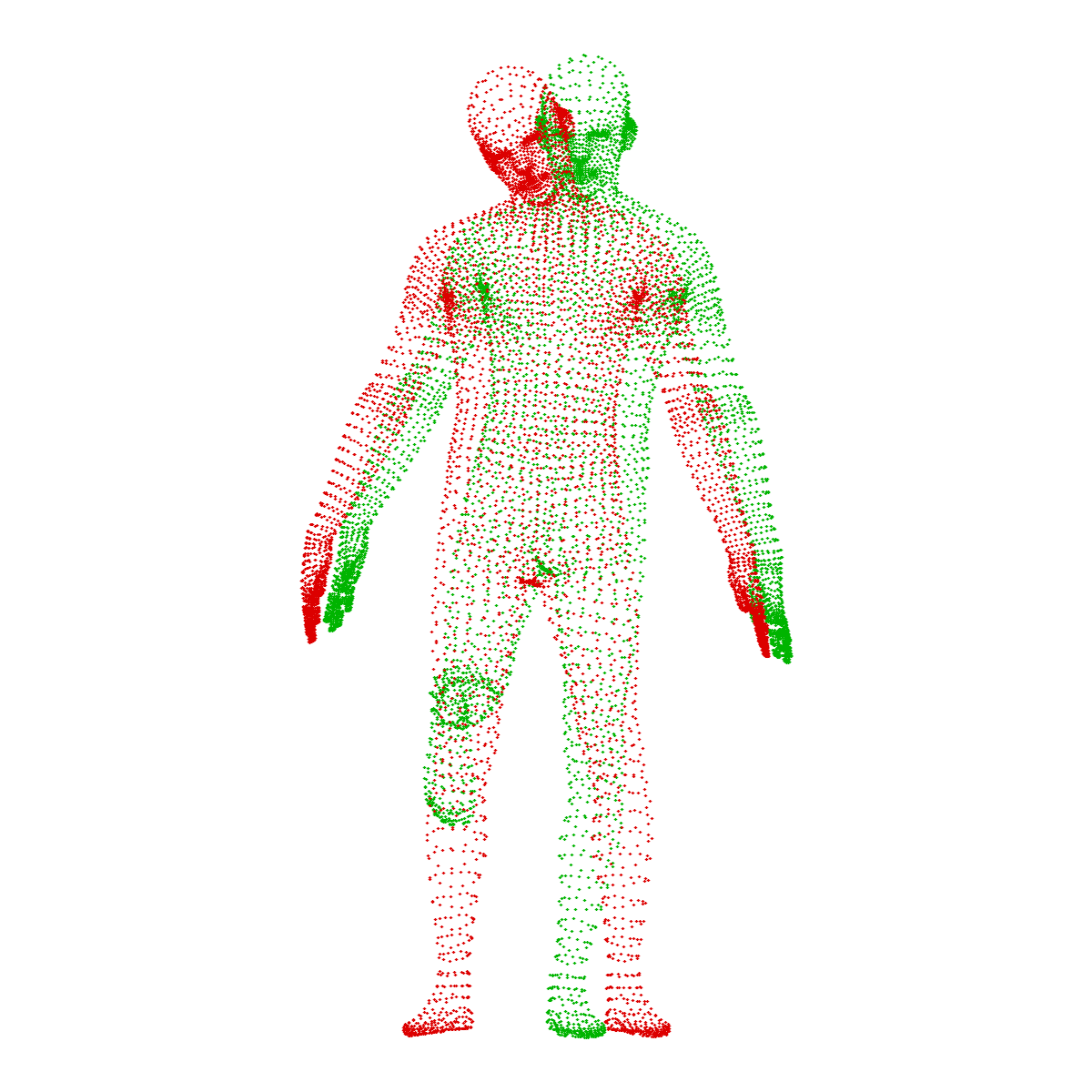}
&
\includegraphics[width=\linewidth]{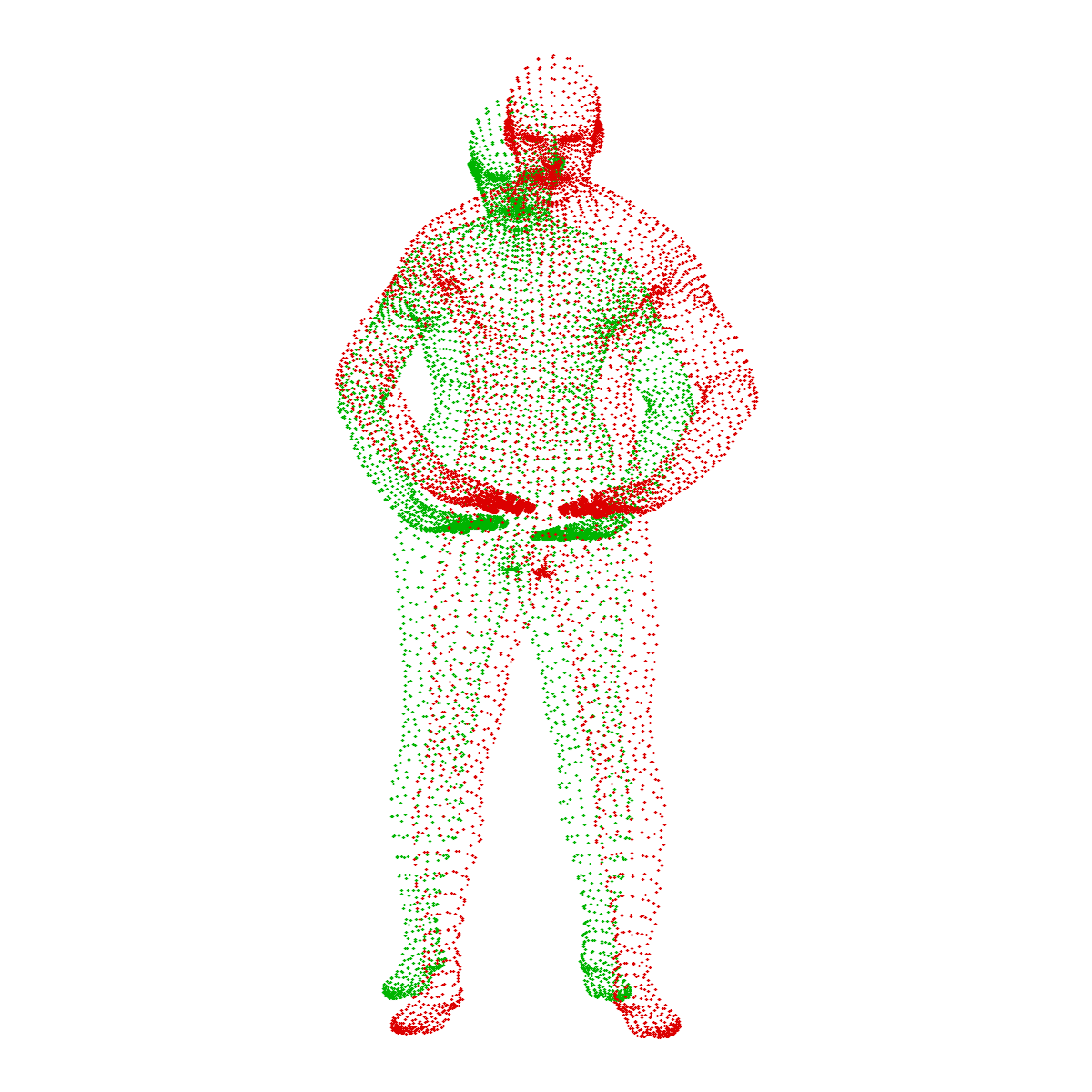}
\\[-0.25em]
\scriptsize (a) Pair 1, RMSE \(=8.90\times10^{-2}\)
&
\scriptsize (b) Pair 2, RMSE \(=1.44\times10^{-1}\)
&
\scriptsize (c) Pair 3, RMSE \(=1.57\times10^{-1}\)
\\[0.45em]

\includegraphics[width=\linewidth]{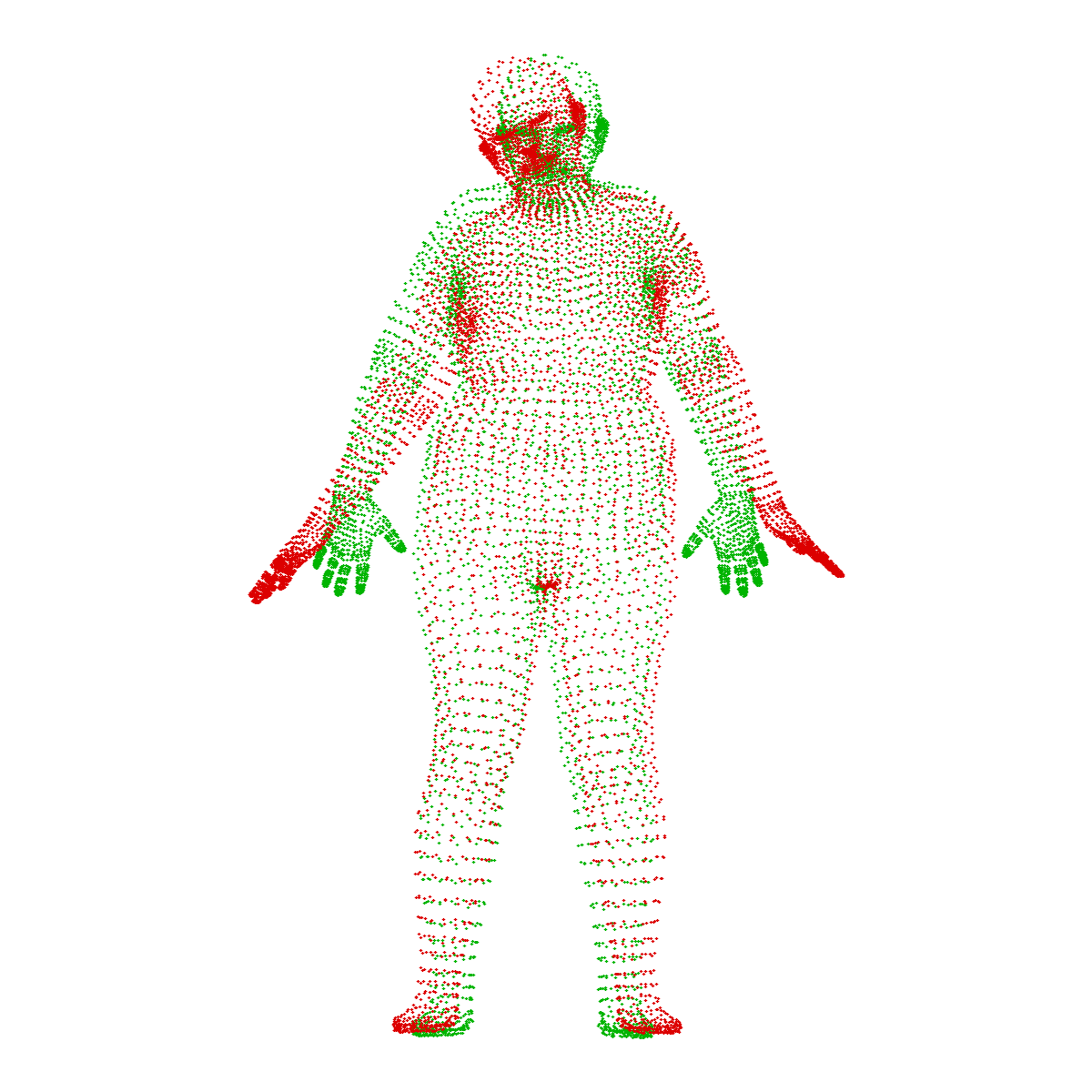}
&
\includegraphics[width=\linewidth]{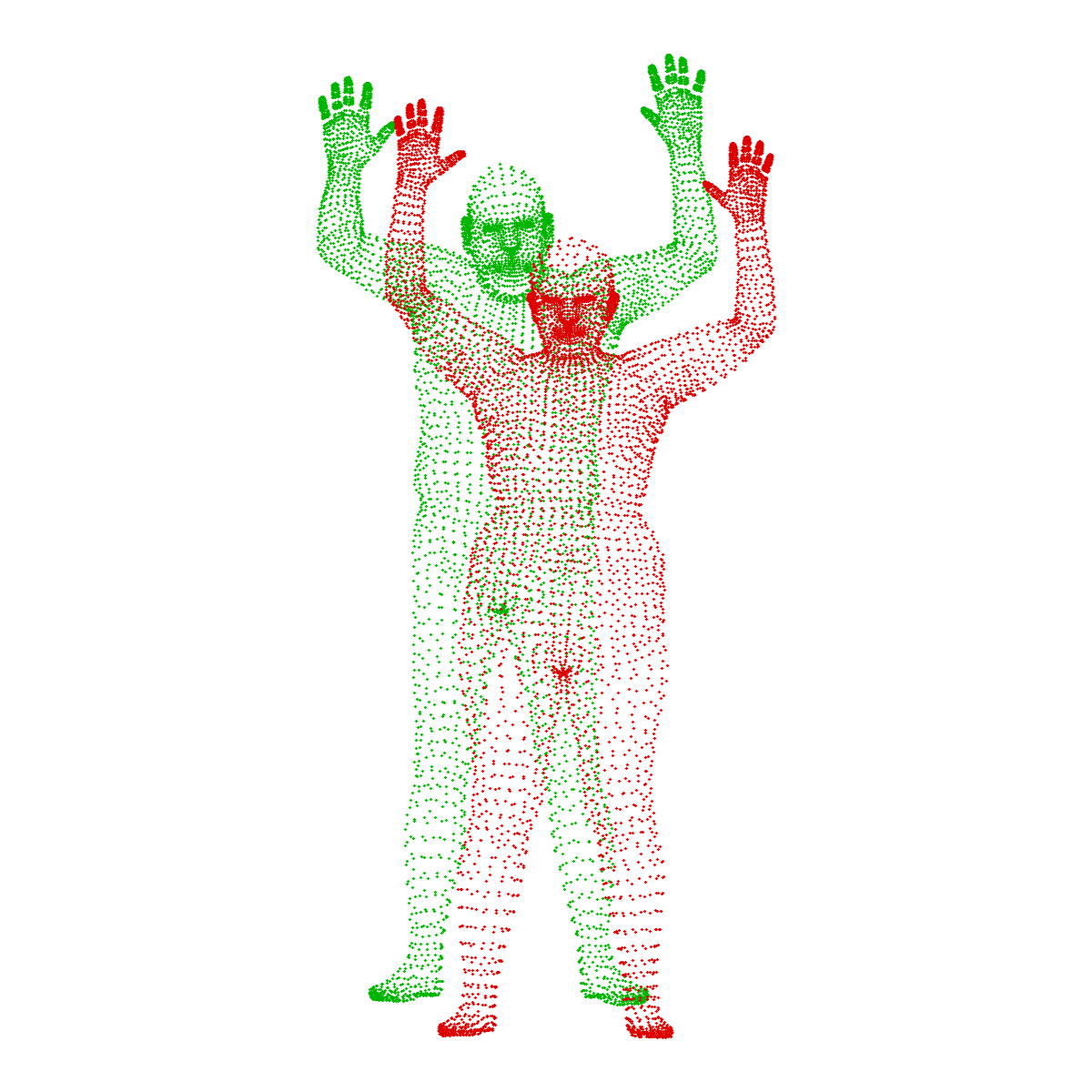}
&
\includegraphics[width=\linewidth]{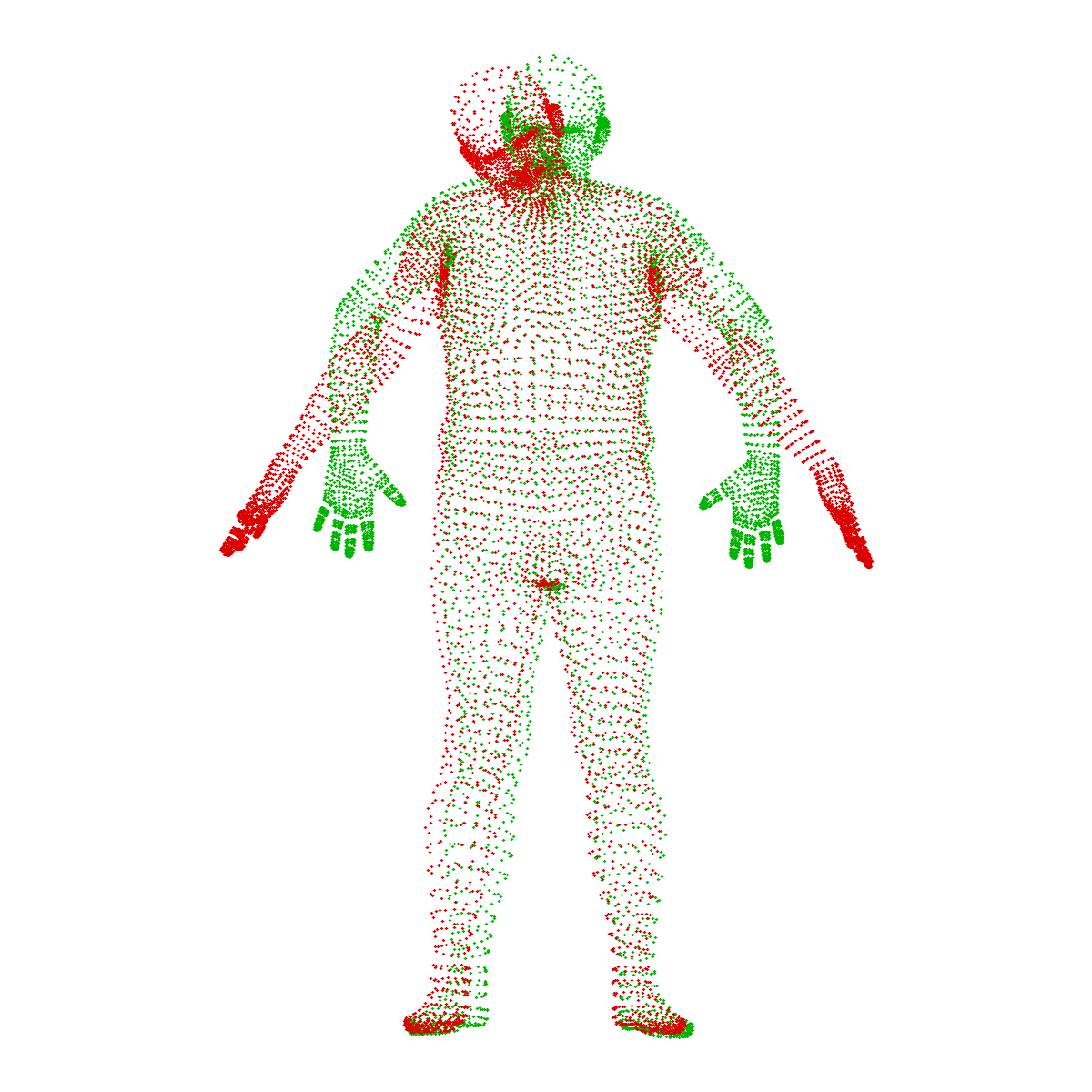}
\\[-0.25em]
\scriptsize (d) Pair 4, RMSE \(=1.20\times10^{-1}\)
&
\scriptsize (e) Pair 5, RMSE \(=2.10\times10^{-1}\)
&
\scriptsize (f) Pair 6, RMSE \(=1.30\times10^{-1}\)
\\[0.45em]

\includegraphics[width=\linewidth]{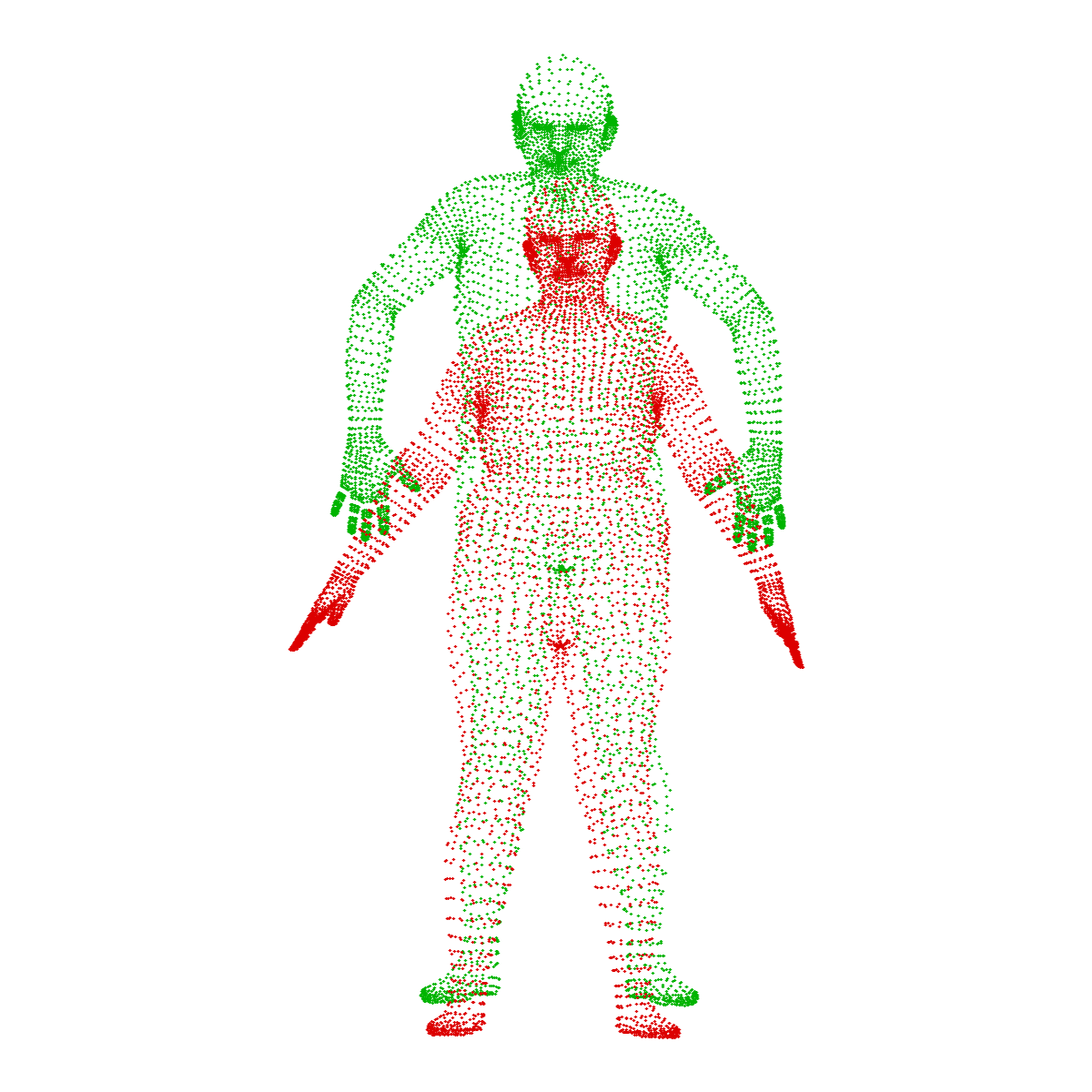}
&
\includegraphics[width=\linewidth]{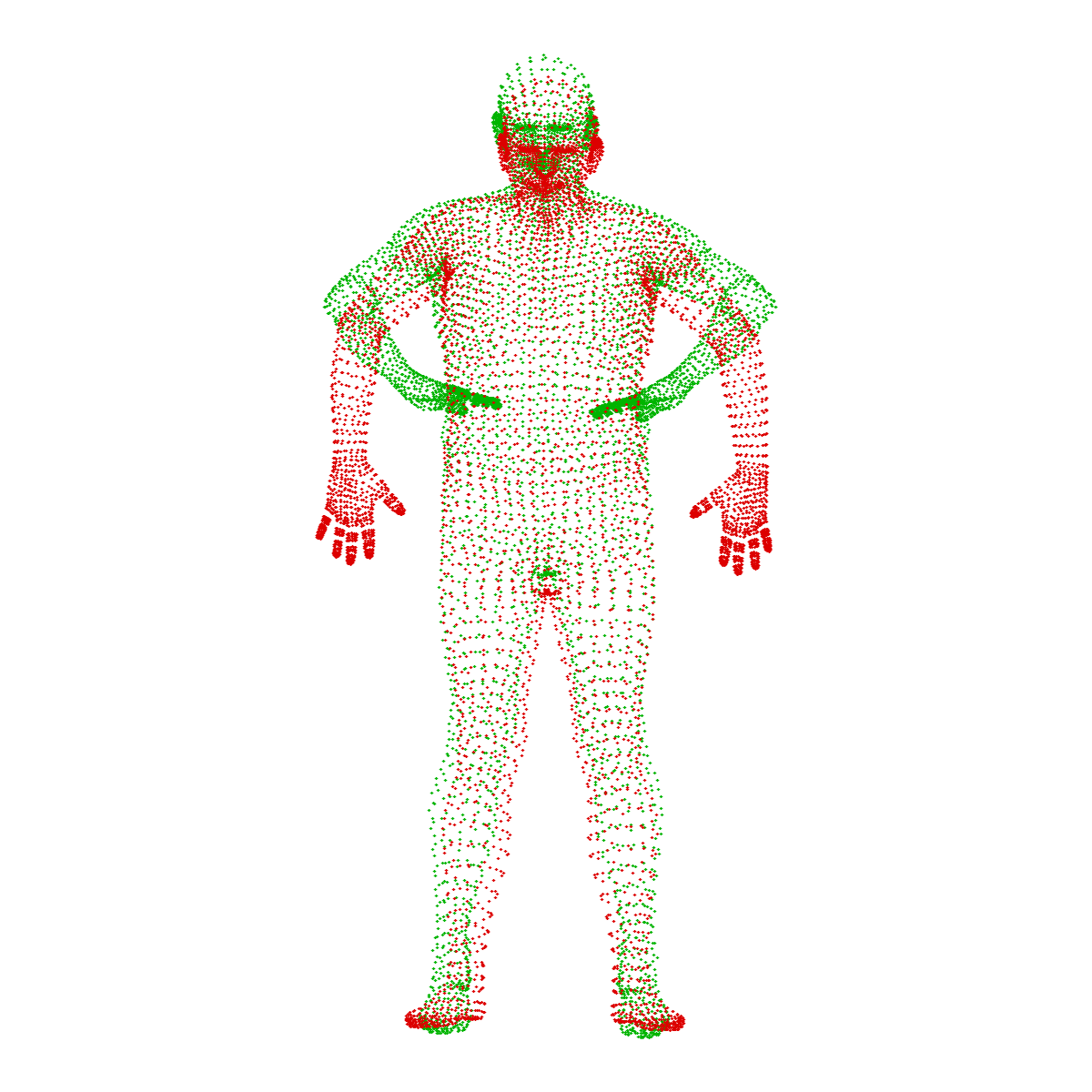}
&
\includegraphics[width=\linewidth]{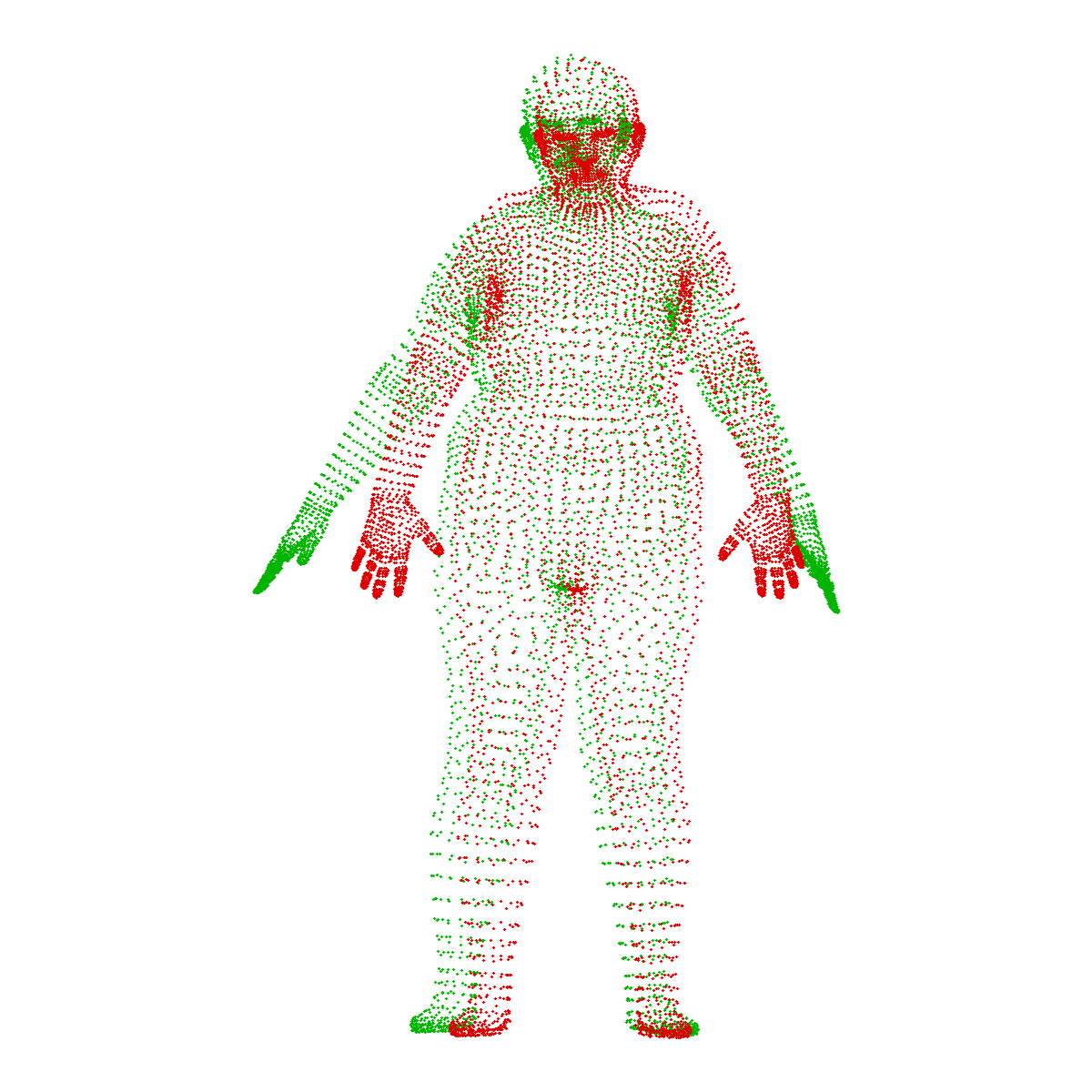}
\\[-0.25em]
\scriptsize (g) Pair 7, RMSE \(=2.30\times10^{-1}\)
&
\scriptsize (h) Pair 8, RMSE \(=3.10\times10^{-1}\)
&
\scriptsize (i) Pair 9, RMSE \(=1.20\times10^{-1}\)
\end{tabular}

\endgroup

\caption{
Initial configurations of the nine FAUST registered human-shape pairs used in the small-deformation statistics.
Red denotes the fixed point cloud, and green denotes the moving point cloud.
The pairs cover small-to-moderate initial deformation magnitudes in the normalized coordinate system.
}
\label{fig:faust-nine-initial-configurations}
\end{figure}

\subsection{Ablation on Degree Continuation}
\label{subsec:degree-ablation}

We examine the role of the degree-continuation strategy using five randomly generated three-dimensional large-deformation cases. All cases are produced by the same smooth non-analytic deformation generator with different random seeds. Each degree strategy is evaluated on the same five deformation instances, with
the same maximum iteration number \(T_{\max}=100\).

We compare fixed-order structured analytic mappings with \(q=1,2,5,10\) against the proposed increasing-degree continuation from \(q=1\) to \(q_{\max}=10\) with decreasing stage lengths. Low fixed orders are expected to be stable but may underfit large nonlinear deformation, whereas high fixed orders provide more expressive power but may become unstable when activated before the posterior correspondences are reliable.

\begin{table}[t]
\centering
\caption{
Ablation study on degree continuation over five randomly generated
three-dimensional large-deformation cases. All strategies use the same maximum iteration number \(T_{\max}=100\). Final errors and times are averaged over successful runs only.
}
\label{tab:degree-continuation-ablation}

\setlength{\tabcolsep}{3pt}
\renewcommand{\arraystretch}{1.08}

\begin{tabular}{
@{}
>{\raggedright\arraybackslash}p{0.43\linewidth}
>{\centering\arraybackslash}p{0.22\linewidth}
>{\centering\arraybackslash}p{0.16\linewidth}
>{\centering\arraybackslash}p{0.13\linewidth}
@{}
}
\toprule
Degree strategy & Final error & Time (s) & Failures \\
\midrule
Fixed \(q=1\) 
& \(5.67\times 10^{-2}\) 
& \(189.03\) 
& \(0/5\) \\

Fixed \(q=2\) 
& \(9.12\times 10^{-3}\) 
& \(228.95\) 
& \(0/5\) \\

Fixed \(q=5\) 
& \(1.37\times 10^{-1}\) 
& \(245.55\) 
& \(4/5\) \\

Fixed \(q=10\) 
& -- 
& -- 
& \(5/5\) \\

Continuation \(1\!\rightarrow\!10\) 
& \(\mathbf{4.12\times 10^{-7}}\) 
& \(342.06\) 
& \(0/5\) \\
\bottomrule
\end{tabular}
\end{table}

Table~\ref{tab:degree-continuation-ablation} shows that degree continuation is essential for stable high-order analytic fitting. Fixed \(q=1\) and \(q=2\) complete all runs but underfit the deformation. Directly activating higher orders is unstable: fixed \(q=5\) fails in four out of five cases, and fixed \(q=10\) fails in all cases. In contrast, the proposed continuation strategy
succeeds in all five cases and achieves the lowest final error. This confirms that Analytic-CPD should first stabilize the alignment and posterior correspondences with low-order analytic mappings, and only later introduce higher-order deformation modes for nonlinear refinement.

\subsection{Discussion and Limitations}
\label{subsec:discussion-limitations}

Analytic-CPD can be viewed as a probabilistic realization of the SAM principle inside the CPD framework. SAM provides a structured analytic approximation model for smooth vector-valued mappings, while CPD provides a probabilistic soft-correspondence mechanism. Through the barycentric condensation of the CPD posterior objective, Analytic-CPD converts the non-rigid CPD M-step into a
weighted structured analytic fitting problem. This changes the deformation representation from a point-indexed kernel displacement field to a compact, interpretable, and degree-controllable analytic mapping space, without abandoning the probabilistic correspondence layer.

The analytic M-step should not be interpreted as a low-order polynomial heuristic. Structured Taylor approximants provide a general approximation framework for smooth mappings on compact domains, and the compositional update inherits the order-amplification property of SAM. Hence, even when each
individual M-step uses a moderate Taylor order, the accumulated map can have rapidly increasing effective expressive order. This explains why Analytic-CPD can handle not only model-matched analytic deformations, but also smooth non-analytic deformations and registered FAUST human-shape pairs that are not generated by a single finite-order Taylor map.

Compared with BCPD and related kernel-field methods, Analytic-CPD adopts a different organization of the deformation space rather than a simply weaker one. Kernel-field methods use point- or kernel-indexed displacement variables with coherence regularization, whereas Analytic-CPD organizes smooth deformation modes by analytic degree and composition. For smooth topology-preserving deformations on bounded domains, the practical difference lies mainly in basis organization, parameter dimension, numerical conditioning, and optimization strategy.

The current implementation is still a direct prototype. It uses a single expansion center, a prescribed maximum order, a fixed degree-continuation schedule, and the direct \(\mathcal{O}(MN)\) posterior computation inherited from CPD. Strongly inhomogeneous smooth deformations may benefit from multi-center analytic bases, local analytic patches, or adaptive degree selection, while very large point sets require accelerated posterior computation. Cases involving severe missing data, partial overlap, topology change, discontinuities, or ambiguous repeated structures also require
additional correspondence and data-modeling mechanisms. These are limitations of the present registration pipeline rather than of the structured analytic approximation principle itself.

\section{Conclusion}
\label{sec:conclusion}

We introduced \textbf{Analytic-CPD}, a new unsupervised non-rigid point set registration framework that couples the Gaussian-mixture posterior mechanism of CPD with the structured analytic deformation model of SAM. The method preserves the probabilistic posterior correspondence layer of CPD, but lifts the non-rigid M-step from point-indexed kernel displacement estimation to structured analytic mapping estimation. In this formulation, posterior probabilities provide soft probabilistic geometric guidance, while the deformation is estimated over a finite-dimensional hierarchy of analytic function spaces.

This structured analytic reformulation brings three main consequences. First, the coefficient dimension of the deformation model is governed by the ambient dimension and analytic order rather than by the number of moving points. Second, the deformation model becomes interpretable and degree-controllable through the nested Taylor mapping spaces. Third, the compositional update and increasing-degree continuation strategy allow low-order analytic maps to stabilize early posterior correspondences, while higher-order modes are activated later to refine nonlinear residual deformation. The ablation study
confirms that direct high-order fitting can be unstable, whereas degree continuation substantially improves robustness and final accuracy.

Experiments on controlled analytic deformations, smooth non-analytic model-mismatch tests, multi-model multi-seed statistics, and registered human-motion cases demonstrate the effectiveness of the proposed formulation. The results support structured analytic mappings as a compact, interpretable, and degree-controllable alternative to point-indexed kernel deformation models in CPD-type non-rigid registration.

The current direct implementation focuses on smooth topology-preserving \(n\)-to-\(n\) registration. Cases involving severe missing data, partial overlap, topology change, discontinuity, or strong ambiguity require additional correspondence and data-modeling mechanisms. Future work will investigate adaptive Taylor-degree flow, in which the analytic order evolves over nested
Taylor mapping spaces according to posterior reliability, residual structure, and numerical conditioning. Multi-center analytic bases, local analytic patches, accelerated posterior computation, and noise-aware coefficient regularization may further improve the flexibility, stability, and scalability
of structured analytic point-set registration.

\ifCLASSOPTIONcompsoc
  \section*{Acknowledgments}

\else
  \section*{Acknowledgment}
\fi

This work was supported in part by the National Natural Science Foundation of China (Grant Nos.\ 62571503 and 62171421) and in part by the TaiShan Scholar Youth Expert Program of Shandong Province (Grant No.\ tsqn202306096).

%\ifCLASSOPTIONcaptionsoff
  %\newpage
%\fi

\appendices

\section{Complexity Analysis}
\label{app:complexity-analysis}

Let \(N\) and \(M\) denote the numbers of fixed and moving points, respectively, and let \(d\) be the ambient dimension. For a structured analytic mapping truncated at order \(q\), the number of Taylor basis terms is
\[
S_{d,q}=\binom{q+d}{d},
\]
and the number of scalar coefficients of a vector-valued mapping is
\[
K_{d,q}=dS_{d,q}.
\]
These quantities depend on the analytic order \(q\) and the dimension \(d\), but not on the number of moving points.

The E-step of Analytic-CPD is the same as that of CPD in the direct implementation: it computes posterior probabilities between all pairs of fixed and moving points. Therefore, the direct posterior computation costs
\[
\mathcal{O}(MNd)
\]
time, and storing the full posterior matrix costs \(\mathcal{O}(MN)\) memory. The sufficient statistics needed by the M-step, such as \(P\mathbf 1\), \(P^\top\mathbf 1\), and \(PX\), can in principle be accumulated in a streaming manner without storing the full posterior matrix.

The main difference from non-rigid CPD lies in the M-step. The dense CPD M-step estimates a Gaussian-kernel displacement field associated with the moving points. It requires forming and storing an \(M\times M\) Gaussian kernel matrix and solving the associated dense linear system, leading to \(\mathcal{O}(M^2)\) memory and \(\mathcal{O}(M^3)\) solver cost.

In contrast, Analytic-CPD solves a weighted structured analytic fitting problem. After barycentric condensation, the M-step involves a Taylor design matrix of size
\[
M_s\times S_{d,q},
\]
where \(M_s\le M\) is the number of retained soft-target pairs in the weighted fitting system. Forming and solving the corresponding dense least-squares system costs, up to lower-order feature-evaluation terms,
\[
\mathcal{O}\!\left(M_sS_{d,q}^{2}+S_{d,q}^{3}\right).
\]

For example, when \(q=10\),
\[
S_{2,10}=66,\qquad K_{2,10}=132,
\]
and
\[
S_{3,10}=286,\qquad K_{3,10}=858.
\]
Thus, for moderate analytic orders, the M-step system of Analytic-CPD is much smaller than a point-indexed kernel system whose size is determined by \(M\). The overall per-iteration cost of the direct Analytic-CPD implementation is therefore
\[
\mathcal{O}\!\left(
MNd
+
M_sS_{d,q}^{2}
+
S_{d,q}^{3}
\right).
\]

Acceleration techniques such as fast Gauss transform, sparse truncation, or low-rank kernel approximation can reduce the cost of CPD, but they introduce additional approximation choices. The distinction analyzed here is model-level: Analytic-CPD retains the CPD posterior layer while replacing the kernel-displacement M-step with a compact finite-dimensional structured analytic mapping M-step.

\section{Implementation Details}
\label{app:implementation-details}

This appendix provides additional implementation details for the Analytic-CPD algorithm. The notation follows the main paper:
\(X=\{x_n\}_{n=1}^{N}\subset\mathbb{R}^d\) denotes the fixed point set, \(Y^{(t)}=\{y_m^{(t)}\}_{m=1}^{M}\subset\mathbb{R}^d\) denotes the current moving point set,
\(P^{(t)}=(P_{mn}^{(t)})\in\mathbb{R}^{M\times N}\) denotes the CPD posterior matrix, and \(\sigma_t^2\) denotes the current variance parameter. The following details are not part of the abstract mathematical definition of the method, but are important for numerical stability and reproducibility.

\subsection{Data normalization and expansion center.}
Before registration, the input point sets are centered and scaled to a common normalized coordinate system, so that the coordinate magnitudes are of order one. This normalization is an essential part of the Analytic-CPD pipeline. Since the structured analytic mapping is represented by finite-order Taylor basis functions, the numerical conditioning of the design matrix and the relative strength of different-order terms depend directly on the coordinate
scale. A common normalized coordinate system prevents high-order monomials from being artificially amplified or suppressed by the raw data scale.

The Analytic-CPD iteration is performed in the normalized coordinate system. The expansion center of the structured analytic mapping is fixed after normalization and is chosen as the origin,
\[
c=\mathbf 0 .
\]
This choice improves the conditioning of the Taylor basis, especially when higher-order terms are used. In the current implementation, \(c\) is not estimated. It is treated as a fixed parameter of the analytic basis, while the Taylor coefficient set \(\Theta_t\) is estimated at each iteration from the current weighted soft-target fitting problem. Estimating or adapting the
expansion center is possible, but it would introduce an additional nonlinear component into the M-step and is left for future work.

After registration, the transformed moving point set can be mapped back to the original coordinate scale for visualization or application-specific use. Unless otherwise stated, the quantitative errors reported in the main experiments are computed in the common normalized coordinate system.

\subsection{Coefficient organization.}
The structured analytic mapping is implemented order by order. For ambient dimension \(d\) and analytic order \(q\), the coefficient set is stored as
\[
\{A_r\}_{r=0}^{q},
\qquad
A_r\in\mathbb R^{d\times C_{d,r}},
\]
where
\[
C_{d,r}
=
\binom{r+d-1}{d-1}
\]
is the number of monomials of exact total degree \(r\) in \(d\) variables. Thus, the \(r\)-th order coefficient block contains one column for each monomial of degree \(r\), and each column is a \(d\)-dimensional output coefficient vector.

For \(d=2\), each order-\(r\) coefficient block has size
\[
A_r\in\mathbb R^{2\times(r+1)}.
\]
For \(d=3\), each order-\(r\) coefficient block has size
\[
A_r\in\mathbb R^{3\times\binom{r+2}{2}}.
\]
The monomial ordering is fixed once and used consistently in basis evaluation, design-matrix assembly, coefficient fitting, and mapping evaluation. The ordering itself has no mathematical significance, as long as it is used consistently.

\subsection{Taylor basis evaluation.}
For two-dimensional point sets, let
\[
\delta_1=y_1-c_1,
\qquad
\delta_2=y_2-c_2.
\]
The order-\(r\) Taylor basis terms are evaluated as
\[
\frac{1}{r!}\binom{r}{k}
\delta_1^{\,r-k}\delta_2^{\,k},
\qquad
k=0,\ldots,r.
\]
Equivalently, this is the multi-index basis
\[
\frac{(y-c)^\alpha}{\alpha!},
\qquad
|\alpha|=r,
\]
written explicitly in two dimensions.

For three-dimensional point sets, the order-\(r\) basis consists of all monomials
\[
\frac{(y_1-c_1)^{\alpha_1}
      (y_2-c_2)^{\alpha_2}
      (y_3-c_3)^{\alpha_3}}
     {\alpha_1!\alpha_2!\alpha_3!},
\qquad
\alpha_1+\alpha_2+\alpha_3=r.
\]
The factorial and multinomial factors are included in the basis functions rather than in the coefficient matrices. This convention makes the estimated coefficients directly correspond to derivative-like Taylor coefficients.

\subsection{Posterior statistics and soft targets.}
The E-step computes the CPD posterior matrix
\[
P^{(t)}=(P_{mn}^{(t)})\in\mathbb R^{M\times N}.
\]
In addition to \(P^{(t)}\), the following posterior statistics are accumulated:
\[
\rho^{(t)}
=
P^{(t)}\mathbf 1_N
\in\mathbb R^M,
\qquad
\eta^{(t)}
=
(P^{(t)})^\top\mathbf 1_M
\in\mathbb R^N,
\]
and
\[
S_X^{(t)}
=
P^{(t)}X
\in\mathbb R^{M\times d},
\]
where \(X\in\mathbb R^{N\times d}\) also denotes the coordinate matrix whose rows are \(x_n^\top\). The scalar
\[
N_P^{(t)}
=
\mathbf 1_M^\top P^{(t)}\mathbf 1_N
=
\sum_{m=1}^{M}\sum_{n=1}^{N}P_{mn}^{(t)}
\]
is the effective posterior mass assigned to the Gaussian components.

For every row with \(\rho_m^{(t)}>0\), the soft target is computed as the posterior barycenter
\[
z_m^{(t)}
=
\frac{1}{\rho_m^{(t)}}
\sum_{n=1}^{N}
P_{mn}^{(t)}x_n.
\]
Equivalently, in matrix form,
\[
Z^{(t)}
=
\operatorname{Diag}(\rho^{(t)})^{-1}S_X^{(t)}
\]
on the rows with positive posterior mass. These statistics are also used in the matrix form of the variance update.

In the current implementation, the full posterior matrix is explicitly computed. However, the quantities \(\rho^{(t)}\), \(\eta^{(t)}\),
\(S_X^{(t)}\), and \(N_P^{(t)}\) can also be accumulated directly while traversing point pairs, without storing the full posterior matrix. This streaming variant can reduce memory usage for large point sets.

\subsection{Treatment of small posterior masses.}
If a moving point receives an extremely small posterior mass,
\[
\rho_m^{(t)}\approx 0,
\]
the corresponding soft target may be numerically unstable because it involves division by \(\rho_m^{(t)}\). Therefore, in implementation, rows satisfying
\[
\rho_m^{(t)}<\varepsilon_\rho
\]
may be omitted from the weighted analytic fitting system. This pruning is used only as a numerical safeguard. Before such negligible-weight rows are removed, the barycentric condensation is exact for the squared Euclidean loss.

Let \(M_s\) denote the number of retained source--target pairs after this filtering. If all moving points are retained, then \(M_s=M\).

\subsection{Degree feasibility and continuation.}
At each iteration, the analytic order \(q_t\) is selected by the
degree-continuation schedule. The selected order is also checked against a feasibility condition. The number of Taylor basis terms of total degree no larger than \(q_t\) is
\[
S_{d,q_t}
=
\binom{q_t+d}{d}.
\]
Since all output coordinates share the same Taylor design matrix, a necessary condition for avoiding an underdetermined weighted least-squares system is
\[
M_s
\ge
S_{d,q_t}.
\]
Equivalently, this condition can be written in scalar-equation form as
\[
dM_s
\ge
K_{d,q_t},
\qquad
K_{d,q_t}=dS_{d,q_t}.
\]
If the condition is violated, the order is reduced until the fitting system is not underdetermined. This condition does not guarantee good conditioning, but it prevents rank deficiency caused solely by an insufficient number of retained observations. For very high orders, additional conditioning checks may be used.

\subsection{Weighted least-squares solver.}
For retained source--target pairs \((\widehat y_m,\widehat z_m)\) with weights
\(\omega_m\), the analytic M-step solves
\[
\min_{\Theta}
\sum_{m=1}^{M_s}
\omega_m
\left\|
\widehat z_m
-
\mathcal A(\widehat y_m;\Theta,c)
\right\|^2.
\]
The weights are applied by multiplying each point equation by
\(\sqrt{\omega_m}\). All \(d\) coordinate equations associated with the same source--target pair share the same weight \(\omega_m\).

The structured analytic mapping is linear in its unknown coefficients. Therefore, after the Taylor design matrix is assembled, the M-step is a weighted linear least-squares problem. Although normal equations reveal the algebraic structure of the problem, explicitly inverting the normal matrix is not recommended, especially for high-order mappings or nearly degenerate point configurations. We therefore use rank-revealing least-squares solvers, such as pivoted QR or SVD, when numerical robustness is prioritized. For moderate-order and well-conditioned normalized data, a normal-equation implementation may also be used as an efficient option.

\subsection{Stopping criteria.}
The internal CPD-style residual is defined as
\[
E_{\mathrm{soft}}^{(t)}
=
\sqrt{d\,\sigma_t^2}.
\]
It is consistent with the posterior-weighted likelihood objective and is used for internal convergence monitoring. The iteration stops when the change of the internal residual, the variance, or the moving-set update becomes smaller than a prescribed tolerance, or when the maximum number of iterations is reached.

When ordered pointwise correspondences are available, as in synthetic benchmarks, an external pointwise RMSE can also be computed:
\[
E_{\mathrm{rmse}}^{(t)}
=
\sqrt{
\frac{1}{M}
\sum_{m=1}^{M}
\|x_m-y_m^{(t)}\|^2
}.
\]
This external RMSE is not used to estimate the posterior matrix or the analytic mapping coefficients. It is used only as an evaluation-side safeguard for ordered benchmarks: the best external state is recorded, and if the external RMSE rises clearly above the best recorded value, the iteration is stopped and the best external state is returned.

In the reported implementation, the external rebound condition is
\[
E_{\mathrm{rmse}}^{(t)}
>
(1+\tau_{\mathrm{rise}})E_{\mathrm{best}}
+
\epsilon_{\mathrm{rise}},
\]
with
\[
\tau_{\mathrm{rise}}=0.02,
\qquad
\epsilon_{\mathrm{rise}}=10^{-12},
\qquad
p_{\mathrm{rise}}=1.
\]
Here \(p_{\mathrm{rise}}\) is the number of consecutive rebound iterations allowed after the minimum iteration number. For unordered point sets without meaningful pointwise correspondences, this external guard is disabled and the best state is selected using the internal CPD-style residual.

\subsection{Direct and accelerated implementations.}
Unless otherwise specified, the reported implementation uses direct posterior computation and exact weighted analytic fitting. Approximation-based acceleration schemes, such as fast Gauss transform, sparse Gaussian truncation, or low-rank kernel approximation, are not used in the main controlled comparison. The reason is that such schemes introduce additional approximation
parameters and may trade accuracy, robustness, or implementation simplicity for speed.

Since Analytic-CPD retains the CPD posterior layer, fast Gaussian summation techniques for computing posterior statistics can in principle be applied to Analytic-CPD as well. The main comparison in this work therefore focuses on the exact model-level difference: CPD estimates a point-indexed Gaussian-kernel
displacement field, whereas Analytic-CPD estimates a compact structured analytic mapping.

\section{Weighted Analytic Least-Squares Details}
\label{app:weighted-analytic-ls}

This appendix provides the algebraic details of the weighted analytic fitting problem used in the Analytic-CPD M-step. The main text uses the compact matrix form, while here we give the corresponding normal equation, vectorized adjustment form, and correction formulation.

\subsection{Matrix normal equation}

Recall the weighted structured analytic fitting problem
\[
A^\star
=
\arg\min_A
\left\|
\Omega^{1/2}
\left(
\widehat Z-\Phi A^\top
\right)
\right\|_F^2 ,
\]
where
\[
\Phi\in\mathbb R^{M_s\times S_{d,q}},
\qquad
\widehat Z\in\mathbb R^{M_s\times d},
\qquad
\Omega=\operatorname{Diag}(\omega_1,\ldots,\omega_{M_s}).
\]
The corresponding normal equation is
\begin{equation}
\left(\Phi^\top\Omega\Phi\right)(A^\star)^\top
=
\Phi^\top\Omega\widehat Z .
\label{eq:app-weighted-sam-normal-equation}
\end{equation}
This equation shows that all output coordinates share the same weighted Taylor design matrix \(\Phi^\top\Omega\Phi\). Only the right-hand side differs across coordinate dimensions. Therefore, the \(d\)-dimensional mapping can be estimated by solving a weighted least-squares system whose basis size is \(S_{d,q}\).

\subsection{Vectorized adjustment form}

The same fitting problem can also be written in a vectorized adjustment form, which is closer to the indirect adjustment formulation used in \cite{feng2026structured}. Let
\[
\theta=\operatorname{vec}(A)\in\mathbb R^{K_{d,q}},
\qquad
K_{d,q}=dS_{d,q}.
\]
For each source point \(\widehat y_m\), define the design block
\begin{equation}
B_m
=
\phi_q(\widehat y_m;c)^\top\otimes I_d
\in\mathbb R^{d\times K_{d,q}},
\label{eq:app-sam-design-block}
\end{equation}
where \(\otimes\) denotes the Kronecker product. The pointwise observation equation is
\begin{equation}
B_m\theta \approx \widehat z_m .
\label{eq:app-point-observation-equation}
\end{equation}
Stacking all source-target pairs gives
\begin{equation}
B\theta\approx \ell ,
\label{eq:app-global-observation-equation}
\end{equation}
where
\[
B=
\begin{bmatrix}
B_1\\
B_2\\
\vdots\\
B_{M_s}
\end{bmatrix}
\in\mathbb R^{dM_s\times K_{d,q}},
\qquad
\ell=
\begin{bmatrix}
\widehat z_1\\
\widehat z_2\\
\vdots\\
\widehat z_{M_s}
\end{bmatrix}
\in\mathbb R^{dM_s}.
\]
The block-diagonal weight matrix is
\begin{equation}
W
=
\operatorname{Diag}
\left(
\omega_1 I_d,
\omega_2 I_d,
\ldots,
\omega_{M_s} I_d
\right).
\label{eq:app-block-weight-matrix}
\end{equation}
The vectorized weighted least-squares problem is
\begin{equation}
\theta^\star
=
\arg\min_{\theta}
\left\|
W^{1/2}(B\theta-\ell)
\right\|^2 ,
\label{eq:app-weighted-ls-vector-form}
\end{equation}
with normal equation
\begin{equation}
B^\top W B\,\theta
=
B^\top W\ell .
\label{eq:app-weighted-normal-equation}
\end{equation}

\subsection{Correction form}

The weighted analytic fitting problem can also be written in a correction form. Given a current coefficient vector \(\theta^{(0)}\), define the residual vector
\begin{equation}
\ell^{(0)}
=
\ell-B\theta^{(0)} .
\label{eq:app-correction-residual}
\end{equation}
The correction \(\Delta\theta\) is obtained by solving
\begin{equation}
\Delta\theta
=
\arg\min_{\Delta}
\left\|
W^{1/2}
\left(
B\Delta-\ell^{(0)}
\right)
\right\|^2 ,
\label{eq:app-weighted-correction-model}
\end{equation}
and the coefficients are updated as
\begin{equation}
\theta^{(1)}
=
\theta^{(0)}+\Delta\theta .
\label{eq:app-coefficient-update}
\end{equation}
Because the structured analytic mapping is linear in the coefficients, one correction step gives the exact weighted least-squares solution when the expansion center \(c\), the truncation order \(q\), and the basis are fixed. The correction form is consistent with the standard indirect adjustment viewpoint and is convenient for implementation.

For numerical stability, the least-squares system can be solved using rank-revealing solvers such as QR or SVD rather than by explicitly forming a matrix inverse. The normal equation forms in this appendix are presented to clarify the algebraic structure of the weighted analytic M-step.

\section{Baseline Implementations and Runtime Protocol}
\label{app:baseline-runtime-protocol}

This appendix provides additional details on the baseline implementations, runtime protocol, and thread-control settings used in the experiments. The purpose is to clarify reproducibility and to distinguish raw in-house single-threaded comparisons from official-implementation reference runtimes.

\subsection{Evaluation protocol.}
All methods are evaluated from their final registered point sets using the same external pointwise RMSE whenever ordered pointwise correspondences are available:
\[
E_{\mathrm{rmse}}
=
\left(
\frac{1}{M}
\sum_{m=1}^{M}
\|x_m-y_m^{\mathrm{reg}}\|^2
\right)^{1/2}.
\]
Method-specific internal objective values, stopping residuals, or likelihood values are not used for cross-method accuracy comparison. If a method performs its own normalization internally, its output is converted back to the common normalized coordinate system before computing the final RMSE.

\subsection{Hardware and runtime measurement.}
Unless otherwise specified, all experiments are conducted on the same computer equipped with an AMD Ryzen 7 5800H CPU at 3.20\,GHz and 32\,GB RAM. Runtime is measured as the wall-clock time required by the registration routine, excluding data loading, file writing, and visualization. For methods with stochastic components or random initialization, the reported statistics are computed over
the same deformation instances and random seeds whenever applicable.

\subsection{In-house C++ implementations.}
The in-house implementations include CPD, TPS-RPM, Analytic-ICP, and Analytic-CPD. These implementations are 64-bit single-threaded C++ programs. They do not use fast Gauss transform, low-rank kernel approximation, multi-threaded BLAS, GPU acceleration, or other external acceleration mechanisms. Therefore, their runtimes are directly comparable within the in-house implementation group.

For CPD, unless otherwise specified, we use
\[
\lambda=2.0,\qquad
\beta=2.0,\qquad
w=0.1.
\]
The method is evaluated in its direct form without fast Gaussian summation or low-rank acceleration.

For TPS-RPM, unless otherwise specified, we use
\[
\lambda=1,\qquad
T=1,\qquad
r=0.9,
\]
where \(T\) is the initial annealing temperature and \(r\) is the annealing ratio.

Analytic-ICP follows the staged rigid--affine--structured Taylor fitting pipeline described in~\cite{feng2026structured}. It uses the same structured analytic mapping model as Analytic-CPD, but correspondence estimation is based on ICP-style hard nearest-neighbor assignment rather than CPD-style posterior probabilities.

For Analytic-CPD, unless otherwise specified, we use
\[
T_{\max}=55,\qquad
q_{\max}=10,\qquad
w=0.1.
\]
The analytic degree follows the increasing-degree continuation schedule with decreasing stage lengths. The expansion center is fixed at the origin after normalization, and \(\sigma_0^2\) is initialized as the average pairwise squared distance, as in CPD.

\subsection{BCPD.}
BCPD is evaluated using its official Windows executable. Unless otherwise specified, we use its official similarity-plus-nonrigid registration mode with the parameter setting used in the main experiments. To reduce the influence of parallel numerical backends, the following environment variables are set before
running BCPD:
\[
\begin{array}{l}
\texttt{OMP\_NUM\_THREADS=1},\\
\texttt{OPENBLAS\_NUM\_THREADS=1},\\
\texttt{GOTO\_NUM\_THREADS=1}.
\end{array}
\]
This setting reduces thread-level parallel effects, but it does not make BCPD a raw-code-equivalent comparison with our in-house C++ programs. The official executable may still benefit from optimized numerical kernels, memory layout, SIMD vectorization, and implementation-specific stopping rules. Therefore, BCPD runtime is reported as an official-implementation reference value rather than a strict single-threaded raw-code comparison.

\subsection{ClusterReg.}
ClusterReg is evaluated in the three-dimensional experiments using its official implementation. Unless otherwise specified, we use
\[
\theta=0.5,\qquad
\beta=0.5,\qquad
\lambda=0.1,\qquad
m=\lceil0.3M\rceil .
\]
The released implementation explicitly uses parallel components for pairwise distance computation, fuzzy-membership updates, and related matrix operations, and it also invokes thread-control routines associated with its numerical backend. Therefore, ClusterReg runtime is reported as an official multi-threaded implementation reference rather than a raw single-threaded
implementation-level comparison with our in-house code.

\subsection{Interpretation of runtime comparisons.}
The runtime comparisons should therefore be interpreted in two groups. First, CPD, TPS-RPM, Analytic-ICP, and Analytic-CPD form a controlled in-house single-threaded comparison group. Second, BCPD and ClusterReg provide external official-implementation reference values. The accuracy comparison is unified by recomputing the final RMSE from the registered point sets, whereas the runtime
comparison is reported with explicit implementation-status annotations.



%\bibliographystyle{IEEEtran}

\bibliographystyle{IEEEtran}
\bibliography{references}

\vfill


\end{document}